\documentclass{article}

\usepackage{arxiv}

\usepackage[utf8]{inputenc} 
\usepackage[T1]{fontenc}    
\usepackage{hyperref}       
\usepackage{url}            
\usepackage{booktabs}       
\usepackage{amsfonts}       
\usepackage{nicefrac}       
\usepackage{microtype}      
\usepackage{lipsum}		
\usepackage{graphicx}
\usepackage{natbib}
\usepackage{doi}
\usepackage{multirow} 
\usepackage{amsmath}
\usepackage{graphicx}
\usepackage{array}
\usepackage{subcaption}
\title{Interpretable Classification via a Rule Network with Selective Logical Operators}

\date{} 					

\author{ \href{https://orcid.org/0009-0005-3204-0255}{\includegraphics[scale=0.06]{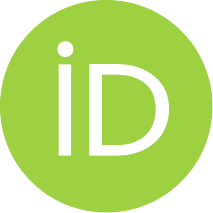}\hspace{1mm}Bowen Wei} \\
	Department of Computer Science\\
	George Mason University\\
	Fairfax, VA 22030 \\
	\texttt{bwei2@gmu.edu} \\
	\And
	 \href{https://orcid.org/0000-0002-3990-4774}{\includegraphics[scale=0.06]{orcid.pdf}\hspace{1mm}Ziwei Zhu} \\
	Department of Computer Science\\
	George Mason University\\
	Fairfax, VA 22030 \\
	\texttt{zzhu20@gmu.edu} 
}

\begin{document}
\maketitle

\begin{abstract}
We introduce the Rule Network with Selective Logical Operators (RNS), a novel neural architecture that employs \textbf{selective logical operators} to adaptively choose between AND and OR operations at each neuron during training. Unlike existing approaches that rely on fixed architectural designs with predetermined logical operations, our selective logical operators treat weight parameters as hard selectors, enabling the network to automatically discover optimal logical structures while learning rules. The core innovation lies in our \textbf{selective logical operators} implemented through specialized  Logic Selection Layers (LSLs) with adaptable AND/OR neurons, a Negation Layer for input negations, and a Heterogeneous Connection Constraint (HCC) to streamline neuron connections. We demonstrate that this selective logical operator framework can be effectively optimized using adaptive gradient updates with the Straight-Through Estimator to overcome gradient vanishing challenges. Through extensive experiments on 13 datasets, RNS demonstrates superior classification performance, rule quality, and efficiency compared to 25 state-of-the-art alternatives, showcasing the power of RNS in rule learning. Code and data are available at \url{https://anonymous.4open.science/r/RNS_-3DDD}.
\end{abstract}

\section{Introduction}
Unlike explainable models~\citep{rudin2019stopexplainingblackbox}, which clarify black-box predictions by analyzing input feature contributions, interpretable models~\citep{molnar2020interpretable} are inherently transparent, enabling direct human comprehension of their inference process (e.g., decision trees). This transparency is crucial for ensuring reliability, safety, and trust, especially in high-stakes domains like healthcare, finance, and law, where justifications are as important as the outcomes.

Rule-based models~\citep{yin2003cpar,10.5555/645527.657305,cohen1995fast,quinlan2014c4,yang2017scalable,marton2023grande} are widely recognized for their inherent interpretability. A range of traditional approaches has been proposed, including example-based rule learning algorithms~\citep{Michalski1973AQVAL1ComputerIO}, systems for learning first-order Horn clauses and Conjunctive Normal Form (CNF) or Disjunctive Normal Form (DNF) rules~\citep{quinlan1990learning,cohen1995fast,10.5555/645527.657305, Michalski1973AQVAL1ComputerIO,10.1023/A:1022641700528,mooney1995encouraging,beck2023layerwise}, ensemble methods and fuzzy rule systems~\citep{ke2017lightgbm,breiman2001random}, and Bayesian frameworks~\citep{Letham_2015,wang2017bayesian,yang2017scalable}. Despite extensive development, these traditional models often struggle with limitations in prediction accuracy~\citep{quinlan2014c4,loh2011classification,cohen1995fast}, suboptimal interpretability~\citep{ke2017lightgbm,breiman2001random}, or poor scalability to large datasets~\citep{Letham_2015,wang2017bayesian,yang2017scalable}.

Neural networks offer significant potential for representing and learning interpretable logical rules due to their expressiveness, generalization capability, robustness, and data-driven nature~\citep{wang2020transparent,wang2021scalable, Wang_2024,10.1145/3583780.3614884}. This enables the automatic and efficient learning of complex rules at scale, combining the computational power of neural methods with the clarity and interpretability of rule-based reasoning.

\textbf{A major limitation in current rule-based neural networks is their reliance on fixed logical operators and non-learnable structural design.} These models enforce a predetermined architecture, with layers configured with static logical operations (either AND or OR) that cannot adapt during training~\citep{wang2020transparent,beck2023layerwise}. \textbf{The inability to dynamically select optimal logical operators} significantly compromises both model performance and rule quality. Furthermore, their training methods are often heuristic-driven~\citep{beck2023layerwise} or require repeated forward and backward passes~\citep{wang2020transparent,wang2021scalable, Wang_2024}, reducing efficiency. These constraints severely limit the models' ability to discover the most appropriate logical structures for different datasets and generate sophisticated, high-quality rules for reliable interpretations.

\begin{figure*}[ht]
    \centering
    \includegraphics[width=0.7\textwidth]{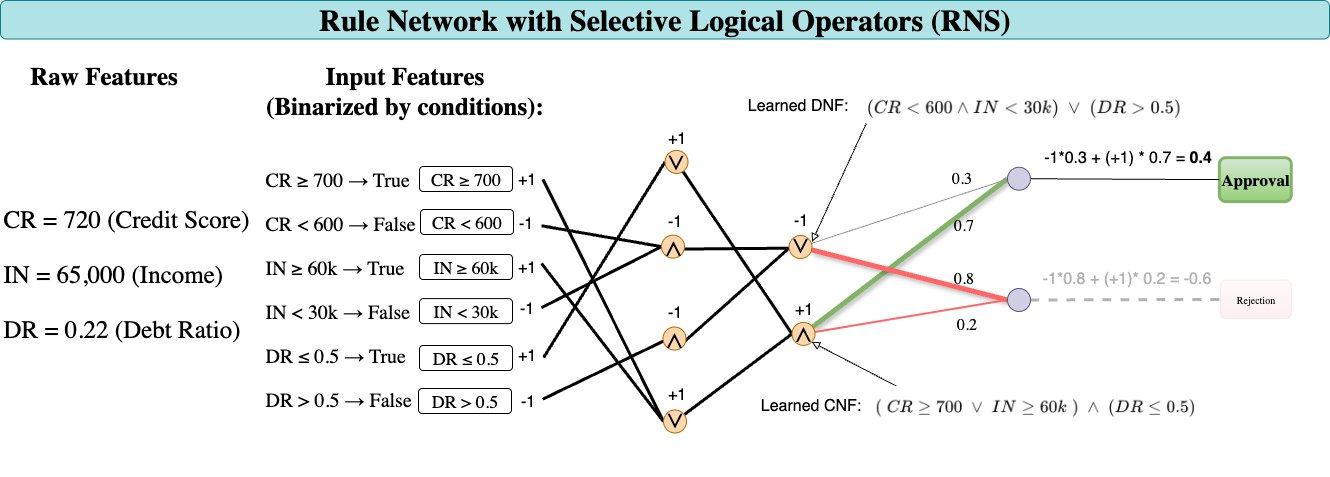}
    \caption{\textbf{Loan-application example of the RNS.} Raw attributes are binarized and passed through two learnable LSLs (AND/OR selected per neuron). The output layer learns class-specific weights from each clause to Approval/Rejection and sums them.  Aggregating with learned weights gives Approval: $0.4$ and Rejection: $-0.6$, yielding an \emph{Approved} decision.}
    \label{fig:intro}

\end{figure*}

To address these fundamental limitations, we introduce the \textbf{Rule Network with Selective Logical Operators (RNS)}, which automatically learns neurons as AND or OR operators and simultaneously learns the connections between neurons, thereby forming CNF/DNF rules. RNS enables the network to uncover optimal logical structures while learning rules in both CNF and DNF formats, ensuring accurate classifications and interpretable insights through a transparent inference process. Figure~\ref{fig:intro} illustrates this process with a loan application example: raw features are binarized and passed through two learnable Logic Selection Layers (LSLs). The output layer assigns class-specific weights to each rule and aggregates them into final decisions. In this example, aggregation yields Approval $0.4$ and Rejection $-0.6$, resulting in an Approved decision.

Key innovations of RNS include: two specialized Logic Selection Layers (LSLs), where weight parameters serve as hard selectors to determine AND or OR operations at each neuron; a Negation Layer for implementing logical negation; a Heterogeneous Connection Constraint (HCC) to efficiently learn neuron connections; and the use of the Straight-Through Estimator (STE) for optimizing the discrete network. Through this design, RNS offers four unique advantages: (1) \textbf{Adaptive Logical Structure}: Neurons dynamically select logical operations, enabling flexible CNF or DNF rule learning; (2) \textbf{Complete Logic Operation}: Incorporates negation to achieve functional logic completeness; (3) \textbf{Efficient Rule Discovery}: Efficiently identifies optimal rules within the large search space of neuron connections; and (4) \textbf{Reliable Optimization}: Effectively optimizes discrete networks with non-differentiable components, mitigating the risk of gradient vanishing. Unlike existing approaches that rely on complex logical activation functions, RNS addresses the gradient vanishing problem through a lightweight design that combines STE with carefully designed logical activation functions.

Experiments on 13 datasets against 25 baselines demonstrate that RNS achieves superior performance in three critical aspects: (1) \textbf{rule quality}, (2) \textbf{}, and (3) \textbf{training efficiency}. Notably, RNS substantially outperforms state-of-the-art (SOTA) methods in rule quality, \textbf{a factor even more critical than prediction accuracy for interpretable classification}. Its rule sets are not only more \textbf{accurate} and \textbf{diverse} but also considerably lower in \textbf{complexity}. Code is available at \url{https://anonymous.4open.science/r/RNS_-3DDD}.

\section{Preliminaries}

\subsection{Problem Formulation}

A set of instances is represented as $\mathcal{X}$, where each instance $\mathbf{x} \in \mathcal{X}$ is defined by a feature vector $\mathbf{x} = [x_1, x_2, \ldots, x_n]$, comprising continuous or categorical features. Each instance is associated with a discrete class label $y$. The classification task aims to learn a function $f: \mathbf{x} \rightarrow y$. In this work, we design $f$ as a rule-based model that automatically learns feature-based rules for prediction, with the learned rules providing inherent interpretability.

\subsection{Feature Binarization}
\label{sec:binarization}

Logical rules operate on Boolean truth values, so each feature must be represented in a True/False form. Since raw feature values cannot be directly evaluated as Boolean, we transform them into binary literals using conditions such as thresholds for continuous features or one-hot encodings for categorical features. For a categorical feature $b_i$, we use one-hot encoding to produce the corresponding binary vector $\mathbf{b}_i$. For a continuous feature $c_j$, we adopt the feature binning method from~\citep{wang2021scalable, Wang_2024}: a set of $k$ upper bounds $[\mathcal{H}_{j,1}, \dots, \mathcal{H}_{j,k}]$ and $k$ lower bounds $[\mathcal{L}_{j,1}, \dots, \mathcal{L}_{j,k}]$ is randomly sampled from the value range of $c_j$, and the binary representation of $c_j$ is then derived as $\tilde{\mathbf{c}} = [q(c_j - \mathcal{L}_{j,1}), \ldots, q(c_j - \mathcal{L}_{j,k}), q(\mathcal{H}_{j,1} - c_j), \ldots, q(\mathcal{H}_{j,k} - c_j)]$, where $q(x) = 1$ if $x > 0$ and $q(x) = -1$ otherwise. The full binarized input feature vector is $\tilde{\mathbf{x}} = [\mathbf{b}_1, \ldots, \mathbf{b}_p, \tilde{\mathbf{c}}_1, \ldots, \tilde{\mathbf{c}}_t]$. Beyond this random binning strategy, other heuristic- and learning-based methods can also be applied: AutoInt~\citep{zhang2023learning} learns optimized bins jointly with model training, KInt~\citep{dougherty1995supervised} partitions values into clusters using K-means, and EntInt~\citep{wang2020transparent} selects bins that minimize label uncertainty. Designing an algorithm for how to bin features is beyond the scope of this work; instead, in Section~\ref{Ablation_HCC}, we empirically compare different binning strategies.

\subsection{Normal Form Rules as Model Interpretation}\label{model-interpretation via normal forms}
Propositional logic is crucial to mathematical logic, focusing on propositions—statements with definite truth values -- and logical connectives (e.g., $\land$, $\lor$, $\neg$) to form logical expressions. Among these, Conjunctive Normal Form (CNF) and Disjunctive Normal Form (DNF) are fundamental constructs: a formula $z$ is in CNF if $z = \bigwedge_i \bigvee_j l_{ij}$, and in DNF if $z = \bigvee_i \bigwedge_j l_{ij}$, where each literal $l_{ij}$ represents an atom or its negation. These standardized formats simplify logical deduction and analysis, facilitating efficient conversion of arbitrary logical expressions into forms suitable for both theoretical and practical applications. Building on this foundation, our work develops a logical rule-based classification model that predicts outcomes using automatically learned CNF and DNF expressions (rules). Specifically, we learn a set of logical rules $\mathfrak{R} = \{z_1, \ldots, z_m\}$, where each rule $z$ is constructed from binary features and their negations as literals. For each rule, we also learn a set of contribution scores $\{s_{z,1}, \ldots, s_{z, Y}\}$ that quantify its impact on each class $y \in \{1, \ldots, Y\}$. Given an input $\tilde{\mathbf{x}}$, we evaluate the truth value of each rule in $\mathfrak{R}$ and compute the logit for class $k$ as $\hat{y}_k = \sum_{i=1}^m z_i \times s_{z_i,k}$. The learned rules thus provide a transparent representation of the inference process, ensuring both accuracy and interpretability.

\noindent\textbf{Example: Loan Approval Decision.}
Consider Figure~\ref{fig:intro}, where an applicant with a credit score CR=720, income of IN=\$65,000, and debt ratio DR=0.22 applies for a loan. The RNS first binarizes these features into logical conditions using learned thresholds, yielding the $\pm1$ encoded values: CR$\geq$700$\rightarrow$+1, CR$<$600$\rightarrow$-1, IN$\geq$60k$\rightarrow$+1, IN$<$30k$\rightarrow$-1, DR$\leq$0.5$\rightarrow$+1, DR$>$0.5$\rightarrow$-1. Two LSL neurons then process these inputs: the first learns a DNF rule $(CR<600 \land IN<30k) \lor (DR>0.5)$ capturing rejection conditions, which evaluates to -1 (false) since neither conjunction holds; the second learns a CNF rule $(CR\geq700 \lor IN\geq60k) \land (DR\leq0.5)$ capturing approval conditions, which evaluates to +1 (true) as both clauses are satisfied. The output layer aggregates these logical decisions with learned weights, computing approval score $(-1)\times0.3+(+1)\times0.7=0.4$ and rejection score $(-1)\times0.8+(+1)\times0.2=-0.6$. Since the approval score exceeds the rejection score, the model outputs ``Approved''.

\section{Method}
\label{sec:method}

\subsection{Overall Structure}

To achieve the goal in Section~\ref{model-interpretation via normal forms}, we propose a rule network that supports logical rule computation and can be trained end-to-end. We identify four key challenges: (1) How to enable each neuron to select a logical operator dynamically? (2) How to incorporate negation, ensuring functional logic completeness? (3) How to efficiently learn logical connections within a large search space? (4) How to support effective optimization despite non-differentiable operations and gradient vanishing?

\begin{figure*}[t]
    \centering
    \includegraphics[width=0.8\linewidth]{ 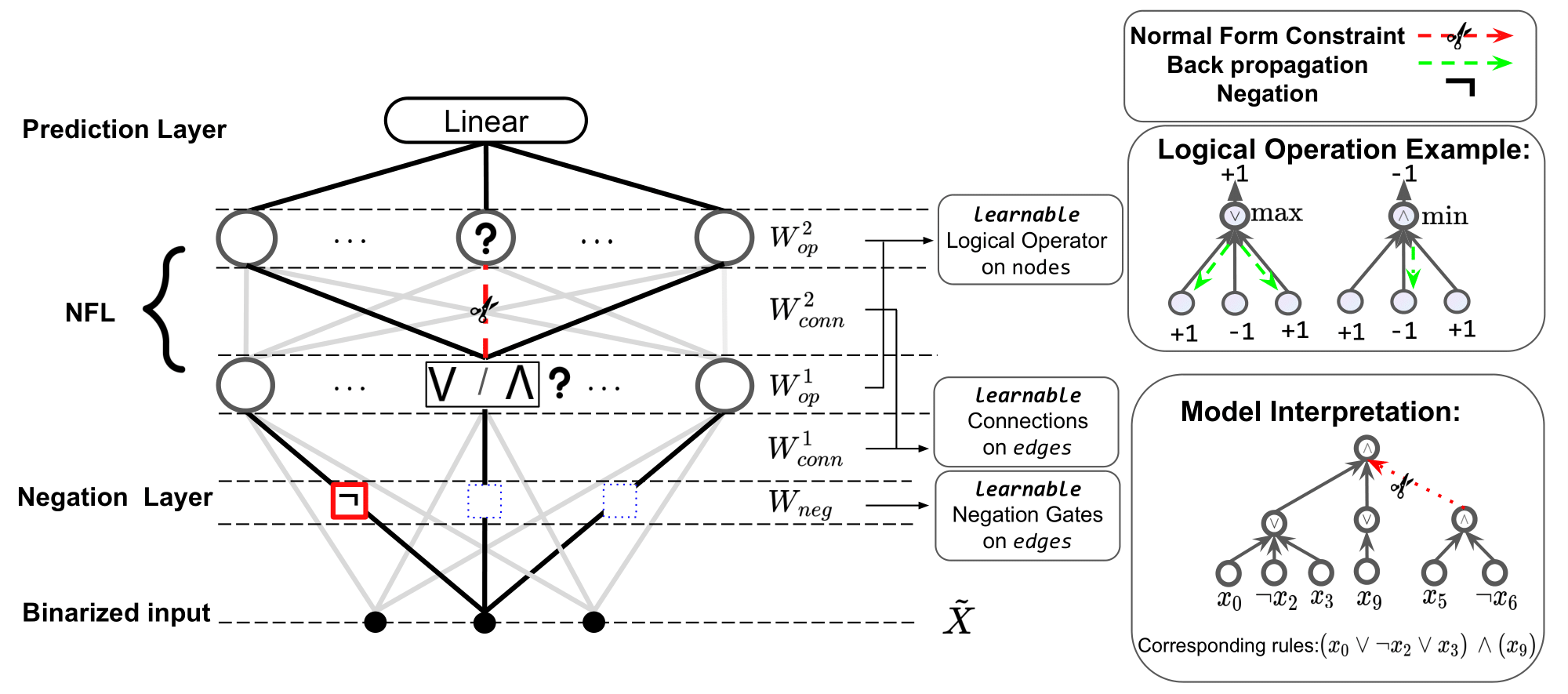}
    \caption{\textbf{RNS architecture and rule extraction.} From binarized inputs $\tilde{X}$, RNS learns interpretable CNF/DNF rules through a negation layer and two logic selection layers (NFL) constrained by normal form. The negation layer applies learnable gates ($W_{neg}$) to select literals or their negations. NFL layers use learnable connection weights ($W^1_{conn}$, $W^2_{conn}$) and operator weights ($W^1_{op}$, $W^2_{op}$) to construct logical conjunctions and disjunctions. The normal form constraint enforces alternating operators between layers via masked back-propagation. A linear prediction layer aggregates rule activations for classification. The extracted rules (bottom-right) are human-readable logical formulas combining selected literals, e.g., $(x_0 \lor \lnot x_2 \lor x_3) \land (x_9)$.}
    \label{fig:model_overview}
\end{figure*}

\textbf{Model Structure.}  
To address the first challenge, we design the  Logic Selection Layer (LSL) with neurons that are learnable to select AND or OR logical operations, enabling connections to represent rules. To address the second challenge, a \textit{Negation Layer} with learnable gates that flip the sign of input features when necessary. For the third challenge, we propose a \textit{Normal Form Constraint} (HCC) that restricts valid connections across LSLs to reduce the search space. These components are trained using the proposed logical activation functions described in Section~\ref{optimization}, which enables faithful logical computation with robust gradient flow. Non-differentiable steps such as $\mathrm{sign}(\cdot)$ are handled using the Straight-Through Estimator (STE). The overall architecture of RNS is shown in Figure~\ref{fig:model_overview}.

Given an input sample, the input feature vector $\mathbf{x}$ is binarized into $\mathbf{\tilde{x}}$ and passed through the first LSL, where conjunction or disjunction operations are applied depending on the neuron’s operator selector. The Negation Layer operates on these connections to produce literals using original or negated features. The second LSL applies the next level of logic, generating a set of interpretable rules $\mathfrak{R}$, with each output neuron representing a learned logical clause. These rule outputs are then linearly combined with learned weights to produce class logits.

\smallskip
\textbf{Binary Neuron Values.}  
A key characteristic of RNS is that all neuron values (excluding predicted logits) are constrained to $\pm1$, as is the binarized feature input described in Section~\ref{sec:binarization}. Unlike existing works~\citep{wang2020transparent,wang2021scalable, Wang_2024} that use $0/1$, this design avoids the gradient vanishing problem, significantly improving the optimization process, detailed in Section~\ref{optimization} and Appendix~\ref{gradient_vanishing}.

\subsection{Logic Selection Layer (LSL)}

\textbf{Neuron Operator Selection.}
To implement the nested structure of CNF and DNF, we stack two Logic Selection Layers (LSLs). Each LSL contains $K$ neurons ${u_1, \ldots, u_K}$, each as a learnable AND ($\land$) or OR ($\lor$) operator. The selection for the $K$ neurons is parameterized by a weight vector $\mathbf{w}_{op} \in \mathbb{R}^K$. For a neuron $u_i$, the operator is determined by the sign of its weight: $\tilde{\mathbf{w}}_{op}^i=\mathrm{sign}(\mathbf{w}_{op}^i)\in\{+1,-1\}$. The selection mechanism is defined as:
\begin{equation}
        u_i=
    \begin{cases}
      \land, & \text{if}\ \tilde{w}_{op}^i=1 \\
      \lor, & \text{otherwise}
    \end{cases}
\end{equation}

\textbf{Neuron Connection.} Similar to an MLP, we learn a weight matrix $\mathbf{W}_{conn}$ to model connections between neurons in consecutive layers, where $\mathbf{W}_{conn}^{i,j} \in \mathbb{R}$ denotes the weight from neuron $u_j$ to $u_i$. RNS uses two such matrices: one for the \textit{(input layer, 1st LSL)} and another for the \textit{(1st LSL, 2nd LSL)}. We binarize the weights via sign: $\widetilde{\mathbf{W}}_{conn}^{i,j} = \mathrm{sign}(\mathbf{W}_{conn}^{i,j}) \in \{+1, -1\}$. A connection exists if $\widetilde{\mathbf{W}}_{conn}^{i,j} = 1$; otherwise, it is inactive. The output $v_i$ of a neuron $u_i$ in the LSL is defined as:
\begin{equation}
    v_i=
    \begin{cases}
      \underset{\widetilde{\mathbf{W}}_{conn}^{i,j}=1}{\bigwedge}v_j, & \text{if}\,\, u_i=\land \\
       \underset{\widetilde{\mathbf{W}}_{conn}^{i,j}=1}{\bigvee}v_j, & \text{if}\,\, u_i=\lor
    \end{cases}
\end{equation}
where $v_j$ are the outputs of neurons from the previous layer connected to $u_i$.

\textbf{Logical Activation Functions.} In propositional logic with $m$ inputs, the AND operation outputs $+1$ only if all connected neurons are $+1$, while the OR operation outputs $+1$ if at least one connected neuron is $+1$. Based on these principles, we use \textit{min} and \textit{max} functions as logical activation functions in RNS. The AND ($\land$) operation is defined as the minimum of connected inputs, and the OR ($\lor$) operation as the maximum:
\begin{equation}\label{logic_output}
    v_i=
    \begin{cases}
      \underset{\widetilde{\mathbf{W}}_{conn}^{i,j}=1}{\mathrm{min}}\, v_j, & \text{if}\,\, u_i=\land \\
       \underset{\widetilde{\mathbf{W}}_{conn}^{i,j}=1}{\mathrm{max}}\, v_j, & \text{if}\,\, u_i=\lor
    \end{cases}
\end{equation}
Compared to the accumulative-multiplication-based activation functions proposed in prior works~\citep{wang2020transparent,wang2021scalable, Wang_2024}, this min-max-based method is more friendly for backpropagation and prevents gradient vanishing, detailed in Section~\ref{optimization}.

\subsection{Negation Layer}
Most existing works~\citep{wang2020transparent,wang2021scalable, Wang_2024} support only \{AND, OR\}, which is not functionally complete for propositional logic~\citep{mendelson1997schaum,enderton2001mathematical}. To enable RNS to theoretically express any logical rule, we include the functionally complete set of operators \{AND, OR, NEGATION\} by introducing a negation layer to perform negation operations on features.

Since negation operates on input features, we apply the negation layer after the input layer. For each connection between $v_j$ (from $\tilde{\mathbf{x}}$) and a neuron $u_i$ in the 1st LSL, we add a negation gate to decide whether to input the original value $v_j$ or its negation $\neg v_j$ to $u_i$. Each gate is parameterized by a weight $\mathbf{W}_{neg}^{i,j} \in \mathbb{R}$, and its binary version is obtained as $\widetilde{\mathbf{W}}_{neg}^{i,j} = \mathrm{sign}(\mathbf{W}_{neg}^{i,j}) \in \{+1, -1\}$. The negation operation is then defined as: $Neg(v_j, \widetilde{\mathbf{W}}_{neg}^{i,j}) = v_j \times \widetilde{\mathbf{W}}_{neg}^{i,j}$.

\subsection{Optimization}
\label{optimization}
Learning a binary neural network is difficult because its discrete functions—$\mathrm{sign}(\cdot)$, $\mathrm{max}(\cdot)$, and $\mathrm{min}(\cdot)$—are non-differentiable, making gradient computation challenging.

\textbf{Gradient of the Sign Function.} Inspired by the approach of searching for discrete solutions in continuous space~\citep{courbariaux2015binaryconnect}, we use the Straight-Through Estimator (STE) algorithm to propagate gradients through non-differentiable operations during backpropagation. The STE assumes that the derivative of the sign function with respect to its input is 1, enabling the gradient to ``pass through'' the non-differentiable operation unchanged: $\frac{\partial \mathrm{sign}(x)}{\partial x} = 1$.

\textbf{Gradient of Logical Activation Functions.} The max and min operations identify the maximum or minimum value across input neurons. While prior work~\citep{lowe2022logical} primarily applies these functions in continuous pairwise scenarios, we extend them to handle discrete and multiple inputs. In pairwise cases, logical operators select one or two inputs, making gradient assignment and parameter updates straightforward. For multiple inputs, logical operators may involve several input neurons simultaneously, requiring careful gradient distribution to ensure proper updates. During the backward pass, uniform gradient updates are applied when multiple neurons share the same maximum or minimum value. This ensures fairness by equally distributing gradients among these neurons, promoting stability during optimization. Formally, for an input vector $\mathbf{x}$ with elements ${x_1, x_2, \ldots}$, the gradient of an input element $x_i$ is:
\begin{equation}
\begin{aligned}
    \frac{\partial \max(\mathbf{x})}{\partial x_i} &= \frac{1}{\sum_j\mathbb{I}(x_j=\max(\mathbf{x}))},\quad \\
    \frac{\partial \min(\mathbf{x})}{\partial x_i} &= \frac{1}{\sum_j\mathbb{I}(x_j=\min(\mathbf{x}))}.
    \label{equ:max_min_gradient}
\end{aligned}
\end{equation}

\textbf{Gradient Vanishing.} Our design addresses the gradient vanishing problem that affects existing methods~\citep{wang2020transparent,wang2021scalable}, caused by two main factors. First, these methods use binary states $\{0, 1\}$, resulting in many neurons outputting $0$. These zero outputs propagate during the forward pass, nullifying gradients in the backward pass. Second, these methods rely on cumulative multiplications for logical activation functions like AND and OR. For instance, the AND operation is defined as: $AND(\mathbf{x})=\prod_{i} x_i$ when $x_i\in\{0,1\}$. The gradient of the AND operation with respect to an input $x_j$ is given by: $\partial AND(\mathbf{x})/\partial x_j=\prod_{i\neq j} x_i$. If any $x_i = 0$, then the gradient becomes $0$. 

In contrast, we adopt $\pm1$ for defining binary states in the neural network, preventing the generation of 0 outputs and 0 gradients. Furthermore, instead of designing complex continuous activation functions, our simple yet effective max/min functions avoid the use of multiplication. The straightforward uniform gradient update introduced in Equation~\ref{equ:max_min_gradient} alleviates the problem of gradient vanishing. A comprehensive convergence analysis with both theoretical and empirical evidence is provided in Appendix~\ref{appendix:convergence}, and a more detailed analysis of gradient behavior is provided in Appendix~\ref{gradient_vanishing}.

\subsection{Heterogeneous Connection Constraint (HCC)}
Next, we turn our attention to a critical challenge in RNS -- learning the connections between the two LSLs is computationally intensive. With $K$ neurons in each LSL, learning and determining $K^2$ potential connections significantly reduces the model's efficiency and efficacy. To mitigate this, we design a normal form constraint that reduces the search space for learning the connections between two LSLs. Specifically, since CNF and DNF have a nested structure -- where operations at the two levels must differ (CNF is a conjunction of disjunction rules, while DNF is a disjunction of conjunction rules) -- we enforce a constraint that only neurons of different types from the two LSLs can be connected. For neurons $u_i$ and $u_j$ from two LSLs, we define a mask parameter $M^{i,j}$ as:
\begin{equation}
    M^{i,j} = \tilde{\mathbf{w}}^i_{op} \oplus \tilde{\mathbf{w}}^j_{op} = - \tilde{\mathbf{w}}^i_{op}\tilde{\mathbf{w}}^j_{op},
\end{equation}
Where $\oplus$ denotes the XOR operation. A connection exists between $u_i$ and $u_j$ if $M^{i,j}\times\widetilde{\mathbf{W}}_{conn}^{i,j}=1$, otherwise, no connection is formed. During optimization, we update $\mathbf{W}_{conn}^{i,j}$ only when $M^{i,j}=1$, ensuring that connections are limited to valid neuron pairs and reducing the computational complexity of the learning process.

Assuming there are $C_1$ and $C_2$ conjunction neurons in two LSLs respectively, and $D_1$ and $D_2$ disjunction neurons respectively ($C_1+D_1=C_2+D_2=K$), then potential connections under HCC are $C_1D_2+C_2D_1\leq K^2$. The empirical study in Section~\ref{Ablation_HCC} demonstrates that HCC benefits both the efficiency and efficacy of the model.

\section{Experiment}

\begin{table*}[ht]
\centering
\small
\setlength{\tabcolsep}{1.3pt}
\renewcommand{\arraystretch}{0.8}
\caption{F1 scores (\%) across 13 datasets. The top rows show non-interpretable models; the bottom rows show interpretable models. Bold denotes best performance. N-Mean and AvgRank are computed across all methods. }
\begin{tabular}{l|ccccccccc|cccc|cc}
\toprule
\textbf{Model} & \textbf{adult} & \textbf{bank} & \textbf{banknote} & \textbf{chess} & \textbf{c-4} & \textbf{letRecog} & \textbf{magic} & \textbf{tic-tac-toe} & \textbf{wine} & \textbf{activity} & \textbf{dota2} & \textbf{fb} & \textbf{fashion} & \textbf{N-Mean} $\uparrow$ & \textbf{AvgRank} $\downarrow$ \\
\midrule
\multicolumn{16}{c}{\textit{Non-interpretable Models}} \\
\midrule
BNN        & 77.26 & 72.49 & 99.64 & 78.55 & 61.94 & 81.06 & 79.50 & 98.92 & 95.77 & 97.86 & 54.76 & 85.94 & 85.33 & 0.800(14) & 15.31(15) \\
FT         & 79.01 & 77.04 & 99.93 & 80.64 & 72.45 & 97.17 & 85.95 & 97.84 & 94.63 & 98.56 & 59.70 & 86.52 & 89.23 & 0.843(7) & 8.62(8) \\
LGBM       & 80.36 & 75.28 & 99.48 & 80.58 & 70.53 & 96.51 & 86.67 & 99.40 & 98.44 & 99.41 & 58.81 & 85.87 & 89.91 & 0.845(6) & 7.46(6) \\
NODE       & 80.55 & 77.16 & 99.93 & 80.64 & 71.90 & \textbf{97.20} & 86.20 & \textbf{100.0} & 97.78 & 97.70 & 60.01 & 87.93 & 89.63 & 0.850(3) & 5.77(3) \\
PLNN       & 73.55 & 72.40 & \textbf{100.0} & 77.85 & 64.55 & 92.34 & 83.07 & \textbf{100.0} & 76.07 & 98.27 & 59.46 & 89.43 & 89.36 & 0.806(13) & 11.77(11) \\
RF         & 79.22 & 72.67 & 99.40 & 75.00 & 62.72 & 96.59 & 86.48 & \textbf{100.0} & 98.31 & 97.80 & 57.39 & 87.49 & 88.35 & 0.828(10) & 10.69(10) \\
SAINT      & 79.31 & 75.60 & 99.04 & 79.37 & 72.85 & 96.72 & 85.46 & 96.95 & 95.50 & 98.94 & 59.58 & 88.79 & 89.69 & 0.842(8) & 8.77(9) \\
STG        & 76.38 & 60.24 & 90.49 & 60.33 & 62.17 & 78.18 & 68.38 & 93.43 & 94.39 & 84.89 & 41.02 & 60.18 & 81.15 & 0.697(24) & 23.15(26) \\
SVM        & 63.63 & 66.78 & \textbf{100.0} & 79.58 & 69.85 & 95.57 & 79.43 & \textbf{100.0} & 96.05 & 98.67 & 57.76 & 87.20 & 84.46 & 0.808(11) & 12.54(12) \\
TabNet     & 80.94 & 77.54 & 96.34 & 80.78 & 72.13 & 93.44 & 85.08 & \textbf{100.0} & 98.37 & 96.07 & 59.16 & 86.53 & 88.98 & 0.840(9) & 8.00(7) \\
TabTrans & 79.35 & 75.67 & 93.78 & 77.64 & 71.24 & 81.75 & 81.03 & 95.77 & 95.16 & 93.07 & 59.12 & 86.86 & 87.65 & 0.807(12) & 14.85(14) \\
VIME       & 78.24 & 76.59 & 98.91 & 76.27 & 52.27 & 80.05 & 83.19 & 93.48 & 92.18 & 92.37 & 57.01 & 88.52 & 81.12 & 0.783(20) & 17.08(18) \\
XGB        & 80.64 & 74.71 & 99.55 & 80.66 & 70.65 & 96.38 & \textbf{86.69} & 99.48 & 97.78 & 99.38 & 58.53 & 88.90 & 89.82 & 0.847(4) & 6.38(4) \\
\midrule
\multicolumn{16}{c}{\textit{Interpretable Models}} \\
\midrule
C4.5       & 77.77 & 71.24 & 98.45 & 79.90 & 61.66 & 88.20 & 82.44 & 98.45 & 95.48 & 94.24 & 52.08 & 80.71 & 80.49 & 0.793(16) & 17.31(19) \\
CART       & 77.06 & 71.38 & 97.85 & 79.15 & 61.24 & 87.62 & 81.20 & 97.85 & 94.39 & 93.35 & 51.91 & 81.50 & 79.61 & 0.787(18) & 18.85(21) \\
CORELS     & 70.56 & 66.86 & 98.49 & 24.86 & 51.72 & 61.13 & 77.37 & 98.49 & 97.43 & 51.61 & 46.21 & 34.93 & 38.06 & 0.582(25) & 22.08(25) \\
CRS        & 80.95 & 73.34 & 94.93 & 80.21 & 65.88 & 84.96 & 80.87 & 94.93 & 97.78 & 95.05 & 56.31 & 91.38 & 66.92 & 0.795(15) & 13.85(13) \\
DRNET      & 76.83 & 68.45 & 97.92 & 76.34 & 62.18 & 88.73 & 79.26 & 95.87 & 92.41 & 93.52 & 52.67 & 81.93 & 80.37 & 0.780(21) & 19.54(22) \\
KNN        & 77.57 & 75.61 & \textbf{100.0} & 75.21 & 65.18 & 91.92 & 77.92 & 92.33 & 96.24 & 97.09 & 51.61 & 68.61 & 82.35 & 0.785(19) & 16.00(16) \\
LORD       & 80.72 & 74.90 & 99.51 & 80.61 & 70.77 & 96.32 & 86.52 & 99.45 & 97.85 & 99.32 & 58.61 & 88.75 & 89.76 & 0.847(5) & 7.00(5) \\
LR         & 78.43 & 69.81 & 98.82 & 33.06 & 49.87 & 72.05 & 75.72 & 98.82 & 95.16 & 98.47 & 59.34 & 88.62 & 84.53 & 0.742(23) & 16.54(17) \\
PTURS      & 77.92 & 71.84 & 96.73 & 77.58 & 65.43 & 89.37 & 76.18 & 95.42 & 90.85 & 94.76 & 53.82 & 84.27 & 82.19 & 0.789(17) & 18.38(20) \\
RIPPER     & 74.69 & 69.76 & 96.00 & 70.95 & 64.78 & 92.94 & 77.92 & 97.79 & 89.16 & 88.08 & 55.67 & 64.18 & 78.66 & 0.758(22) & 20.38(23) \\
RRL        & 80.42 & 77.18 & \textbf{100.0} & 79.66 & 72.01 & 96.14 & 86.24 & \textbf{100.0} & 98.37 & 98.96 & 60.08 & 90.11 & 89.64 & 0.851(2) & 4.62(2) \\
SBRL       & 79.88 & 72.67 & 94.44 & 26.44 & 48.54 & 64.32 & 82.52 & 94.44 & 95.84 & 11.34 & 34.83 & 31.16 & 47.38 & 0.552(26) & 21.15(24) \\
\textbf{RNS }& \textbf{81.24} & \textbf{77.62} & \textbf{100.0} & \textbf{81.19} & \textbf{72.93} & 95.82 & 86.68 & \textbf{100.0} & \textbf{98.80} & \textbf{99.49} & \textbf{60.17} & \textbf{90.93} & \textbf{90.04} & \textbf{0.857(1)} & \textbf{1.77(1)} \\
\bottomrule
\end{tabular}

\label{tab:overall_results_new}
\end{table*}

\begin{figure*}[ht]
  \centering
  \includegraphics[width=0.2\linewidth]{ 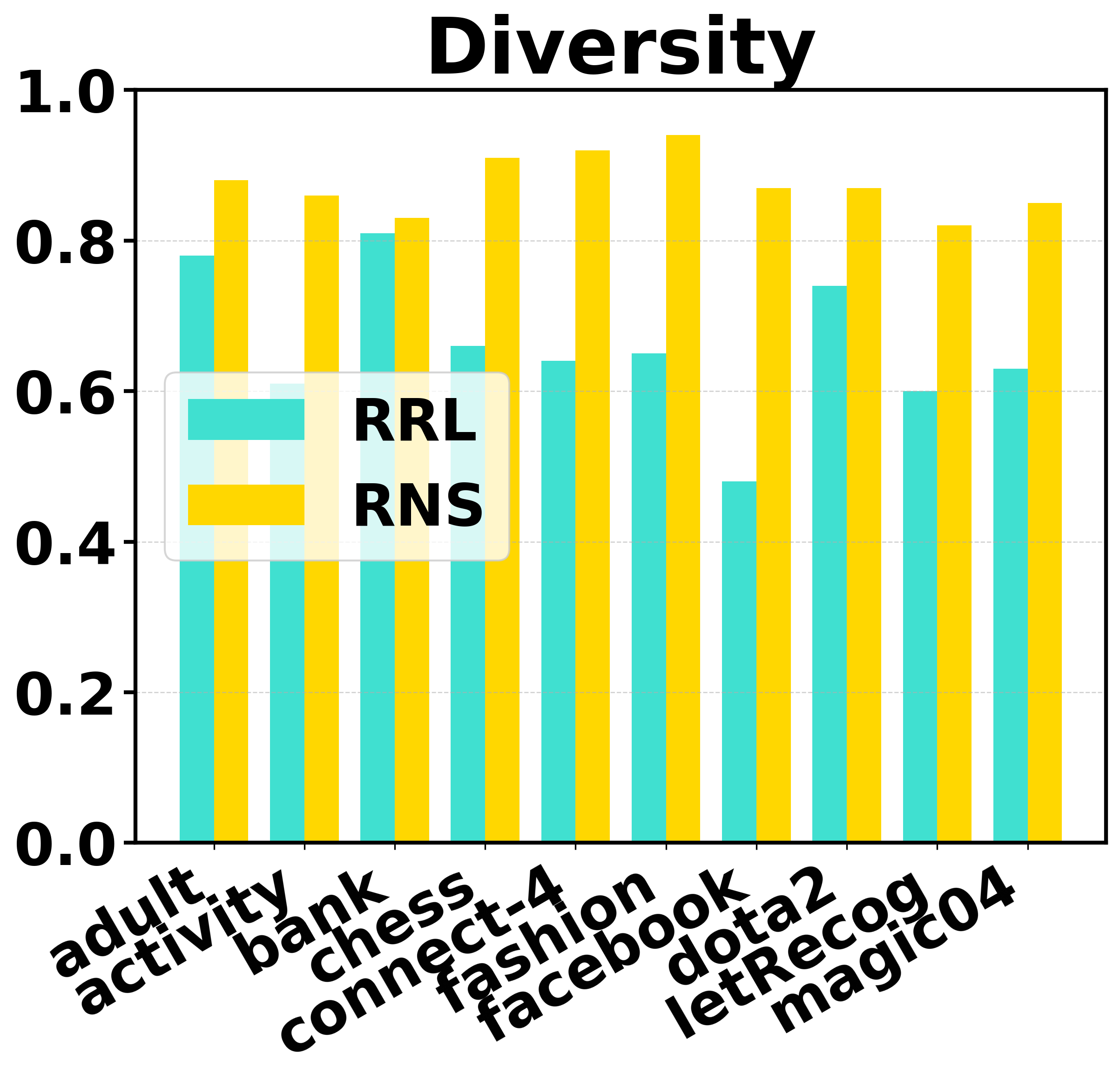}\hfill
  \includegraphics[width=0.2\linewidth]{ 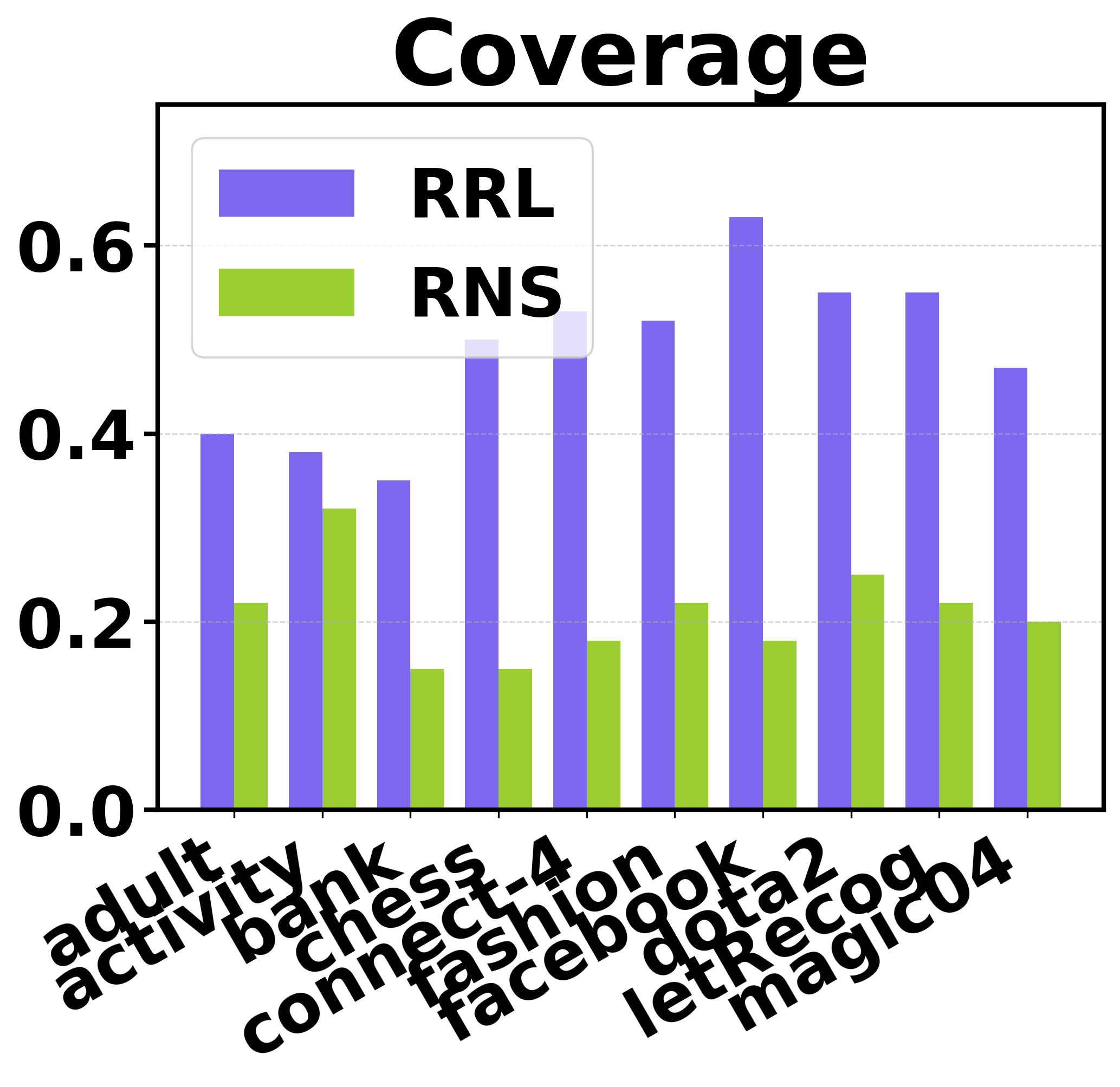}\hfill
  \includegraphics[width=0.2\linewidth]{ 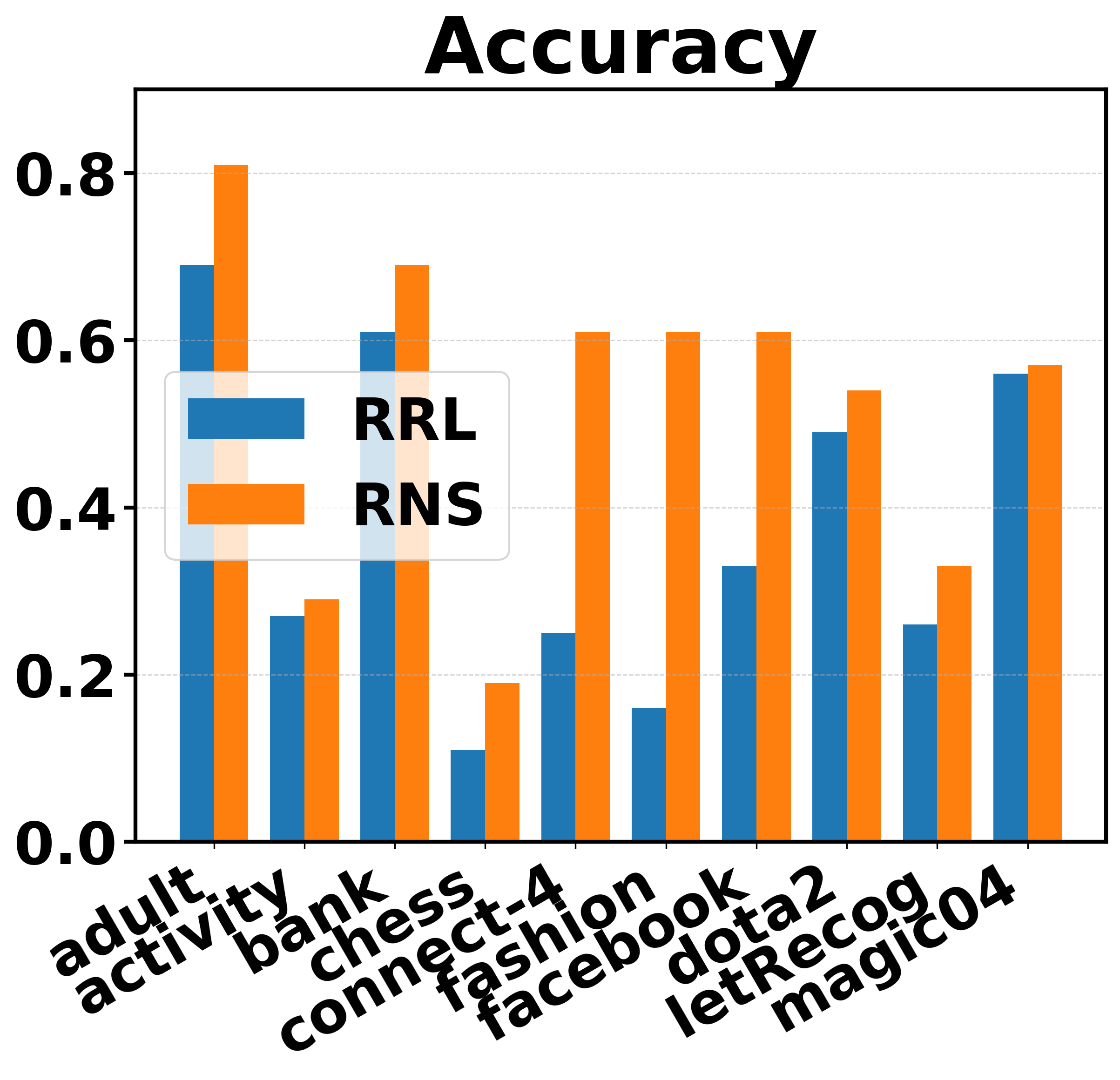}\hfill
  \includegraphics[width=0.2\linewidth]{ 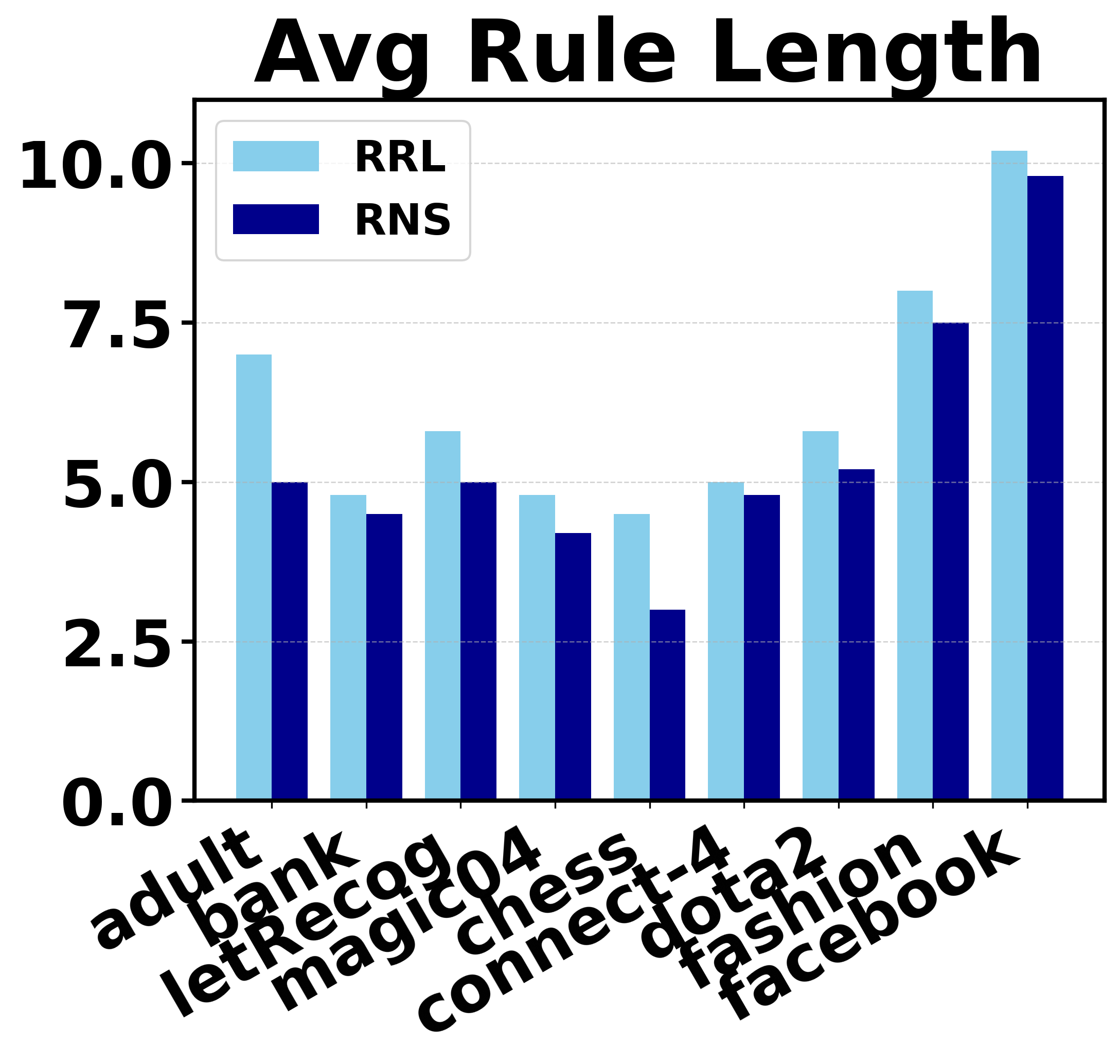}\hfill
  \includegraphics[width=0.2\linewidth]{ 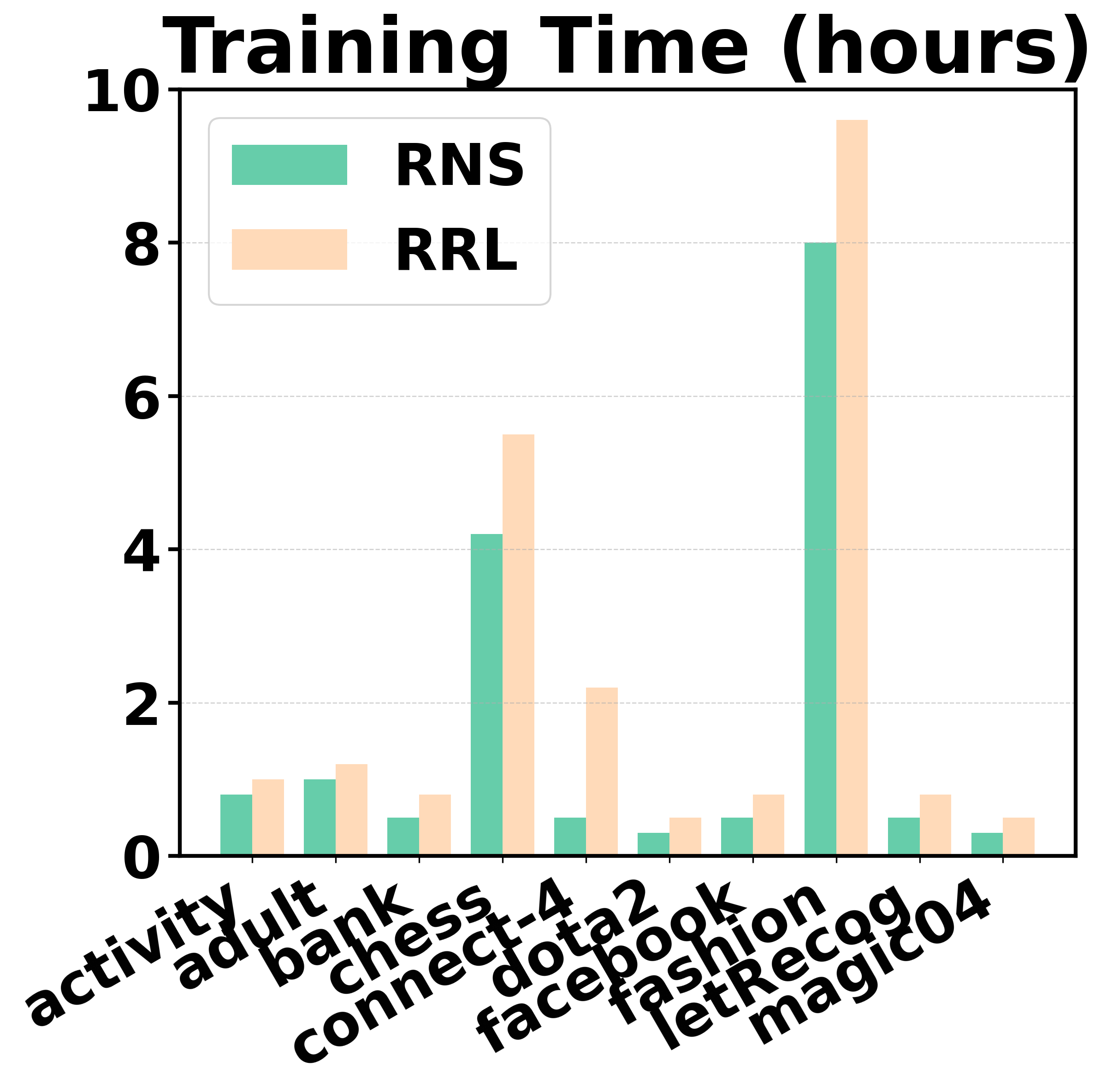}
  \caption{(a) Diversity. (b) Coverage. (c) Accuracy. (d) Avg Rule Length. (e) Training Time.}
  \label{fig:rule_analysis}
\end{figure*}

We conduct comprehensive experiments to evaluate RNS and answer the following research questions: \textbf{RQ1}: Learning high-quality rules is the core goal of an interpretable model. Can RNS learn high-quality rules? \textbf{RQ2}: How does RNS perform w.r.t. classification accuracy compared to SOTA baselines? \textbf{RQ3}: How efficient is RNS in terms of model complexity and training time? \textbf{RQ4}: How do the proposed Negation Layer, HCC, and different binning functions impact RNS? Note that detailed ablation studies for RQ4 are provided in Appendix~\ref{Ablation_HCC}. \textbf{RQ5}: What are the impacts of different hyperparameters in RNS? 

\subsection{Experimental Settings}

\textbf{Datasets.} Extensive experiments are conducted on nine small datasets (adult, bank-marketing, banknote, chess, c-4, letRecog, magic04, wine, tic-tac-toe) and four large datasets (activity, dota2, facebook, fashion-mnist). These datasets are widely used for evaluating classification performance~\citep{Letham_2015,wang2017bayesian,yang2017scalable}. Details are provided in Table~\ref{tab:dataset_stats}. 


\textbf{Performance Evaluation.} We assess performance using F1 score with 5-fold cross-validation. To provide a comprehensive evaluation, we compare models across all datasets using AvgRank~\citep{demvsar2006statistical} and N-Mean~\citep{marton2023grande}.
We evaluate RNS against interpretable and non-interpretable complex methods. The interpretable models include RRL~\citep{wang2021scalable, Wang_2024}, RIPPER~\citep{cohen1995fast}, CRS~\citep{wang2020transparent}, PTURS~\citep{yang2024probabilistictrulyunorderedrule}, DRNET~\citep{qiao2021learningaccurateinterpretabledecision}, LORD~\citep{beck2023layerwise}, C4.5~\citep{quinlan2014c4}, CART~\citep{breiman2017classification}, SBRL~\citep{yang2017scalable}, CORELS~\citep{angelino2018learning}, Logistic Regression (LR)~\citep{kleinbaum2008logistic}, KNN~\citep{peterson2009k}. The non-interpretable models include BNN~\citep{courbariaux2016binarized}, PLNN~\citep{chu2018exact}, SVM~\citep{scholkopf2002learning}, RF~\citep{breiman2001random}, LightGBM (LGBM)~\citep{ke2017lightgbm}, XGBoost (XGB)~\citep{chen2016xgboost}, FT~\citep{gorishniy2021revisiting}, SAINT~\citep{somepalli2021saint}, NODE~\citep{popov2019neural}, STG~\citep{yamada2020feature}, TabNet~\citep{arik2020tabnetattentiveinterpretabletabular}, TabTrans~\citep{huang2020tabtransformer}, VIME~\citep{yoon2020vime}.

\begin{figure}[t]
\centering
\includegraphics[width=0.49\textwidth]{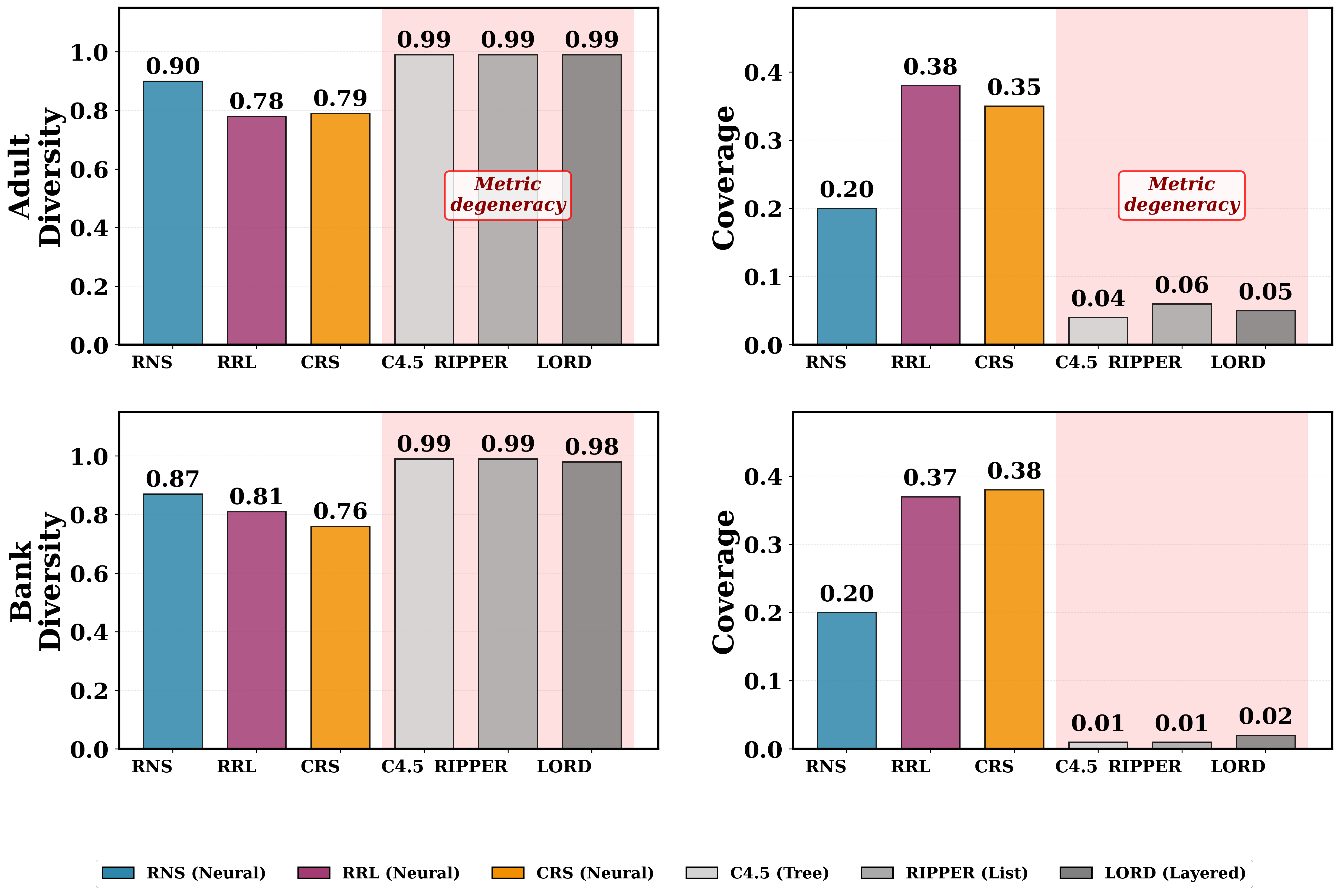}
\caption{Extended rule quality analysis comparing neural and classical baselines.}
\label{fig:extended_rule_quality}
\end{figure}

\subsection{Rule Quality (RQ1)}
The primary advantage of rule learning models over black-box methods is their ability to provide transparent and interpretable insights. However, this advantage is entirely dependent on the quality of the rules they generate. Learned rules must be of high quality, possessing characteristics like simplicity, accuracy, and generalizability. Such rules are crucial for fostering user trust, facilitating debugging, and discovering meaningful knowledge from data. Poorly constructed or overly complex rules defeat the purpose of using an interpretable model in the first place.
In this section, we evaluate the quality of the learned rules. Specifically, we benchmark RNS against RRL, the SOTA rule-based neural network, as a comparative baseline. We also examine classical interpretable methods including C4.5 (decision tree), RIPPER (decision list), and LORD (layered rule learning). The evaluation is conducted in two parts: (1) a comparison of the rules generated by these models using various existing \textbf{rule quality metrics}, and (2) a simulation experiment to assess the models' ability to recover the underlying rules.
\subsubsection{Rule Quality Metrics}
Following prior work~\cite{10.1145/3583780.3614884,lakkaraju2016interpretable}, we evaluate rule quality using three key metrics: diversity, coverage, and single-rule accuracy. Extensive experiments are conducted across 10 datasets. \textbf{Accuracy} measures the prediction accuracy of a single rule for the instances it covers. As shown in Figure~\ref{fig:rule_analysis} and Figure~\ref{fig:accuracy_all}, RNS achieves higher or comparable single-rule accuracy on nearly all datasets. \textbf{Coverage} quantifies the proportion of data instances covered by a rule. Lower coverage indicates that rules are more specific and easier for human experts to understand. As depicted in Figure~\ref{fig:rule_analysis} and Figure~\ref{fig:coverage_all}, RNS consistently yields rules with lower average coverage across most datasets, indicating that its rules focus on more localized, less redundant patterns. \textbf{Diversity} measures the overlap ratio between pairs of rules, with higher diversity reflecting that rules capture distinct, non-redundant logic. In Figure~\ref{fig:rule_analysis} and Figure~\ref{fig:diversity_all}, RNS consistently achieves higher diversity scores across all tested datasets. The diversity gap is especially prominent for more complex datasets such as \textit{facebook} and \textit{dota2}, demonstrating RNS's ability to extract a set of rules that cover a wider variety of patterns with minimal redundancy.

\noindent\textbf{Comparison with Classical Interpretable Methods.}
We extend our analysis to include classical interpretable baselines. As shown in Figure~\ref{fig:extended_rule_quality}, these methods exhibit metric degeneracy. This arises from structural constraints—decision trees partition the input space into disjoint regions, forcing diversity near 1 and coverage near $1/\#\text{leaves}$, while decision lists create similar partitions via first-match semantics. These architecturally determined values provide no discriminative information for comparing rule quality. In contrast, neural-based rule methods learn overlapping rule sets where diversity and coverage reflect meaningful structural differences.

In summary, RNS produces rule sets that are not only more accurate but also more diverse and interpretable due to their reduced coverage and increased diversity. This improvement in rule quality—across accuracy, coverage, and diversity—demonstrates the effectiveness of RNS in generating more meaningful and distinct rules.

\begin{figure}[ht]
    \centering
    \begin{minipage}[t]{0.5\linewidth}
        \includegraphics[width=\linewidth]{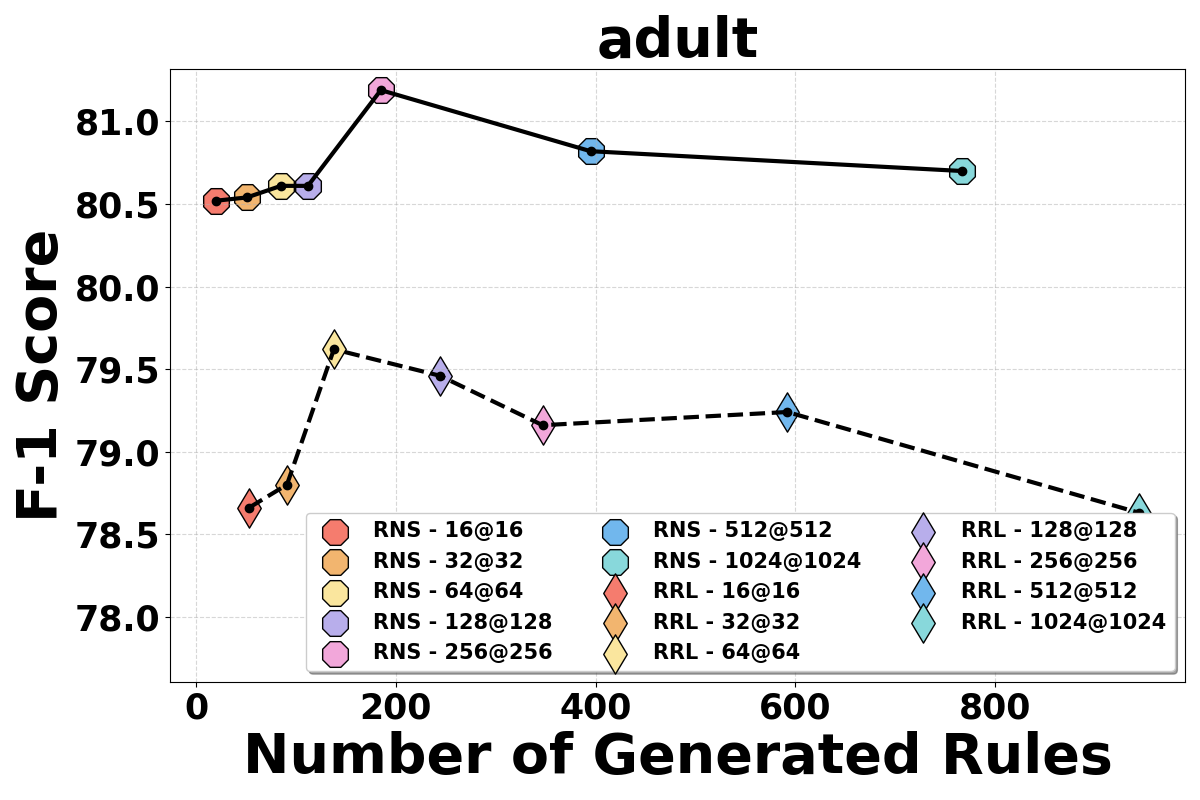}
    \end{minipage}%
    \hfill%
    \begin{minipage}[t]{0.5\linewidth}
        \includegraphics[width=\linewidth]{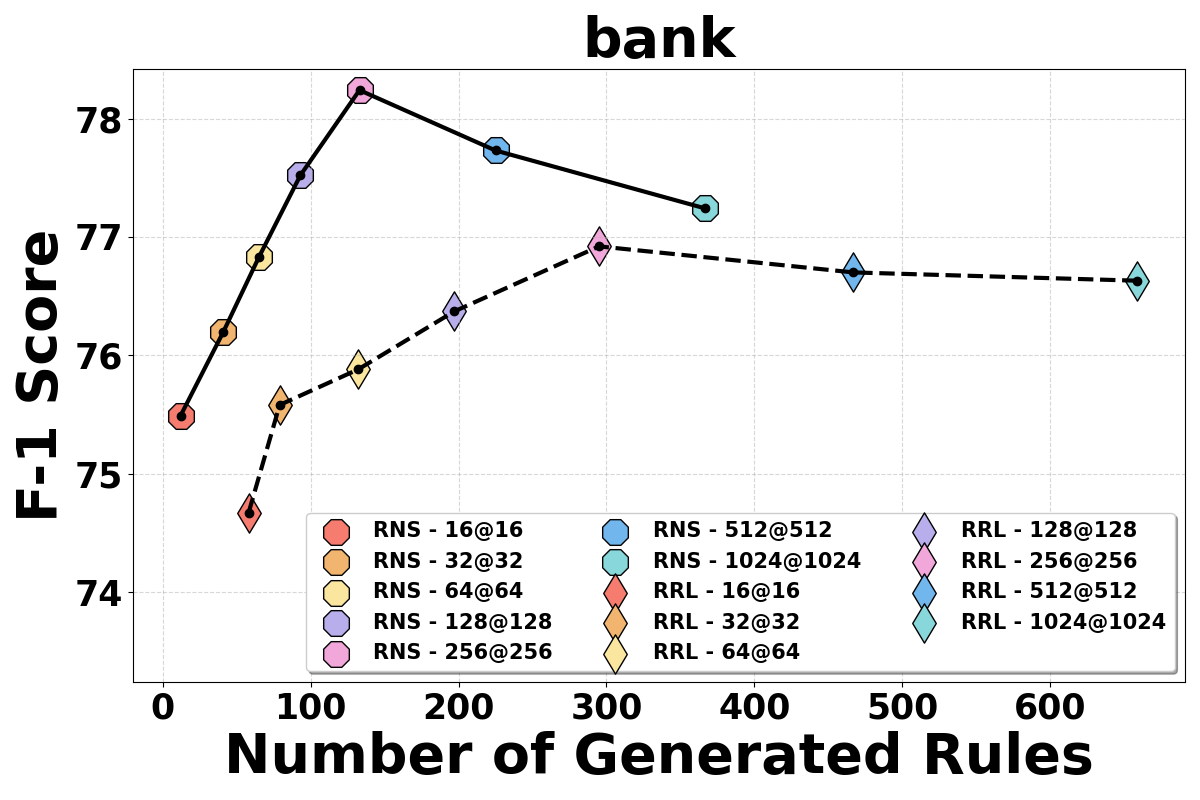}
    \end{minipage}%
\caption{Comparison with Different Numbers of Generated Rules. }

\label{fig:rule_number}
\end{figure}

\subsection{Classification Performance (RQ2)}
Table~\ref{tab:overall_results_new} demonstrates the superior performance of RNS compared to both interpretable and non-interpretable baseline models across 13 datasets. RNS achieves the highest Normalized Mean (N-Mean) score of 0.857, outperforming all baselines, and its average rank of 1.77 further highlights its dominance, as it consistently ranks first or near the top across most datasets. Among interpretable baselines, RRL is the strongest competitor with an N-Mean of 0.851 and average rank of 4.62, yet RNS consistently outperforms it across the majority of datasets—for example, on the bank dataset, RNS achieves 77.62\%, a 0.44\% improvement over RRL's 77.18\%, and on fashion, RNS attains 90.04\%, compared to RRL's 89.64\%. Although complex non-interpretable models such as NODE (N-Mean: 0.850, AvgRank: 5.77), XGBoost (N-Mean: 0.847, AvgRank: 6.38), and LightGBM (N-Mean: 0.845, AvgRank: 7.46) are recognized for their strong performance, RNS surpasses them on several benchmarks, achieving 72.93\% on the c-4 dataset (exceeding LightGBM and XGBoost by 2.40\% and 2.28\%, respectively) and 99.49\% on the activity dataset (outperforming LightGBM at 99.41\% and XGBoost at 99.38\%). Notably, RNS also outperforms LORD, another strong interpretable rule-based baseline (N-Mean: 0.847, AvgRank: 7.00), across most datasets, including achieving superior performance on adult (81.24\% vs. 80.72\%), bank (77.62\% vs. 74.90\%), chess (81.19\% vs. 80.61\%), c-4 (72.93\% vs. 70.77\%), and dota2 (60.17\% vs. 58.61\%). These results highlight that RNS is not only interpretable but also matches or exceeds the performance of the strongest black-box baselines across diverse tabular datasets.

\begin{table*}[t]
\centering
\small
\caption{Logical rules learned by RNS, RRL, and RRL with naive negation on synthetic data. }

\begin{tabular}{|l|l|l|l|}
\hline
\textbf{Ground-Truth Rules} & \textbf{RNS} & \textbf{RRL} & \textbf{RRL w naive negation} \\
\hline
$(x_{1}\vee x_{2})\wedge \neg x_{3}$ 
& $x_{1}\wedge \neg x_{3},\; x_{2}\wedge \neg x_{3}$ 
& $x_{1}\wedge x_{2}\wedge x_{3},\; x_{1},\; x_{2}\wedge x_{3}$ 
& \textbf{$x_{1}\wedge \neg x_{2}\wedge \neg x_{3}$}, $x_{1}$, $\neg x_{3}$ \\
\hline
$x_{1}\vee (\neg x_{2}\wedge \neg x_{3})$ 
& $\neg x_{2}\wedge \neg x_{3},\; x_{1}$ 
& $x_{3}\vee x_{2}\vee (x_{2}\vee x_{1})$ 
& \textbf{$x_{1}$}, $x_{2}$, $x_{1}\wedge \neg x_{2}\wedge x_{3}$ \\
\hline
$x_{1}\wedge \neg x_{2}\wedge x_{3}$ 
& $x_{1}\wedge \neg x_{2}\wedge x_{3}$ 
& $x_{1}\wedge x_{3},\; x_{2},\; (x_{1}\wedge x_{3})\wedge x_{1}$ 
& \textbf{$x_{1}$}, $x_{2}$, $x_{1}\wedge \neg x_{2}\wedge x_{3}$ \\
\hline
$x_{1}\vee \neg x_{2}\vee \neg x_{3}$ 
& $x_{1}\vee \neg x_{2}\vee \neg x_{3}$ 
& $x_{2}\wedge x_{3},\; (x_{2}\wedge x_{3})\wedge x_{2}$ 
& \textbf{$x_{1}\wedge \neg x_{3}$}, $\neg x_{2}\wedge \neg x_{3}$ \\
\hline
\end{tabular}

\label{tab:synthetic}
\end{table*}

\subsubsection{Simulation Experiment}
We conduct a controlled simulation experiment to assess RNS's ability to learn the exact logical rules used to generate synthetic data.

\noindent \textbf{Setup.} We generate synthetic data using predefined logical rules to evaluate rule reconstruction ability. Three probability parameters $\mathbf{p} = (p_1, p_2, p_3)$ correspond to feature variables $\{x_1, x_2, x_3\}$, each drawn from $U(0, 1)$. Using these as Bernoulli parameters, we generate $X_{gen}\in\mathbb{R}^{3\times 50000}$ via Bernoulli sampling, yielding $n=50000$ binary vectors, where each element is sampled from $Bernoulli(p_i)$. Labels are assigned based on logical rules (e.g., $x_1 \land x_2 \land x_3 \rightarrow 1$ assigns label 1 if all features are 1). We define four rule types, as presented in Table~\ref{tab:synthetic}, and split the data evenly into training/test sets. Both RNS and RRL use logical layer dimensions of $64@64$.

\noindent \textbf{Results.} Table~\ref{tab:synthetic} shows RNS achieves near 100\% accuracy while recovering exact ground-truth rule structures. RRL produces overly simplified, redundant, or logically incomplete rules despite using naive negation settings. Detailed and extended analysis is provided in Appendix~\ref{appendix:simu}.

\subsection{Efficiency (RQ3)}
\label{Efficiency all}

We evaluate learning efficiency based on \textbf{the number of learned rules}, \textbf{the length of the learned rules}, and \textbf{computational time}. Large rule sets with lengthy conditions are difficult to interpret, so smaller rule sets with concise rules are preferred. Figure~\ref{fig:rule_number} and Figure~\ref{fig:rule_number_all} illustrate the relationship between the number of learned rules and the F1 score. RNS consistently outperforms RRL, achieving higher performance with fewer rules. Additionally, Figure~\ref{fig:rule_analysis} and Figure~\ref{fig:rule_length_all} show that RNS produces shorter rules, averaging fewer than eight conditions, improving comprehensibility. In terms of computational time, as shown in Figure~\ref{fig:rule_analysis}, RNS converges faster than the best baseline across all datasets. In general, RNS is more efficient than the baseline as it consistently produces fewer rules of a more concise size, using shorter training time.

\subsection{Hyperparameter Study (RQ5)}

\begin{figure}[t]
\centering
\begin{minipage}{0.45\textwidth}
    \includegraphics[width=\linewidth]{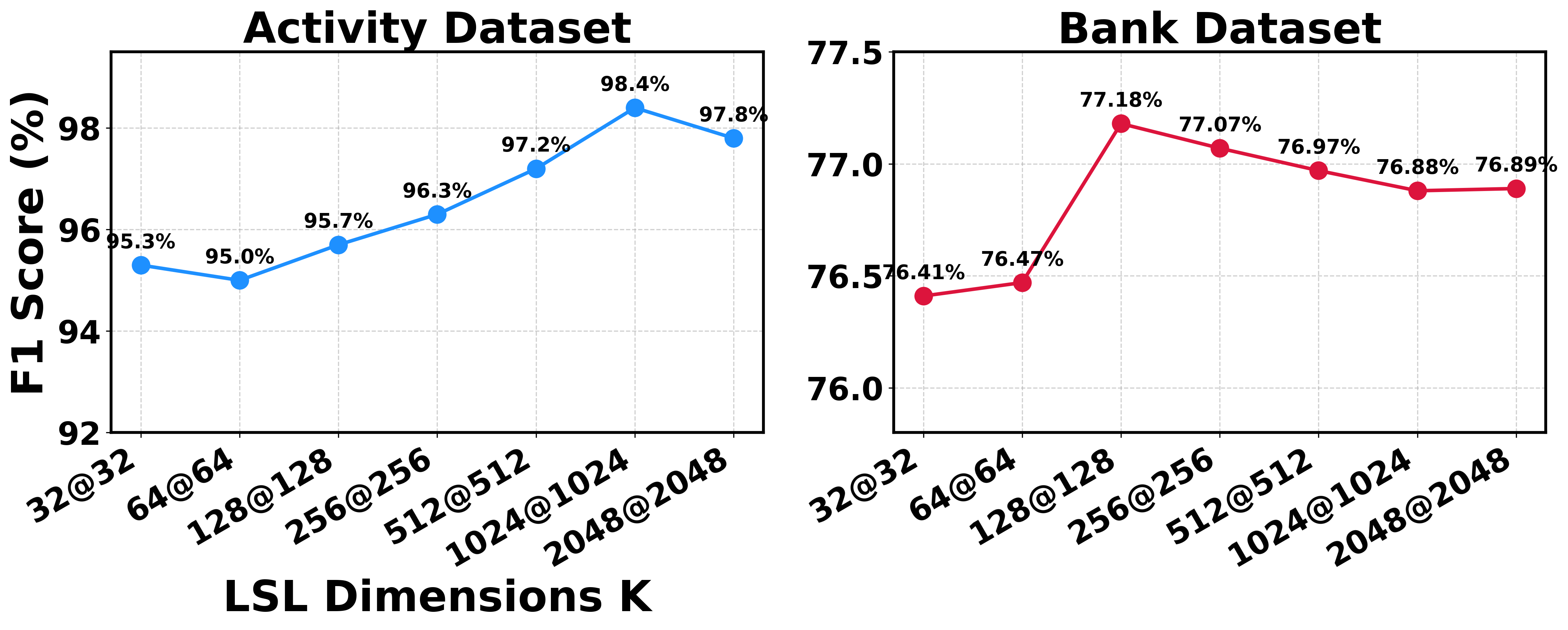}
\end{minipage}\hfill
\begin{minipage}{0.45\textwidth}
    \includegraphics[width=\linewidth]{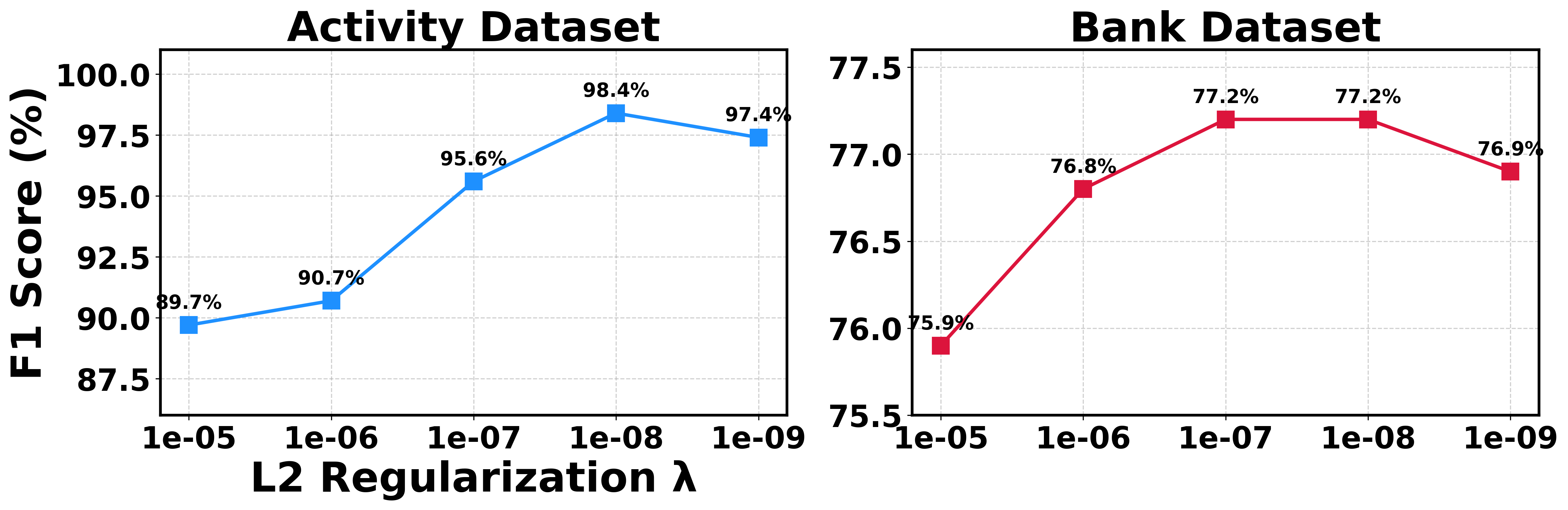}

\end{minipage}\hfill

\caption{Impact of LSL dimension K and L2 regularization.}
\label{fig:K_l2}

\end{figure}

We evaluate LSL dimension $K$ and L2 regularization in Figure~\ref{fig:K_l2}. 
For the \textit{activity} dataset, F1 improves steadily with larger $K$, peaking at $1024@1024$, 
while the \textit{bank} dataset achieves its best performance earlier at $128@128$. 
Similarly, moderate L2 values yield the highest scores, 
with \textit{activity} reaching $98.4\%$ and \textit{bank} stabilizing near $77.2\%$. 
These results suggest that \textit{activity} dataset benefits from larger architectures and balanced regularization, 
whereas \textit{bank} saturates with smaller models and is less sensitive to $\lambda$.

\subsection{Model Interpretation}
RNS provides interpretable rules that capture meaningful patterns across diverse domains. We present qualitative analysis on three datasets in Appendix~\ref{appendix:case_study}. On the Adult dataset, RNS discovers CNF encoding compound economic constraints. On the Wine dataset, RNS captures precise chemical balance requirements. On the bank-marketing dataset (Figure~\ref{fig:case_study}), RNS identifies seasonal effects, the link between longer calls and successful deposits, and demographic tendencies. These examples illustrate how RNS reflects domain knowledge through transparent logical structures.

\section{Advantages and Limitations of RNS}
\textbf{Advantages.} RNS advances interpretable rule learning via (i) learnable AND/OR operator selection per neuron, (ii) functional completeness with NOT via a Negation Layer, and (iii) robust optimization using max/min logical activations, $\pm1$ states, and STE. These yield superior prediction, rule quality, and efficiency over prior work (e.g., RRL).

\textbf{Limitations.} Although more efficient than prior neural rule learners, RNS is heavier than heuristic algorithms (e.g., C4.5, RIPPER). Future work includes improved scalability for extremely high-dimensional or streaming data and integrating domain priors.
\section{Conclusion}
We proposed RNS, a selective discrete neural network that learns CNF/DNF rules via LSLs with learnable AND/OR neurons, a Negation Layer, and an HCC for valid, efficient connections. With STE-enabled training and logical max/min activations, RNS achieves strong performance, rule quality, and efficiency across diverse datasets. Future directions include recommendations and text applications.

\section*{Impact Statement}

This paper presents work whose goal is to advance the field of interpretable machine learning through the development of Rule Network with Selective Logical Operators (RNS). We believe our work has several positive societal implications worth highlighting.

\textbf{Transparency and Trust in AI Systems.} By producing human-readable logical rules, RNS enables stakeholders to understand and verify model decisions. This transparency is particularly valuable in high-stakes domains such as healthcare, finance, and criminal justice, where algorithmic decisions significantly impact human lives. Unlike black-box models, RNS allows domain experts to audit predictions, identify potential biases, and ensure compliance with regulatory requirements.

\textbf{Democratization of AI Understanding.} The interpretable nature of rule-based outputs makes AI systems more accessible to non-technical stakeholders, including patients, loan applicants, and policymakers. This democratization can foster more informed public discourse about AI deployment and governance.

\textbf{Potential Risks and Mitigations.} While interpretability generally enhances accountability, we acknowledge that rule-based explanations could potentially be misused to game systems (e.g., manipulating features to achieve desired outcomes) or to provide oversimplified justifications for complex decisions. We encourage practitioners to complement RNS with appropriate safeguards and to avoid over-reliance on any single interpretable model for critical decisions.

\textbf{Environmental Considerations.} RNS demonstrates improved training efficiency compared to existing neural rule learning methods, potentially reducing the computational resources and energy consumption required for model development.

Overall, we believe the benefits of advancing interpretable machine learning substantially outweigh the risks, and we encourage responsible deployment of such systems with appropriate human oversight.

\bibliographystyle{unsrtnat}
\bibliography{main}  

\begin{thebibliography}{70}
\providecommand{\natexlab}[1]{#1}
\providecommand{\url}[1]{\texttt{#1}}
\expandafter\ifx\csname urlstyle\endcsname\relax
  \providecommand{\doi}[1]{doi: #1}\else
  \providecommand{\doi}{doi: \begingroup \urlstyle{rm}\Url}\fi

\bibitem[Rudin(2019)]{rudin2019stopexplainingblackbox}
Cynthia Rudin.
\newblock Stop explaining black box machine learning models for high stakes decisions and use interpretable models instead, 2019.
\newblock URL \url{https://arxiv.org/abs/1811.10154}.

\bibitem[Molnar(2020)]{molnar2020interpretable}
Christoph Molnar.
\newblock \emph{Interpretable machine learning}.
\newblock Lulu. com, 2020.

\bibitem[Yin and Han(2003)]{yin2003cpar}
Xiaoxin Yin and Jiawei Han.
\newblock Cpar: Classification based on predictive association rules.
\newblock In \emph{Proceedings of the 2003 SIAM international conference on data mining}, pages 331--335. SIAM, 2003.

\bibitem[Frank and Witten(1998)]{10.5555/645527.657305}
Eibe Frank and Ian~H. Witten.
\newblock Generating accurate rule sets without global optimization.
\newblock In \emph{Proceedings of the Fifteenth International Conference on Machine Learning}, ICML '98, page 144–151, San Francisco, CA, USA, 1998. Morgan Kaufmann Publishers Inc.
\newblock ISBN 1558605568.

\bibitem[Cohen(1995)]{cohen1995fast}
William~W Cohen.
\newblock Fast effective rule induction.
\newblock In \emph{Machine learning proceedings 1995}, pages 115--123. Elsevier, 1995.

\bibitem[Quinlan(2014)]{quinlan2014c4}
J~Ross Quinlan.
\newblock \emph{C4. 5: programs for machine learning}.
\newblock Elsevier, 2014.

\bibitem[Yang et~al.(2017)Yang, Rudin, and Seltzer]{yang2017scalable}
Hongyu Yang, Cynthia Rudin, and Margo Seltzer.
\newblock Scalable bayesian rule lists.
\newblock In \emph{International conference on machine learning}, pages 3921--3930. PMLR, 2017.

\bibitem[Marton et~al.(2023)Marton, L{\"u}dtke, Bartelt, and Stuckenschmidt]{marton2023grande}
Sascha Marton, Stefan L{\"u}dtke, Christian Bartelt, and Heiner Stuckenschmidt.
\newblock Grande: Gradient-based decision tree ensembles.
\newblock \emph{arXiv preprint arXiv:2309.17130}, 2023.

\bibitem[Michalski(1973)]{Michalski1973AQVAL1ComputerIO}
Ryszard~S. Michalski.
\newblock Aqval/1--computer implementation of a variable-valued logic system vl1 and examples of its application to pattern recognition.
\newblock In \emph{International Joint Conference on Artificial Intelligence}, 1973.
\newblock URL \url{https://api.semanticscholar.org/CorpusID:60492559}.

\bibitem[Quinlan(1990)]{quinlan1990learning}
J.~Ross Quinlan.
\newblock Learning logical definitions from relations.
\newblock \emph{Machine learning}, 5\penalty0 (3):\penalty0 239--266, 1990.

\bibitem[Clark and Niblett(1989)]{10.1023/A:1022641700528}
Peter Clark and Tim Niblett.
\newblock The cn2 induction algorithm.
\newblock \emph{Mach. Learn.}, 3\penalty0 (4):\penalty0 261–283, mar 1989.
\newblock ISSN 0885-6125.
\newblock \doi{10.1023/A:1022641700528}.
\newblock URL \url{https://doi.org/10.1023/A:1022641700528}.

\bibitem[Mooney(1995)]{mooney1995encouraging}
Raymond~J Mooney.
\newblock Encouraging experimental results on learning cnf.
\newblock \emph{Machine Learning}, 19:\penalty0 79--92, 1995.

\bibitem[Beck et~al.(2023)Beck, F{\"u}rnkranz, and Huynh]{beck2023layerwise}
Florian Beck, Johannes F{\"u}rnkranz, and Van Quoc~Phuong Huynh.
\newblock Layerwise learning of mixed conjunctive and disjunctive rule sets.
\newblock In \emph{International Joint Conference on Rules and Reasoning}, pages 95--109. Springer, 2023.

\bibitem[Ke et~al.(2017)Ke, Meng, Finley, Wang, Chen, Ma, Ye, and Liu]{ke2017lightgbm}
Guolin Ke, Qi~Meng, Thomas Finley, Taifeng Wang, Wei Chen, Weidong Ma, Qiwei Ye, and Tie-Yan Liu.
\newblock Lightgbm: A highly efficient gradient boosting decision tree.
\newblock \emph{Advances in neural information processing systems}, 30, 2017.

\bibitem[Breiman(2001)]{breiman2001random}
Leo Breiman.
\newblock Random forests.
\newblock \emph{Machine learning}, 45:\penalty0 5--32, 2001.

\bibitem[Letham et~al.(2015)Letham, Rudin, McCormick, and Madigan]{Letham_2015}
Benjamin Letham, Cynthia Rudin, Tyler~H. McCormick, and David Madigan.
\newblock Interpretable classifiers using rules and bayesian analysis: Building a better stroke prediction model.
\newblock \emph{The Annals of Applied Statistics}, 9\penalty0 (3), September 2015.
\newblock ISSN 1932-6157.
\newblock \doi{10.1214/15-aoas848}.
\newblock URL \url{http://dx.doi.org/10.1214/15-AOAS848}.

\bibitem[Wang et~al.(2017)Wang, Rudin, Doshi-Velez, Liu, Klampfl, and MacNeille]{wang2017bayesian}
Tong Wang, Cynthia Rudin, Finale Doshi-Velez, Yimin Liu, Erica Klampfl, and Perry MacNeille.
\newblock A bayesian framework for learning rule sets for interpretable classification.
\newblock \emph{The Journal of Machine Learning Research}, 18\penalty0 (1):\penalty0 2357--2393, 2017.

\bibitem[Loh(2011)]{loh2011classification}
Wei-Yin Loh.
\newblock Classification and regression trees.
\newblock \emph{Wiley interdisciplinary reviews: data mining and knowledge discovery}, 1\penalty0 (1):\penalty0 14--23, 2011.

\bibitem[Wang et~al.(2020)Wang, Zhang, Liu, and Wang]{wang2020transparent}
Zhuo Wang, Wei Zhang, Ning Liu, and Jianyong Wang.
\newblock Transparent classification with multilayer logical perceptrons and random binarization, 2020.

\bibitem[Wang et~al.(2021)Wang, Zhang, Liu, and Wang]{wang2021scalable}
Zhuo Wang, Wei Zhang, Ning Liu, and Jianyong Wang.
\newblock Scalable rule-based representation learning for interpretable classification, 2021.

\bibitem[Wang et~al.(2024)Wang, Zhang, Liu, and Wang]{Wang_2024}
Zhuo Wang, Wei Zhang, Ning Liu, and Jianyong Wang.
\newblock Learning interpretable rules for scalable data representation and classification.
\newblock \emph{IEEE Transactions on Pattern Analysis and Machine Intelligence}, 46\penalty0 (2):\penalty0 1121–1133, February 2024.
\newblock ISSN 1939-3539.
\newblock \doi{10.1109/tpami.2023.3328881}.
\newblock URL \url{http://dx.doi.org/10.1109/TPAMI.2023.3328881}.

\bibitem[Yu et~al.(2023)Yu, Li, Zhang, Li, and Zhou]{10.1145/3583780.3614884}
Lu~Yu, Meng Li, Ya-Lin Zhang, Longfei Li, and Jun Zhou.
\newblock Finrule: Feature interactive neural rule learning.
\newblock In \emph{Proceedings of the 32nd ACM International Conference on Information and Knowledge Management}, CIKM '23, page 3020–3029, New York, NY, USA, 2023. Association for Computing Machinery.
\newblock ISBN 9798400701245.
\newblock \doi{10.1145/3583780.3614884}.
\newblock URL \url{https://doi.org/10.1145/3583780.3614884}.

\bibitem[Zhang et~al.(2023)Zhang, Liu, Wang, and Wang]{zhang2023learning}
Wei Zhang, Yongxiang Liu, Zhuo Wang, and Jianyong Wang.
\newblock Learning to binarize continuous features for neuro-rule networks.
\newblock In \emph{Proceedings of the Thirty-Second International Joint Conference on Artificial Intelligence}, pages 4584--4592, 2023.

\bibitem[Dougherty et~al.(1995)Dougherty, Kohavi, and Sahami]{dougherty1995supervised}
James Dougherty, Ron Kohavi, and Mehran Sahami.
\newblock Supervised and unsupervised discretization of continuous features.
\newblock In \emph{Machine learning proceedings 1995}, pages 194--202. Elsevier, 1995.

\bibitem[Mendelson(1997)]{mendelson1997schaum}
Elliott Mendelson.
\newblock \emph{Schaum's outline of theory and problems of beginning calculus}.
\newblock McGraw-Hill, 1997.

\bibitem[Enderton(2001)]{enderton2001mathematical}
Herbert~B Enderton.
\newblock \emph{A mathematical introduction to logic}.
\newblock Elsevier, 2001.

\bibitem[Courbariaux et~al.(2015)Courbariaux, Bengio, and David]{courbariaux2015binaryconnect}
Matthieu Courbariaux, Yoshua Bengio, and Jean-Pierre David.
\newblock Binaryconnect: Training deep neural networks with binary weights during propagations.
\newblock \emph{Advances in neural information processing systems}, 28, 2015.

\bibitem[Lowe et~al.(2022)Lowe, Earle, d'Eon, Trappenberg, and Oore]{lowe2022logical}
Scott~C. Lowe, Robert Earle, Jason d'Eon, Thomas Trappenberg, and Sageev Oore.
\newblock Logical activation functions: Logit-space equivalents of probabilistic boolean operators, 2022.

\bibitem[Dem{\v{s}}ar(2006)]{demvsar2006statistical}
Janez Dem{\v{s}}ar.
\newblock Statistical comparisons of classifiers over multiple data sets.
\newblock \emph{The Journal of Machine learning research}, 7:\penalty0 1--30, 2006.

\bibitem[Yang and van Leeuwen(2024)]{yang2024probabilistictrulyunorderedrule}
Lincen Yang and Matthijs van Leeuwen.
\newblock Probabilistic truly unordered rule sets, 2024.
\newblock URL \url{https://arxiv.org/abs/2401.09918}.

\bibitem[Qiao et~al.(2021)Qiao, Wang, and Lin]{qiao2021learningaccurateinterpretabledecision}
Litao Qiao, Weijia Wang, and Bill Lin.
\newblock Learning accurate and interpretable decision rule sets from neural networks, 2021.
\newblock URL \url{https://arxiv.org/abs/2103.02826}.

\bibitem[Breiman(2017)]{breiman2017classification}
Leo Breiman.
\newblock \emph{Classification and regression trees}.
\newblock Routledge, 2017.

\bibitem[Angelino et~al.(2018)Angelino, Larus-Stone, Alabi, Seltzer, and Rudin]{angelino2018learning}
Elaine Angelino, Nicholas Larus-Stone, Daniel Alabi, Margo Seltzer, and Cynthia Rudin.
\newblock Learning certifiably optimal rule lists for categorical data.
\newblock \emph{Journal of Machine Learning Research}, 18\penalty0 (234):\penalty0 1--78, 2018.

\bibitem[Kleinbaum et~al.(2008)Kleinbaum, Dietz, Gail, Klein, and Klein]{kleinbaum2008logistic}
David~G Kleinbaum, K~Dietz, M~Gail, M~Klein, and M~Klein.
\newblock Logistic regression.
\newblock \emph{A Self-Learning Tekst}, 2008.

\bibitem[Peterson(2009)]{peterson2009k}
Leif~E Peterson.
\newblock K-nearest neighbor.
\newblock \emph{Scholarpedia}, 4\penalty0 (2):\penalty0 1883, 2009.

\bibitem[Courbariaux et~al.(2016{\natexlab{a}})Courbariaux, Hubara, Soudry, El-Yaniv, and Bengio]{courbariaux2016binarized}
Matthieu Courbariaux, Itay Hubara, Daniel Soudry, Ran El-Yaniv, and Yoshua Bengio.
\newblock Binarized neural networks: Training deep neural networks with weights and activations constrained to +1 or -1, 2016{\natexlab{a}}.

\bibitem[Chu et~al.(2018)Chu, Hu, Hu, Wang, and Pei]{chu2018exact}
Lingyang Chu, Xia Hu, Juhua Hu, Lanjun Wang, and Jian Pei.
\newblock Exact and consistent interpretation for piecewise linear neural networks: A closed form solution.
\newblock In \emph{Proceedings of the 24th ACM SIGKDD International Conference on Knowledge Discovery \& Data Mining}, pages 1244--1253, 2018.

\bibitem[Sch{\"o}lkopf and Smola(2002)]{scholkopf2002learning}
Bernhard Sch{\"o}lkopf and Alexander~J Smola.
\newblock \emph{Learning with kernels: support vector machines, regularization, optimization, and beyond}.
\newblock MIT press, 2002.

\bibitem[Chen and Guestrin(2016)]{chen2016xgboost}
Tianqi Chen and Carlos Guestrin.
\newblock Xgboost: A scalable tree boosting system.
\newblock In \emph{Proceedings of the 22nd acm sigkdd international conference on knowledge discovery and data mining}, pages 785--794, 2016.

\bibitem[Gorishniy et~al.(2021)Gorishniy, Rubachev, Khrulkov, and Babenko]{gorishniy2021revisiting}
Yury Gorishniy, Ivan Rubachev, Valentin Khrulkov, and Artem Babenko.
\newblock Revisiting deep learning models for tabular data.
\newblock \emph{Advances in Neural Information Processing Systems}, 34:\penalty0 18932--18943, 2021.

\bibitem[Somepalli et~al.(2021)Somepalli, Goldblum, Schwarzschild, Bruss, and Goldstein]{somepalli2021saint}
Gowthami Somepalli, Micah Goldblum, Avi Schwarzschild, C~Bayan Bruss, and Tom Goldstein.
\newblock Saint: Improved neural networks for tabular data via row attention and contrastive pre-training.
\newblock \emph{arXiv preprint arXiv:2106.01342}, 2021.

\bibitem[Popov et~al.(2019)Popov, Morozov, and Babenko]{popov2019neural}
Sergei Popov, Stanislav Morozov, and Artem Babenko.
\newblock Neural oblivious decision trees for deep learning on tabular data.
\newblock In \emph{Advances in Neural Information Processing Systems}, pages 6285--6295, 2019.

\bibitem[Yamada et~al.(2020)Yamada, Lindenbaum, Negahban, and Kluger]{yamada2020feature}
Yutaro Yamada, Ofir Lindenbaum, Sahand Negahban, and Yuval Kluger.
\newblock Feature selection using stochastic gates.
\newblock In \emph{International Conference on Machine Learning}, pages 10648--10659. PMLR, 2020.

\bibitem[Arik and Pfister(2020)]{arik2020tabnetattentiveinterpretabletabular}
Sercan~O. Arik and Tomas Pfister.
\newblock Tabnet: Attentive interpretable tabular learning, 2020.
\newblock URL \url{https://arxiv.org/abs/1908.07442}.

\bibitem[Huang et~al.(2020)Huang, Khetan, Cvitkovic, and Karnin]{huang2020tabtransformer}
Xin Huang, Ashish Khetan, Milan Cvitkovic, and Zohar Karnin.
\newblock Tabtransformer: Tabular data modeling using contextual embeddings.
\newblock In \emph{arXiv preprint arXiv:2012.06678}, 2020.

\bibitem[Yoon et~al.(2020)Yoon, Zhang, Jordon, and van~der Schaar]{yoon2020vime}
Jinsung Yoon, Yao Zhang, James Jordon, and Mihaela van~der Schaar.
\newblock Vime: Extending the success of self-and semi-supervised learning to tabular domain.
\newblock In \emph{Advances in Neural Information Processing Systems}, volume~33, pages 11033--11043, 2020.

\bibitem[Lakkaraju et~al.(2016)Lakkaraju, Bach, and Leskovec]{lakkaraju2016interpretable}
Himabindu Lakkaraju, Stephen~H Bach, and Jure Leskovec.
\newblock Interpretable decision sets: A joint framework for description and prediction.
\newblock In \emph{Proceedings of the 22nd ACM SIGKDD international conference on knowledge discovery and data mining}, pages 1675--1684, 2016.

\bibitem[S.(1969)]{1571135649817801472}
MICHALSKI~R. S.
\newblock On the quasi-minimal solution of the general covering problem.
\newblock \emph{Proceedings of the 5th International Symposium on Information Processing}, 3:\penalty0 125--128, 1969.
\newblock URL \url{https://cir.nii.ac.jp/crid/1571135649817801472}.

\bibitem[Pagallo and Haussler(1990)]{Pagallo1990BooleanFD}
Giulia Pagallo and David Haussler.
\newblock Boolean feature discovery in empirical learning.
\newblock \emph{Machine Learning}, 5:\penalty0 71--99, 1990.
\newblock URL \url{https://api.semanticscholar.org/CorpusID:5661437}.

\bibitem[Liu et~al.(1998)Liu, Hsu, and Ma]{liu1998integrating}
Bing Liu, Wynne Hsu, and Yiming Ma.
\newblock Integrating classification and association rule mining.
\newblock In \emph{Proceedings of the fourth international conference on knowledge discovery and data mining}, pages 80--86, 1998.

\bibitem[Li et~al.(2001)Li, Han, and Pei]{989541}
Wenmin Li, Jiawei Han, and Jian Pei.
\newblock Cmar: accurate and efficient classification based on multiple class-association rules.
\newblock In \emph{Proceedings 2001 IEEE International Conference on Data Mining}, pages 369--376, 2001.
\newblock \doi{10.1109/ICDM.2001.989541}.

\bibitem[Dries et~al.(2009)Dries, De~Raedt, and Nijssen]{dries2009mining}
Anton Dries, Luc De~Raedt, and Siegfried Nijssen.
\newblock Mining predictive k-cnf expressions.
\newblock \emph{IEEE Transactions on Knowledge and Data Engineering}, 22\penalty0 (5):\penalty0 743--748, 2009.

\bibitem[Jain et~al.(2021)Jain, Gautrais, Kimmig, and De~Raedt]{jain2021learning}
Arcchit Jain, Cl{\'e}ment Gautrais, Angelika Kimmig, and Luc De~Raedt.
\newblock Learning cnf theories using mdl and predicate invention.
\newblock In \emph{IJCAI}, pages 2599--2605, 2021.

\bibitem[Sverdlik(1992)]{sverdlik1992dynamic}
William Sverdlik.
\newblock \emph{Dynamic version spaces in machine learning}.
\newblock Wayne State University, 1992.

\bibitem[Hong and Tsang(1997)]{hong1997generalized}
Tzung-Pai Hong and Shian-Shyong Tsang.
\newblock A generalized version space learning algorithm for noisy and uncertain data.
\newblock \emph{IEEE Transactions on Knowledge and Data Engineering}, 9\penalty0 (2):\penalty0 336--340, 1997.

\bibitem[Frosst and Hinton(2017)]{frosst2017distillingneuralnetworksoft}
Nicholas Frosst and Geoffrey Hinton.
\newblock Distilling a neural network into a soft decision tree, 2017.
\newblock URL \url{https://arxiv.org/abs/1711.09784}.

\bibitem[Ribeiro et~al.(2016)Ribeiro, Singh, and Guestrin]{ribeiro2016should}
Marco~Tulio Ribeiro, Sameer Singh, and Carlos Guestrin.
\newblock " why should i trust you?" explaining the predictions of any classifier.
\newblock In \emph{Proceedings of the 22nd ACM SIGKDD international conference on knowledge discovery and data mining}, pages 1135--1144, 2016.

\bibitem[Petersen et~al.(2022)Petersen, Borgelt, Kuehne, and Deussen]{petersen2022deep}
Felix Petersen, Christian Borgelt, Hilde Kuehne, and Oliver Deussen.
\newblock Deep differentiable logic gate networks.
\newblock In \emph{Advances in Neural Information Processing Systems (NeurIPS)}, volume~35, pages 2006--2019, 2022.

\bibitem[Courbariaux et~al.(2016{\natexlab{b}})Courbariaux, Bengio, and David]{courbariaux2016binaryconnect}
Matthieu Courbariaux, Yoshua Bengio, and Jean-Pierre David.
\newblock Binaryconnect: Training deep neural networks with binary weights during propagations, 2016{\natexlab{b}}.

\bibitem[Rastegari et~al.(2016)Rastegari, Ordonez, Redmon, and Farhadi]{rastegari2016xnornet}
Mohammad Rastegari, Vicente Ordonez, Joseph Redmon, and Ali Farhadi.
\newblock Xnor-net: Imagenet classification using binary convolutional neural networks, 2016.

\bibitem[Bulat and Tzimiropoulos(2019)]{bulat2019xnornet}
Adrian Bulat and Georgios Tzimiropoulos.
\newblock Xnor-net++: Improved binary neural networks, 2019.

\bibitem[Liu et~al.(2018)Liu, Wu, Luo, Yang, Liu, and Cheng]{liu2018bireal}
Zechun Liu, Baoyuan Wu, Wenhan Luo, Xin Yang, Wei Liu, and Kwang-Ting Cheng.
\newblock Bi-real net: Enhancing the performance of 1-bit cnns with improved representational capability and advanced training algorithm, 2018.

\bibitem[Kim and Smaragdis(2016)]{kim2016bitwise}
Minje Kim and Paris Smaragdis.
\newblock Bitwise neural networks, 2016.

\bibitem[Lahoud et~al.(2019)Lahoud, Achanta, Márquez-Neila, and Süsstrunk]{lahoud2019selfbinarizing}
Fayez Lahoud, Radhakrishna Achanta, Pablo Márquez-Neila, and Sabine Süsstrunk.
\newblock Self-binarizing networks, 2019.

\bibitem[Sakr et~al.(2018)Sakr, Choi, Wang, Gopalakrishnan, and Shanbhag]{sakr2018true}
Charbel Sakr, Jungwook Choi, Zhuo Wang, Kailash Gopalakrishnan, and Naresh Shanbhag.
\newblock True gradient-based training of deep binary activated neural networks via continuous binarization.
\newblock In \emph{2018 IEEE international conference on acoustics, speech and signal processing (ICASSP)}, pages 2346--2350. IEEE, 2018.

\bibitem[Cheng et~al.(2019)Cheng, Liu, Li, Shen, Henao, and Carin]{cheng2019straightthrough}
Pengyu Cheng, Chang Liu, Chunyuan Li, Dinghan Shen, Ricardo Henao, and Lawrence Carin.
\newblock Straight-through estimator as projected wasserstein gradient flow, 2019.

\bibitem[Yin et~al.(2019)Yin, Lyu, Zhang, Osher, Qi, and Xin]{yin2019understanding}
Penghang Yin, Jiancheng Lyu, Shuai Zhang, Stanley Osher, Yingyong Qi, and Jack Xin.
\newblock Understanding straight-through estimator in training activation quantized neural nets, 2019.

\bibitem[Kingma and Ba(2014)]{kingma2014adam}
Diederik~P Kingma and Jimmy Ba.
\newblock Adam: A method for stochastic optimization.
\newblock \emph{arXiv preprint arXiv:1412.6980}, 2014.

\bibitem[Paszke et~al.(2019)Paszke, Gross, Massa, Lerer, Bradbury, Chanan, Killeen, Lin, Gimelshein, Antiga, et~al.]{paszke2019pytorch}
Adam Paszke, Sam Gross, Francisco Massa, Adam Lerer, James Bradbury, Gregory Chanan, Trevor Killeen, Zeming Lin, Natalia Gimelshein, Luca Antiga, et~al.
\newblock Pytorch: An imperative style, high-performance deep learning library.
\newblock \emph{Advances in neural information processing systems}, 32, 2019.

\bibitem[Payani and Fekri(2019)]{payani2019learning}
Ali Payani and Faramarz Fekri.
\newblock Learning algorithms via neural logic networks.
\newblock \emph{arXiv preprint arXiv:1904.01554}, 2019.

\end{thebibliography}

\clearpage
\appendix

\section{Related Work}

\subsection{Traditional Rule Learning Methods} 
Historically, example-based rule learning algorithms~\citep{1571135649817801472, Michalski1973AQVAL1ComputerIO, Pagallo1990BooleanFD} were initially proposed by selecting a random example and finding the best rule to cover it. However, due to their computational inefficiency, CN2~\citep{10.1023/A:1022641700528} explicitly changed the strategy to finding the best rule that covers as many examples as possible. Building on these heuristic algorithms, FOIL~\citep{quinlan1990learning}, a system for learning first-order Horn clauses, was subsequently developed. While some algorithms learn rule sets directly, such as RIPPER~\citep{cohen1995fast}, PART~\citep{10.5555/645527.657305}, and CPAR~\citep{yin2003cpar}, others post-process a decision tree~\citep{quinlan2014c4} or construct sets of rules by post-processing association rules, like CBA~\citep{liu1998integrating} and CMAR~\citep{989541}. All these algorithms use different strategies to find and use sets of rules for classification.

Most of these algorithms are based on Disjunctive Normal Form (DNF)~\citep{1571135649817801472, Michalski1973AQVAL1ComputerIO, Pagallo1990BooleanFD, 10.1023/A:1022641700528, liu1998integrating, 989541, 10.5555/645527.657305, yin2003cpar, cohen1995fast} expressions. CNF learners have been shown to perform competitively with DNF learners~\citep{mooney1995encouraging}, inspiring a line of CNF learning algorithms~\citep{dries2009mining, jain2021learning, beck2023layerwise, sverdlik1992dynamic, hong1997generalized}. Traditional rule-based models are valued for their interpretability but struggle to find the global optimum due to their discrete, non-differentiable nature. Extensive exploration of heuristic methods~\citep{quinlan2014c4, loh2011classification, cohen1995fast} has not consistently yielded optimal solutions.

In response, recent research has turned to Bayesian frameworks to enhance model structure~\citep{Letham_2015, wang2017bayesian, yang2017scalable}, employing strategies such as if-then rules~\citep{lakkaraju2016interpretable} and advanced data structures for quicker training~\citep{angelino2018learning}. Despite these advancements, extended search times, scalability challenges, and performance issues limit the practicality of rule-based models compared to ensemble methods like Random Forest~\citep{breiman2001random} and Gradient Boosted Decision Trees~\citep{chen2016xgboost, ke2017lightgbm}, which trade off interpretability for improved performance.

\subsection{Rule Learning Neural Networks}
Neural rule learning-based methods integrate rule learning with advanced optimization techniques, enabling the discovery of complex and nuanced rules that combine the interpretability of symbolic models with the generalization power of neural networks. Unlike tree-based models, which explicitly follow feature-condition rules, neural approaches rely on weight parameters to control the rule learning process, offering improved robustness and scalability through data-driven training. However, existing approaches such as neural decision trees and rule extraction from neural networks~\citep{frosst2017distillingneuralnetworksoft, ribeiro2016should, wang2020transparent, wang2021scalable, zhang2023learning} face challenges in fidelity and scalability. In particular, RRL~\citep{wang2021scalable, Wang_2024}, a state-of-the-art rule-based neural network, requires a predefined structure of CNF and DNF layers, which limits flexibility, leads to inefficient rule discovery, and exacerbates optimization issues such as gradient vanishing~\citep{wang2020transparent}. 

Recent work on differentiable logic gate networks~\citep{petersen2022deep} learns distributions over all 16 binary Boolean operators with fixed sparse connectivity for fast hardware inference ($>10^6$ inferences/s). However, DLGN uses continuous relaxations during training and discretizes only at inference, whereas RNS maintains discrete operations throughout via Straight-Through Estimator. Moreover, DLGN lacks explicit rule extraction mechanisms, while RNS evaluates rule coverage, diversity, and recovery for interpretable CNF/DNF learning.

Our proposed Rule Network with Selective Logical Operators (RNS) addresses these challenges through its novel architecture and optimization strategies, improving scalability, rule quality, and training stability, as detailed in Section~\ref{sec:method}.

\subsection{Binarized Neural Network}
A related topic to this work is Binarized Neural Networks (BNNs), which optimize deep neural networks by employing binary weights. The deployment of deep neural networks typically requires substantial memory storage and computing resources. To achieve significant memory savings and energy efficiency during inference, recent efforts have focused on learning binary model weights while maintaining the performance levels of their floating-point counterparts~\cite{courbariaux2015binaryconnect, courbariaux2016binaryconnect, rastegari2016xnornet, bulat2019xnornet, liu2018bireal}. Innovations such as bit logical operations~\cite{kim2016bitwise} and novel training strategies for self-binarizing networks~\cite{lahoud2019selfbinarizing}, along with integrating scaling factors for weights and activations~\cite{sakr2018true}, have advanced BNNs. However, due to the binary nature of their weights, BNNs face optimization challenges. The Straight-Through Estimator (STE) method~\cite{courbariaux2015binaryconnect, courbariaux2016binaryconnect, cheng2019straightthrough} allows gradients to "pass through" non-differentiable functions, making it particularly effective for discrete optimization.

Despite both using binarized model weights and employing STE for optimization, our work diverges significantly from BNNs. First, RNS adopts specialized logical activation functions to perform logical operations on features, whereas BNNs typically use the Sign function to produce binary outputs. Second, BNNs are fully connected neural networks, while RNS features a learning mechanism for its connections. Most importantly, these distinctions enable RNS to learn logical rules for both prediction and interpretability, setting it apart from BNNs, which are primarily designed to enhance model efficiency.

\section{ Convergence Analysis}
\label{appendix:convergence}

We analyze convergence for a simplified RNS module with a single Logic Selection Layer (LSL) operating on binary inputs $\mathbf{x} \in \{\pm 1\}^d$. Each neuron $u_j$ selects inputs $S_j \subset \{1,\dots,d\}$ and an operator via $\sigma_j \in \mathbb{R}$: if $\sigma_j < 0$, $u_j$ computes $z_j(\mathbf{x}) = \min_{i\in S_j} x_i$ (AND); if $\sigma_j > 0$, $z_j(\mathbf{x}) = \max_{i\in S_j} x_i$ (OR). The network minimizes population loss $L(\theta) = \mathbb{E}_{(\mathbf{x},y)\sim\mathcal{D}}[\ell(f_\theta(\mathbf{x}),y)]$.

\textbf{STE for min/max operations.}
We use straight-through estimators for both min/max operations and the sign of $\sigma_j$. For $z = \max(x_{i_1},\dots,x_{i_k})$, we set
\begin{equation}
\frac{\partial z}{\partial x_{i_m}}
= \begin{cases}
\displaystyle \frac{1}{|\mathcal{M}|}, & x_{i_m} = \max_\ell x_{i_\ell} \\
0, & \text{otherwise},
\end{cases}
\end{equation}
uniformly distributing gradients among all maximal inputs. For the selector, we treat $\operatorname{sign}(\sigma_j)$ as having unit derivative near zero (identity STE). Let $\tilde{\nabla} L(\theta)$ denote the resulting coarse gradient and consider updates $\theta_{t+1} = \theta_t - \eta \, \tilde{\nabla} L(\theta_t)$.

Following convergence analyses for activation-quantized networks~\citep{yin2019understanding}, we assume: (1) bounded data support with $\pm1$ inputs; (2) compact parameter set; (3) $L$-Lipschitz population gradient; (4) proper STE alignment: $\langle \mathbb{E}[\tilde{\nabla} L(\theta)], \nabla L(\theta)\rangle \geq c \|\nabla L(\theta)\|^2$ for $c>0$; (5) step size $0 < \eta < 2/L$.

Under these conditions, coarse-gradient descent satisfies: $L(\theta_{t+1}) \le L(\theta_t)$ (monotone descent); $\sum_{t=0}^\infty \|\nabla L(\theta_t)\|^2 < \infty$, hence $\|\nabla L(\theta_t)\|\to 0$; and every limit point is a critical point of $L$. When the target is exactly an AND/OR over the same inputs, convergence reaches the global minimum.

\begin{figure}[t]
\centering
\begin{subfigure}[b]{0.24\textwidth}
    \centering
    \includegraphics[width=\textwidth]{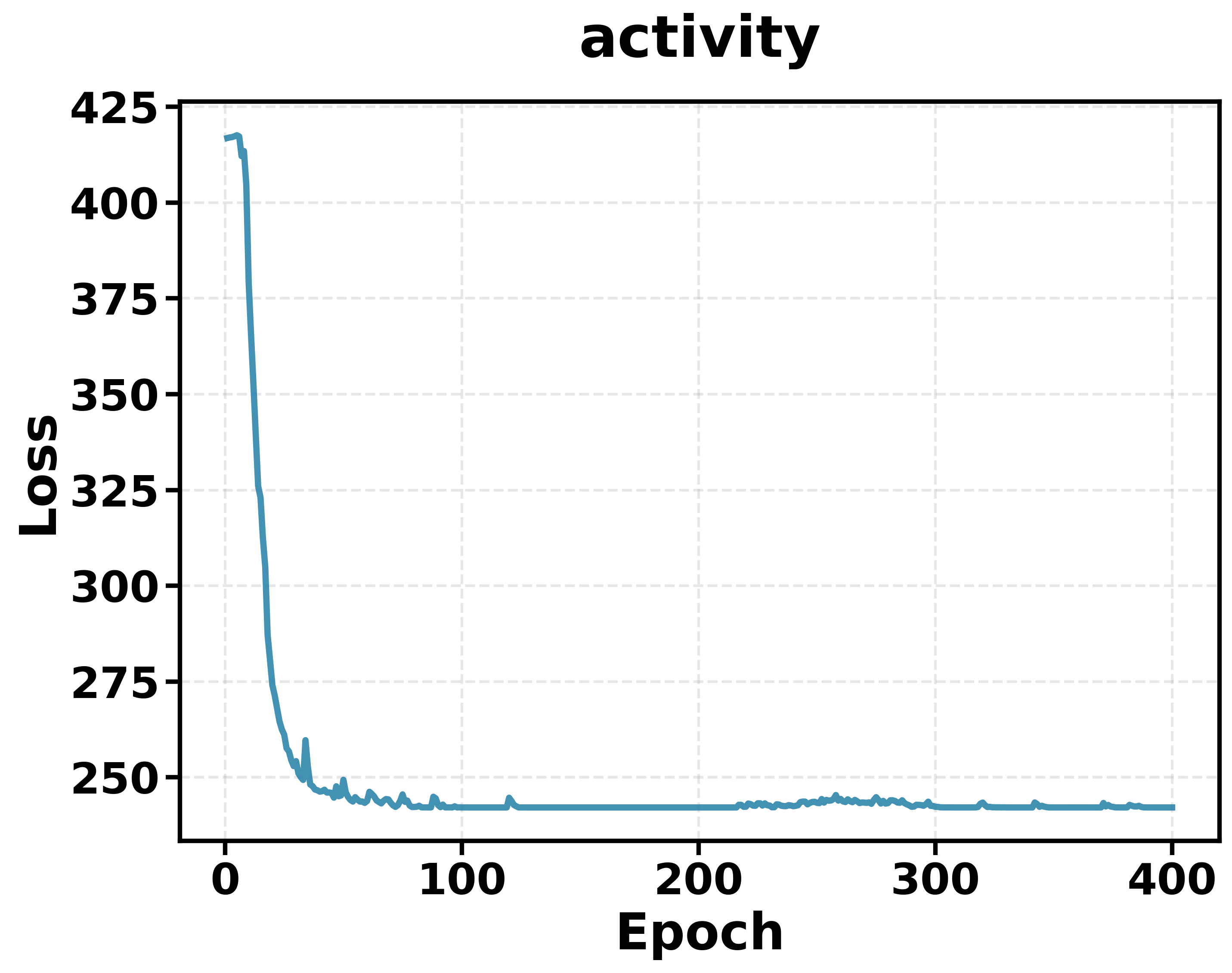}
\end{subfigure}
\hfill
\begin{subfigure}[b]{0.24\textwidth}
    \centering
    \includegraphics[width=\textwidth]{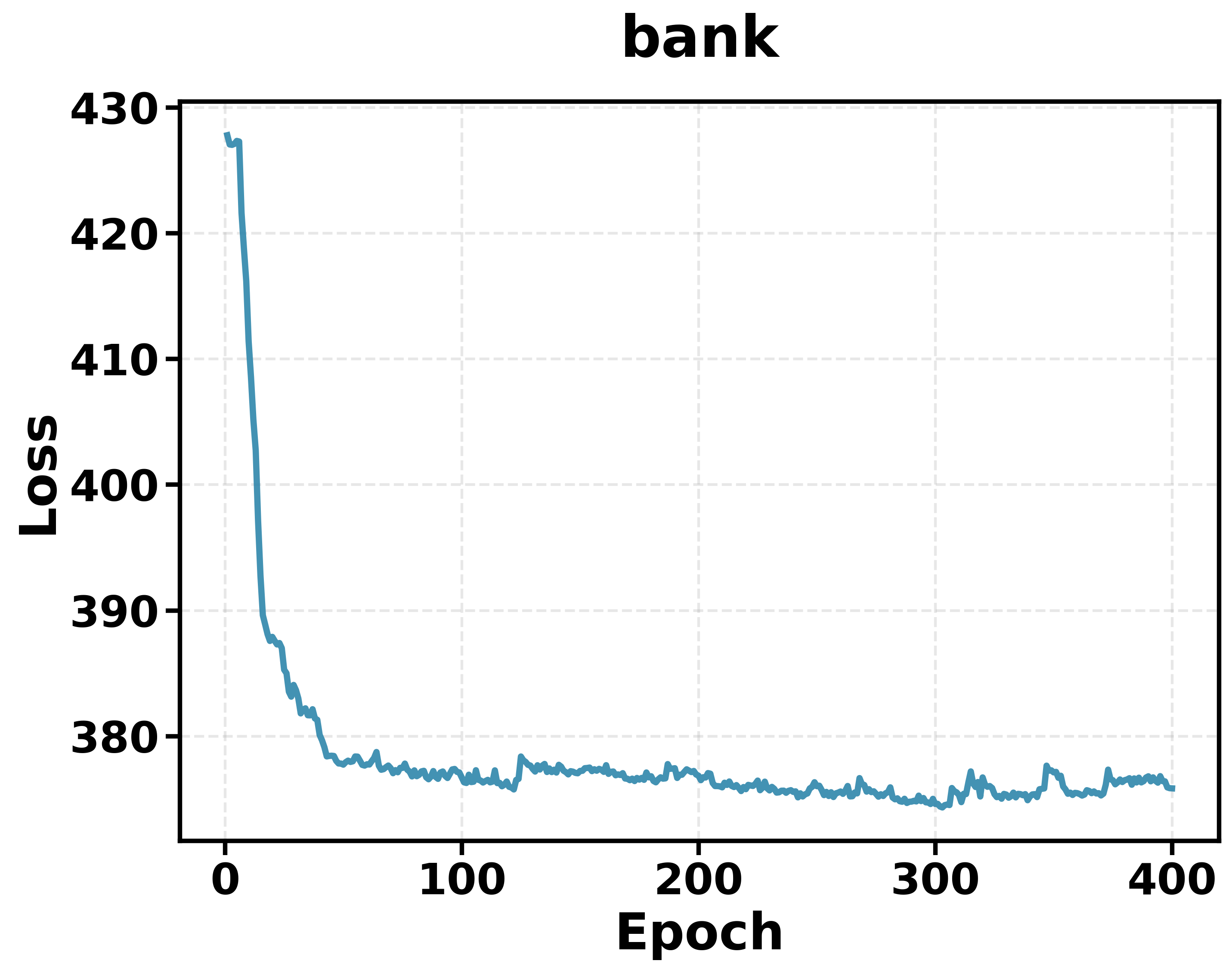}
\end{subfigure}
\hfill
\begin{subfigure}[b]{0.24\textwidth}
    \centering
    \includegraphics[width=\textwidth]{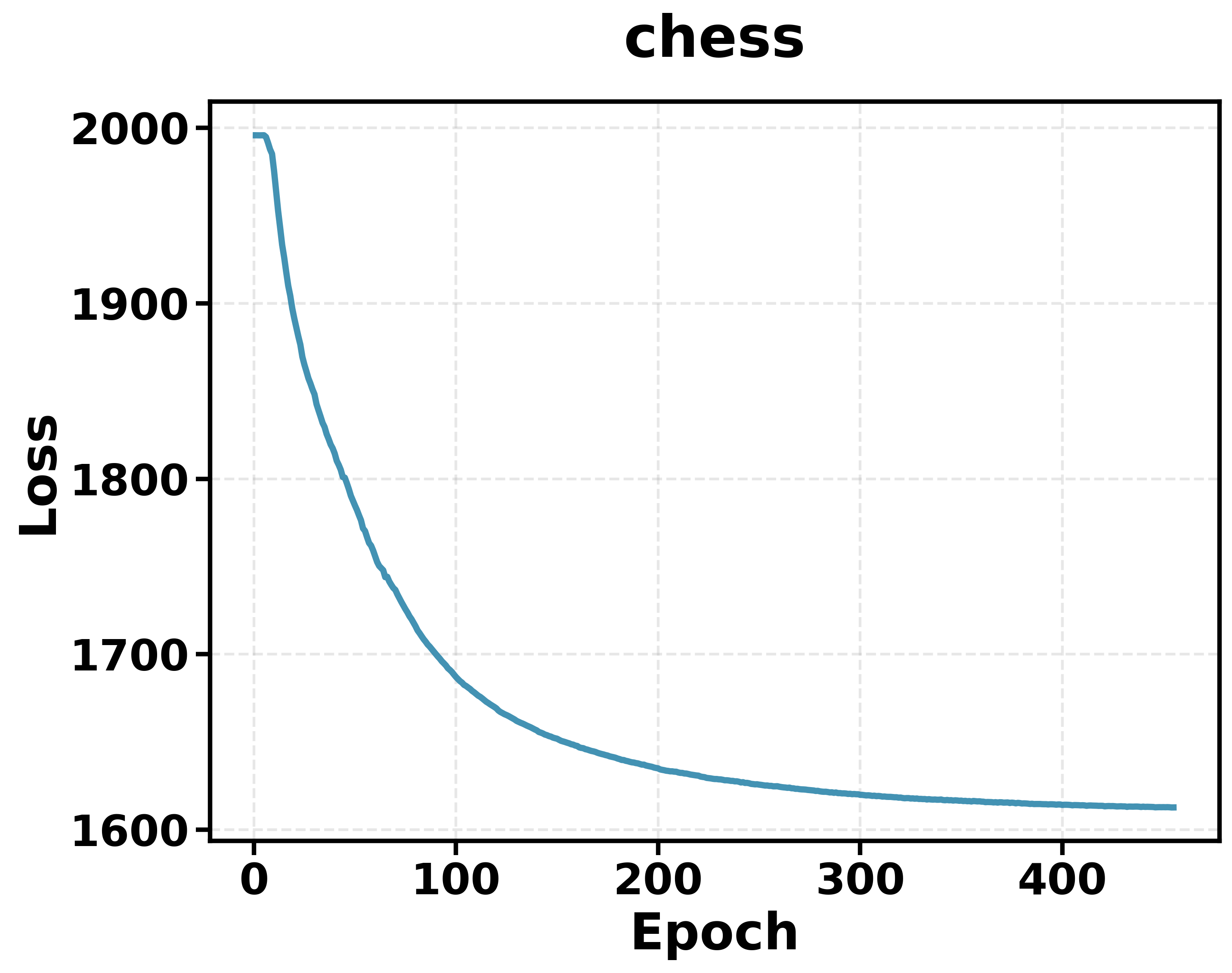}
\end{subfigure}
\hfill
\begin{subfigure}[b]{0.24\textwidth}
    \centering
    \includegraphics[width=\textwidth]{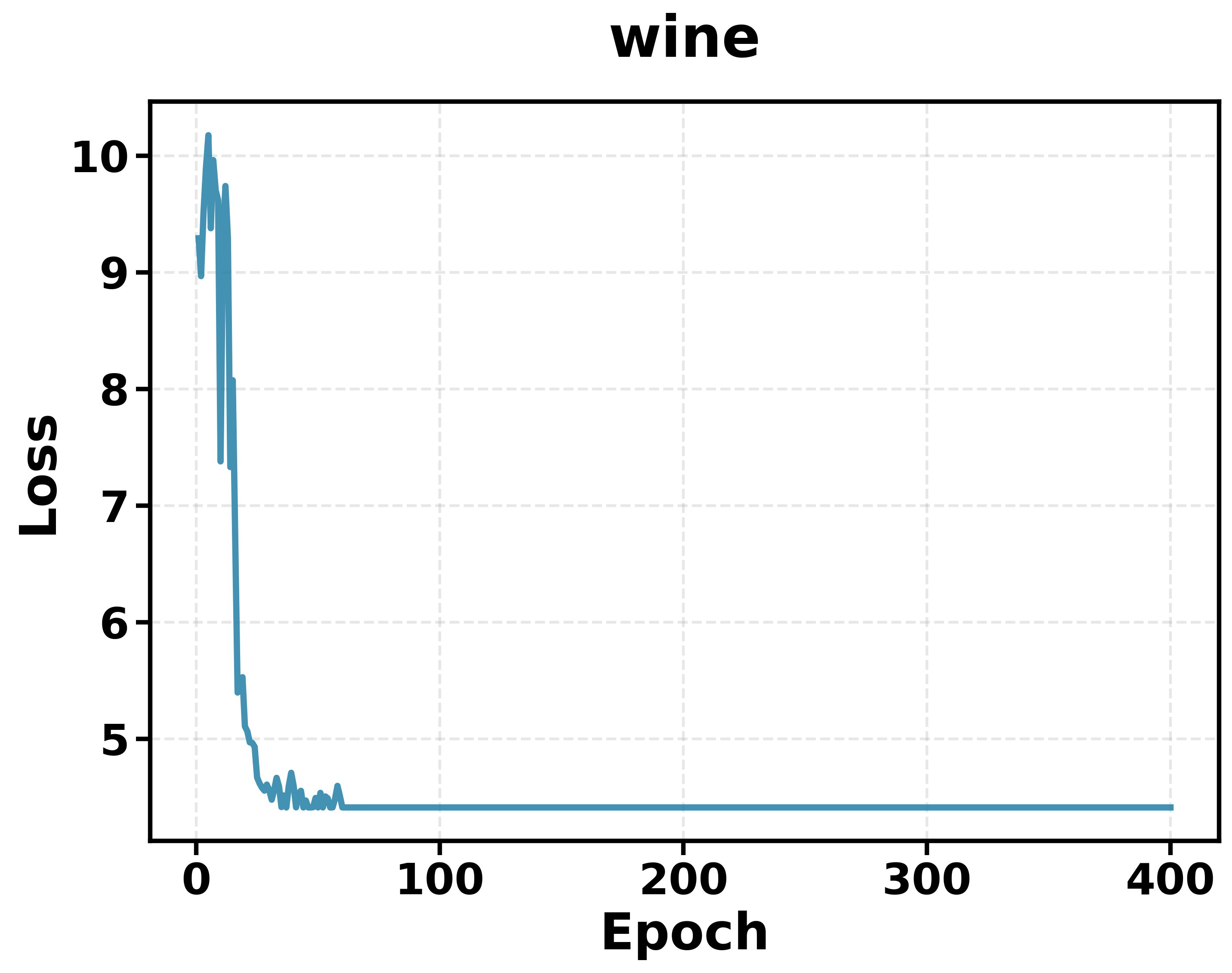}
\end{subfigure}

\begin{subfigure}[b]{0.24\textwidth}
    \centering
    \includegraphics[width=\textwidth]{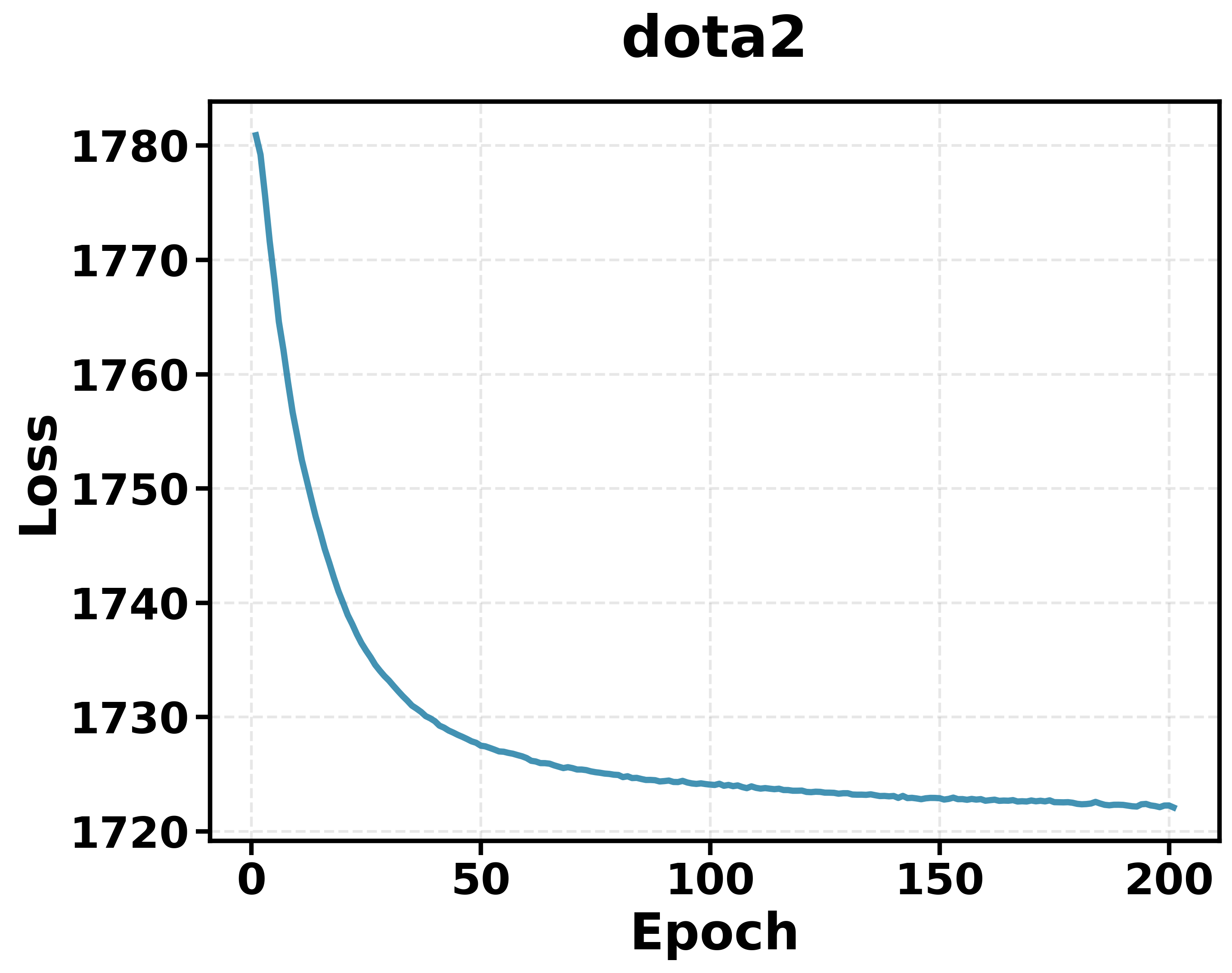}
\end{subfigure}
\hfill
\begin{subfigure}[b]{0.24\textwidth}
    \centering
    \includegraphics[width=\textwidth]{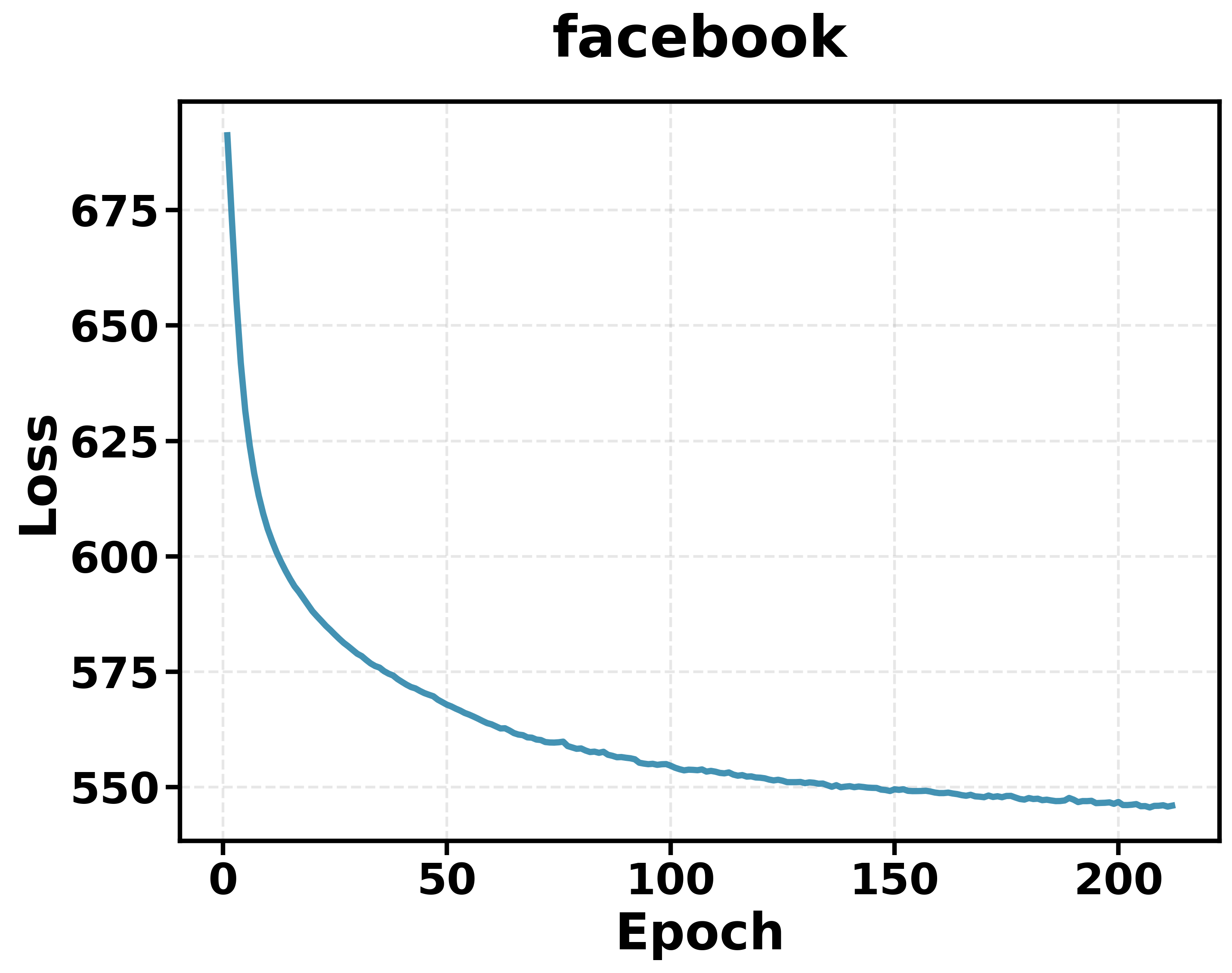}
\end{subfigure}
\hfill
\begin{subfigure}[b]{0.24\textwidth}
    \centering
    \includegraphics[width=\textwidth]{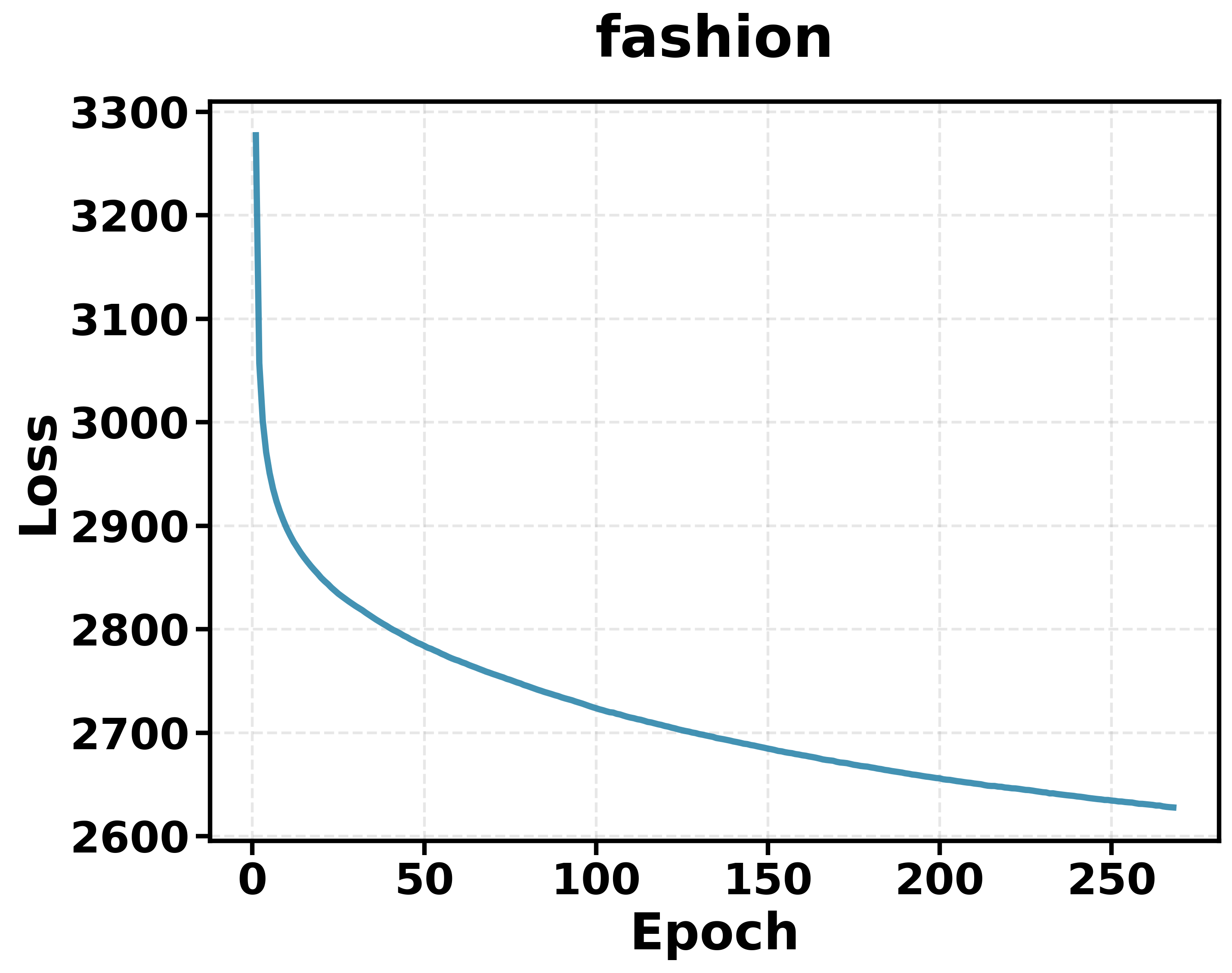}
\end{subfigure}
\hfill
\begin{subfigure}[b]{0.24\textwidth}
    \centering
    \includegraphics[width=\textwidth]{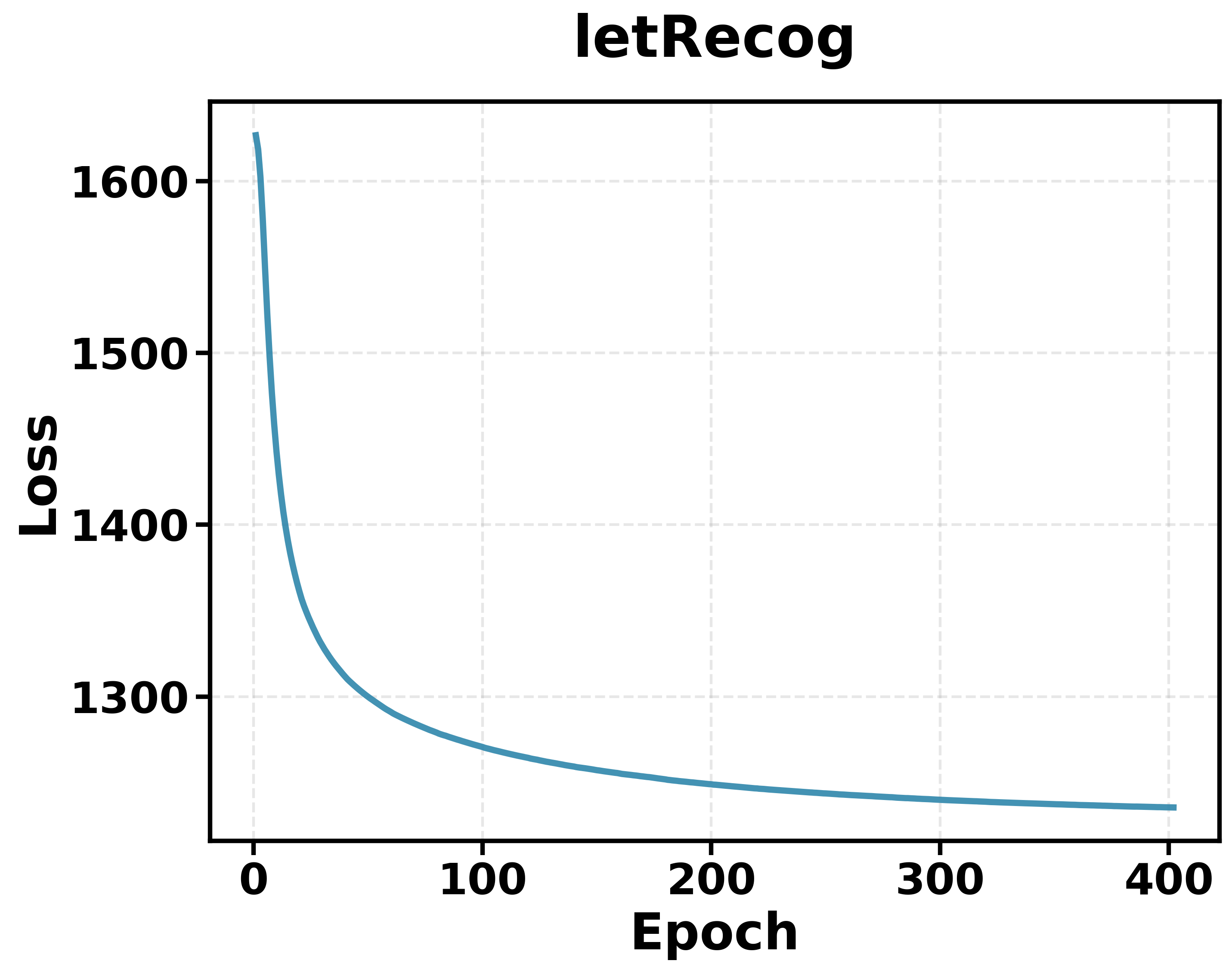}
\end{subfigure}

\caption{\textbf{Training loss convergence across diverse datasets.} Loss trajectories for RNS on eight datasets with varying sizes and complexities. All curves show rapid initial descent and stable convergence, validating our theoretical analysis.}
\label{fig:convergence_curves}
\end{figure}

\textbf{Empirical Convergence Results.} Figure~\ref{fig:convergence_curves} shows training loss trajectories across eight diverse datasets, validating our theoretical guarantees. All datasets exhibit: (1) rapid initial descent indicating effective gradient flow; (2) near-monotonic decrease consistent with $L(\theta_{t+1}) \le L(\theta_t)$; (3) stable convergence to low loss values.

Convergence speeds vary by dataset characteristics: \texttt{wine} converges within 50 epochs due to its small size and clear separation; \texttt{dota2} stabilizes in 50-75 epochs; larger datasets like \texttt{fashion} (784 features, 10 classes) and \texttt{letRecog} require 200-300 epochs. The \texttt{bank} dataset shows minor oscillations at epochs 300-400 from mini-batch stochasticity on imbalanced data, but all datasets achieve stable convergence with $\|\nabla L(\theta_t)\| \to 0$.

These results confirm that proper STE design enables reliable gradient-based optimization of discrete logical networks across diverse problem scales and class distributions, avoiding the vanishing gradient issues of product-based activations.

\section{Simulation Experiment}
\label{appendix:simu}

We conduct controlled simulation experiments to assess RNS's ability to learn the exact logical rules used to generate synthetic data, examining both basic and extended settings with increasing complexity.

\subsection{Basic Setting}

\noindent \textbf{Setup.} We generate synthetic data based on predefined logical rules and train rule-based models to evaluate their ability to reconstruct these ground-truth rules. The dataset is synthesized by defining three probability parameters, $\mathbf{p} = (p_1, p_2, p_3)$, corresponding to feature variables $\{x_1, x_2, x_3\}$, each drawn independently from a uniform distribution $U(0, 1)$. Using these as Bernoulli parameters, we generate $X_{gen}\in\mathbb{R}^{3\times 50000}$ by repeating Bernoulli sampling, resulting in $n=50000$ binary vectors of length 3, where each element is sampled from $Bernoulli(p_i)$. Labels are assigned to these binary vectors based on specific logical rules. For example, a rule $x_1 \land x_2 \land x_3 \rightarrow 1$ assigns a label of 1 if all three features are 1. We define four different types of rules, presented as "Ground-Truth Rules" in Table~\ref{tab:synthetic}. The dataset is split evenly into training and test sets. Both RNS and RRL are configured with logical layer dimensions of $64@64$.

\noindent \textbf{Results.} As shown in Table~\ref{tab:synthetic}, RNS not only achieves near 100\% accuracy but also recovers the exact structure of the ground-truth rules. In contrast, RRL, even with a naive negation setting, tends to produce overly simplified, redundant, or logically incomplete rules. For example, RNS is able to precisely recover rules such as $x_1 \land \lnot x_2 \land x_3$ and $x_2 \land \lnot x_3$, faithfully mirroring the underlying logic used to generate the data. RRL, on the other hand, often produces rules that are either too general (e.g., $x_1$, $x_2$) or combine terms in a way that does not fully reflect the intended logic (e.g., $(x_1 \land x_3), x_2$). This pattern holds even when RRL is augmented with the naive negation setting: RNS consistently recovers all ground-truth rules, whereas RRL fails to identify several rules in their correct logical form and often outputs multiple trivial or repeated variants for a single ground-truth pattern.

\subsection{Extended Setting with Higher Complexity}

To address concerns about the scalability of rule recovery to more complex logical structures, we extend the simulation experiments beyond the 3-variable setting.

\noindent \textbf{Setup.} We extend the simulation framework to higher-dimensional settings with more features. For each complexity level, we generate synthetic datasets following the same procedure: probability parameters $\mathbf{p} = (p_1, \ldots, p_d)$ are sampled uniformly from $U(0,1)$, and binary feature vectors are generated via Bernoulli sampling to produce $n=150{,}000$ samples. Labels are assigned based on predefined logical rules of increasing structural complexity, including nested disjunctions, conjunctions with multiple negations, and rules with longer conjunctive clauses. All models are configured with logical layer dimensions of $2048@2048$, and datasets are split evenly into training and test sets.

\noindent \textbf{Results.} As shown in Table~\ref{tab:extended_recovery}, RNS consistently recovers the exact ground-truth logical structure across both medium-complexity and high-complexity rules, demonstrating robust scalability. In contrast, both RRL variants struggle to capture complete logical structures as complexity increases.

For the medium complexity rules, RNS precisely recovers all ground-truth patterns, including complex structures like $(x_{1}\vee x_{2})\wedge \neg x_{3}\wedge x_{4}$ (decomposed into its two disjunctive components) and $x_{1}\vee (\neg x_{2}\wedge x_{3}\wedge \neg x_{5})$. RRL produces incomplete rules that miss critical negations or introduce spurious disjunctions (e.g., $x_{1}\vee x_{3}\vee (x_{3}\vee x_{5})$ instead of the correct structure). RRL with naive negation performs slightly better but still produces overly simplified rules that fail to capture all logical constraints.

For the high-complexity rules, RNS successfully decomposes the complex three-way disjunction $(x_{1}\wedge x_{2}\wedge \neg x_{3})\vee (x_{4}\wedge x_{5}\wedge x_{7})\vee (\neg x_{8}\wedge x_{9})$ into its exact constituent clauses. In contrast, RRL systematically drops negation constraints (e.g., missing $\neg x_{3}$) and introduces spurious conjunctions (e.g., $x_{7}\wedge x_{9}$), while RRL with naive negation produces overly general rules that lose structural precision (e.g., reducing $\neg x_{8}\wedge x_{9}$ to just $x_{9}$).

These results demonstrate that RNS's advantage in exact rule recovery is not limited to simple, low-dimensional settings but extends to substantially more complex logical structures, validating its utility for interpretable learning in realistic scenarios.

\begin{table*}[t]
\centering
\setlength{\tabcolsep}{3pt}
\begin{tabular}{|p{3.2cm}|p{4.2cm}|p{3cm}|p{3cm}|}
\hline
\textbf{Ground-Truth Rules} & \textbf{RNS} & \textbf{RRL} & \textbf{RRL w/ naive negation} \\
\hline
\multicolumn{4}{|c|}{\textit{Medium Complexity}} \\
\hline
$(x_{1}\vee x_{2})\wedge \neg x_{3}\wedge x_{4}$ 
& $x_{1}\wedge \neg x_{3}\wedge x_{4}$, $x_{2}\wedge \neg x_{3}\wedge x_{4}$ 
& $x_{1}\wedge x_{2}\wedge x_{4}$, $x_{1}\wedge x_{4}$, $x_{2}$ 
& \textbf{$x_{1}\wedge x_{4}$}, $x_{2}\wedge \neg x_{3}$, $x_{4}$ \\
\hline
$x_{1}\vee (\neg x_{2}\wedge x_{3}\wedge \neg x_{5})$ 
& $x_{1}$, $\neg x_{2}\wedge x_{3}\wedge \neg x_{5}$ 
& $x_{1}\vee x_{3}\vee (x_{3}\vee  x_{5})$ 
& \textbf{$x_{1}$}, $x_{3}$, $\neg x_{2}\wedge x_{3}$ \\
\hline
$(x_{1}\wedge x_{2})\vee (x_{3}\wedge \neg x_{4}\wedge x_{5})$ 
& $x_{1}\wedge x_{2}$, $x_{3}\wedge \neg x_{4}\wedge x_{5}$ 
& $x_{1}\wedge x_{2}$, $x_{3}\wedge x_{5}$, $x_{1}\wedge x_{2}\wedge x_{3}$ 
& \textbf{$x_{1}\wedge x_{2}$}, $x_{3}\wedge x_{5}$, $\neg x_{4}$ \\
\hline
\multicolumn{4}{|c|}{\textit{High Complexity}} \\
\hline
$(x_{1}\wedge x_{2}\wedge \neg x_{3})\vee (x_{4}\wedge x_{5}\wedge x_{7})\vee (\neg x_{8}\wedge x_{9})$ 
& $x_{1}\wedge x_{2}\wedge \neg x_{3}$, $x_{4}\wedge x_{5}\wedge x_{7}$, $\neg x_{8}\wedge x_{9}$ 
& $x_{1}\wedge x_{2}$, $x_{4}\wedge x_{5}\wedge x_{7}$, $x_{7}\wedge x_{9}$ 
& \textbf{$x_{1}\wedge x_{2}$}, $x_{4}\wedge x_{5}\wedge x_{7}$, $x_{9}$ \\
\hline
\end{tabular}
\caption{Logical rules learned by RNS, RRL, and RRL with naive negation on extended synthetic data with more features. RNS consistently recovers the exact ground-truth structure, while RRL produces incomplete or overly simplified rules.}
\label{tab:extended_recovery}
\end{table*}
\section{Qualitative Analysis and Model Interpretation}
\label{appendix:case_study}

We examine learned rules from multiple datasets to demonstrate that RNS naturally discovers interpretable logical structures that reflect meaningful domain knowledge.

\subsection{Adult Income Dataset}

We examine a learned rule from the Adult income dataset that predicts low income ($\leq$50K) through a conjunction of disjunctive clauses (CNF structure):
\begin{equation*}
\begin{aligned}
&(\text{capital-gain} < 17{,}939 \vee \text{education-num} < 12.8) \wedge \\
&(\neg\text{Exec-managerial} \vee \text{capital-gain} < 15{,}318) \wedge \\
&(\neg\text{Prof-specialty} \vee \text{education-num} < 12.8)
\end{aligned}
\end{equation*}
Each clause constrains a wealth-building mechanism: (1) requiring either low capital gains or limited education, (2) excluding executives with substantial investments ($>\$15{,}000$), and (3) excluding professional specialists with college degrees. The conjunctive structure encodes that low income results from lacking multiple advantages simultaneously—neither education nor capital gains nor high-status occupation alone suffices to escape this prediction. This demonstrates RNS's ability to capture interpretable domain structure where compound constraints reflect economic reality.

\subsection{Wine Quality Dataset}

We examine a learned rule from the Wine Quality dataset that predicts high-quality wine (Class-2) with 94.37\% support, covering nearly all instances of this quality class:
\begin{equation*}
\begin{aligned}
&(\text{total-SO}_2 < 3.729 \vee \text{free-SO}_2 < 3.490) \wedge \\
&(\text{alcohol} > 0.301 \wedge \text{alcohol} < 0.391) \wedge \\
&(\text{citric-acid} > 1.638 \wedge \text{citric-acid} < 1.700) \wedge \\
&(\text{density} > 0.066 \wedge \text{density} < 0.132) \wedge \\
&(\text{volatile-acidity} < 4.797) \wedge (\text{pH} > 0.120) \wedge \\
&(\text{total-SO}_2 > -0.094 \vee \text{free-SO}_2 > 0.938)
\end{aligned}
\end{equation*}
This rule exhibits CNF structure where wine quality requires multiple chemical properties to simultaneously satisfy constraints. Most clauses are unit literals establishing upper or lower bounds (e.g., $1.638 < \text{citric-acid} < 1.700$ creates a narrow acceptable range of width 0.062), while two clauses are disjunctions offering alternative sulfur dioxide conditions. Unlike the Adult dataset where high income admits multiple alternative paths (DNF), wine quality demands that all conjunctive clauses be satisfied—reflecting the domain knowledge that wine quality depends on balanced chemistry rather than any single exceptional property. The exceptionally high support (94.37\%) indicates this rule captures the essential characteristics defining this quality class.

\subsection{Bank Marketing Dataset}

\begin{figure}[t]
    \centering
    \includegraphics[width=1\linewidth]{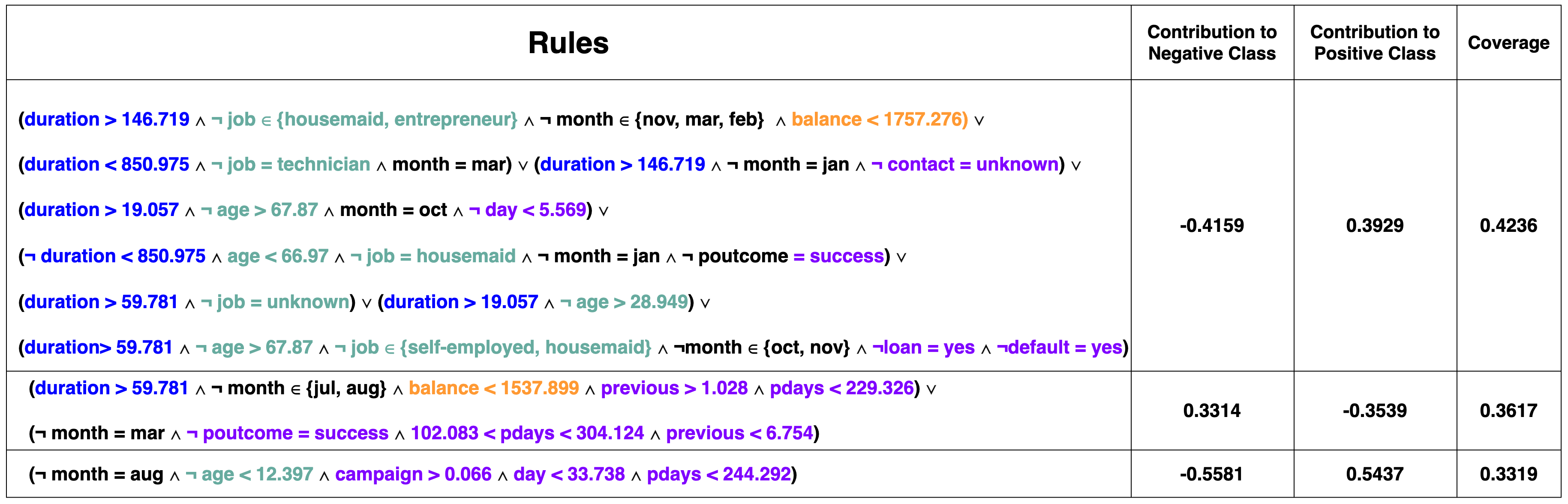}
    \caption{Logical rules from RNS on the bank-marketing dataset. ``Coverage'' denotes the proportion of training samples satisfying a rule.}
    \label{fig:case_study}
\end{figure}

Figure~\ref{fig:case_study} presents the three most discriminative rules extracted from RNS, trained on the bank-marketing dataset to identify customer profiles likely to subscribe to a term deposit through telesales. These rules provide detailed profiles by highlighting specific conditions that increase subscription likelihood.

The transparent interpretations reveal several insights. Subscriptions are less likely during colder seasons, possibly due to reluctance to make financial decisions during holidays. The model identifies a correlation between longer call durations and successful marketing, suggesting that extended dialogues often indicate interest in deposits. It also highlights that housemaids, entrepreneurs, and self-employed individuals are less inclined towards term deposits, potentially due to financial constraints or the need for liquidity. Additionally, the analysis shows age-related trends in call duration, with younger individuals having shorter calls and middle-aged clients engaging in longer discussions. These findings demonstrate the model's ability to reflect plausible real-world behaviors while avoiding direct causal assertions.

\section{Dataset Statistics} \label{dataset stat}
\begin{table}[t]
\centering
\small
\setlength{\tabcolsep}{6pt}
\begin{tabular}{lcccc}
\toprule
Dataset & \#instances & \#classes & \#features & feature type\\
\midrule
adult          & 32561 & 2 & 14 & mixed\\ 
bank           & 45211 & 2 & 16 & mixed\\
banknote       & 1372  & 2 & 4  & continuous\\
chess          & 28056 & 18 & 6 & discrete\\
connect-4      & 67557 & 3 & 42 & discrete\\
letRecog       & 20000 & 26 & 16 & continuous \\ 
magic04        & 19020 & 2 & 10 & continuous \\
tic-tac-toe    & 958   & 2 & 9  & discrete\\
wine           & 178   & 3 & 13 & continuous\\
\midrule
activity       & 10299 & 6 & 561 & continuous\\ 
dota2          & 102944& 2 & 116 & discrete\\ 
facebook       & 22470 & 4 & 4714& discrete \\
fashion        & 70000 & 10 & 784 & continuous\\
\bottomrule
\end{tabular}
\caption{Datasets statistics.}
\label{tab:dataset_stats}

\end{table}
In Table~\ref{tab:dataset_stats}, the first nine data sets are small, while the last four are large. The "Discrete" or "Continuous" feature type indicates that all features in the data set are either discrete or continuous, respectively. The "Mixed" feature type indicates that the corresponding data set contains both discrete and continuous features.

\section{Reproducibility}
\textbf{Data Preprocessing.} All datasets are converted into tabular feature vectors to ensure consistent treatment across different data modalities. For standard tabular benchmarks (Adult, Bank, Chess, Connect-4, etc.), we use the original attributes provided by the datasets. Categorical features are one-hot encoded, while continuous features are treated as real-valued inputs (as summarized in Table~\ref{tab:dataset_stats}) and then binarized using the scheme described in Section~\ref{sec:binarization}. For Fashion-MNIST, each 28×28 grayscale image is flattened into a 784-dimensional continuous feature vector, and pixel values are passed through the same feature binarization layer as other continuous features. The binarization method uses random binning in main experiments, with ablations over EntInt/KInt/AutoInt in Appendix H. No additional feature engineering specific to images is applied; Fashion-MNIST is treated purely as a high-dimensional tabular dataset. All methods operate on the same base preprocessed features to ensure fair comparison: interpretable neuro-rule baselines (RRL, CRS) use the same binarized representation as in their original code following Wang et al.~\citep{wang2021scalable, Wang_2024}, while non-interpretable tabular models (FT-Transformer, SAINT, NODE, TabNet, LGBM, XGB) are trained on the same continuous/tabular features before binarization.

\textbf{Model Configuration.} RNS includes two  Logic Selection Layers (LSLs), with the number of logical neurons grid-searched from 32 to 4096 based on dataset complexity. For training, we use cross-entropy loss with L2 regularization to control model complexity, with the regularization coefficient searched in the range $10^{-5}$ to $10^{-9}$. The number of bins in the feature binarization layer is selected from $\{15, 30, 50\}$. The model is trained using the Adam optimizer~\citep{kingma2014adam} with a batch size of 32. For small datasets, training runs for 400 epochs, with the learning rate reduced by 10\% every 100 epochs. For large datasets, training is conducted for 100 epochs with a similar learning rate schedule, decreasing every 20 epochs. Baseline settings follow those described in~\citep{wang2021scalable, Wang_2024}. RNS is implemented in PyTorch~\citep{paszke2019pytorch}, and we use the high-quality code base from~\citep{wang2021scalable, Wang_2024} for baseline comparisons. Experiments are performed on a Linux server equipped with an NVIDIA A100 80GB GPU. All code and data are available at \url{https://anonymous.4open.science/r/RNS_-3DDD}.



\section{Efficiency} \label{appendix:Efficiency all}
\begin{figure*}[ht]
  \centering
  \begin{minipage}[t]{0.20\linewidth}
    \includegraphics[width=\linewidth]{figures/rule_number_figure/adult_rule_numbers.png}
    \label{fig:adult_number}
  \end{minipage}%
  \hfill%
  \begin{minipage}[t]{0.20\linewidth}
    \includegraphics[width=\linewidth]{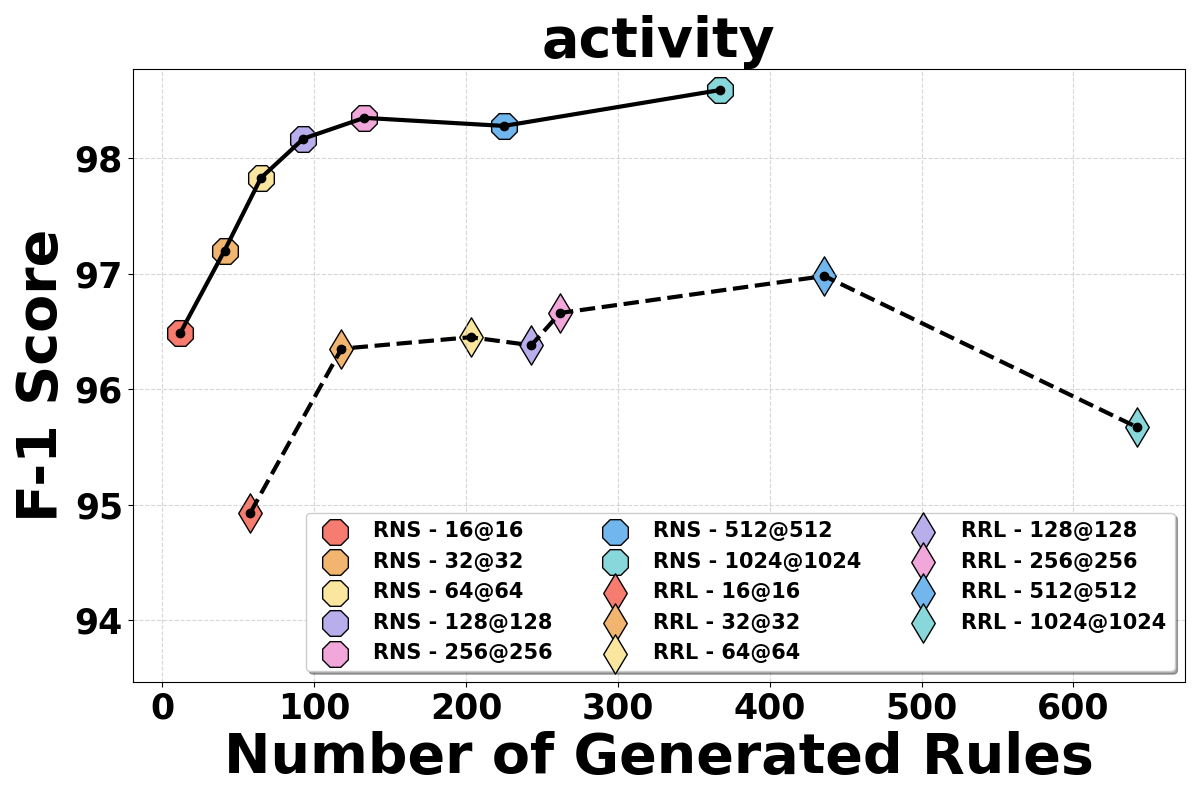}
    \label{fig:activity_number}
  \end{minipage}%
  \hfill%
  \begin{minipage}[t]{0.20\linewidth}
    \includegraphics[width=\linewidth]{figures/rule_number_figure/bank_rule_numbers.png}
    \label{fig:bank_number}
  \end{minipage}%
  \hfill%
  \begin{minipage}[t]{0.20\linewidth}
    \includegraphics[width=\linewidth]{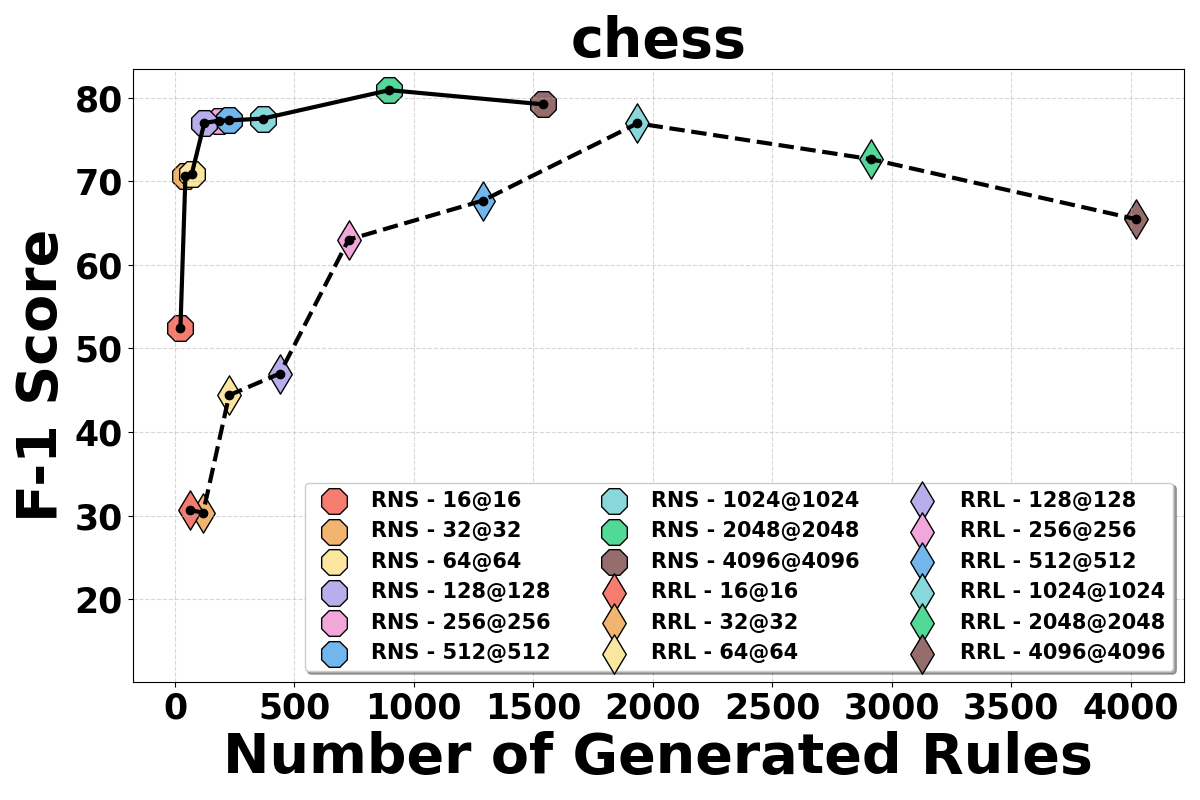}
    \label{fig:chess_number}
  \end{minipage}%
  \hfill%
  \begin{minipage}[t]{0.20\linewidth}
    \includegraphics[width=\linewidth]{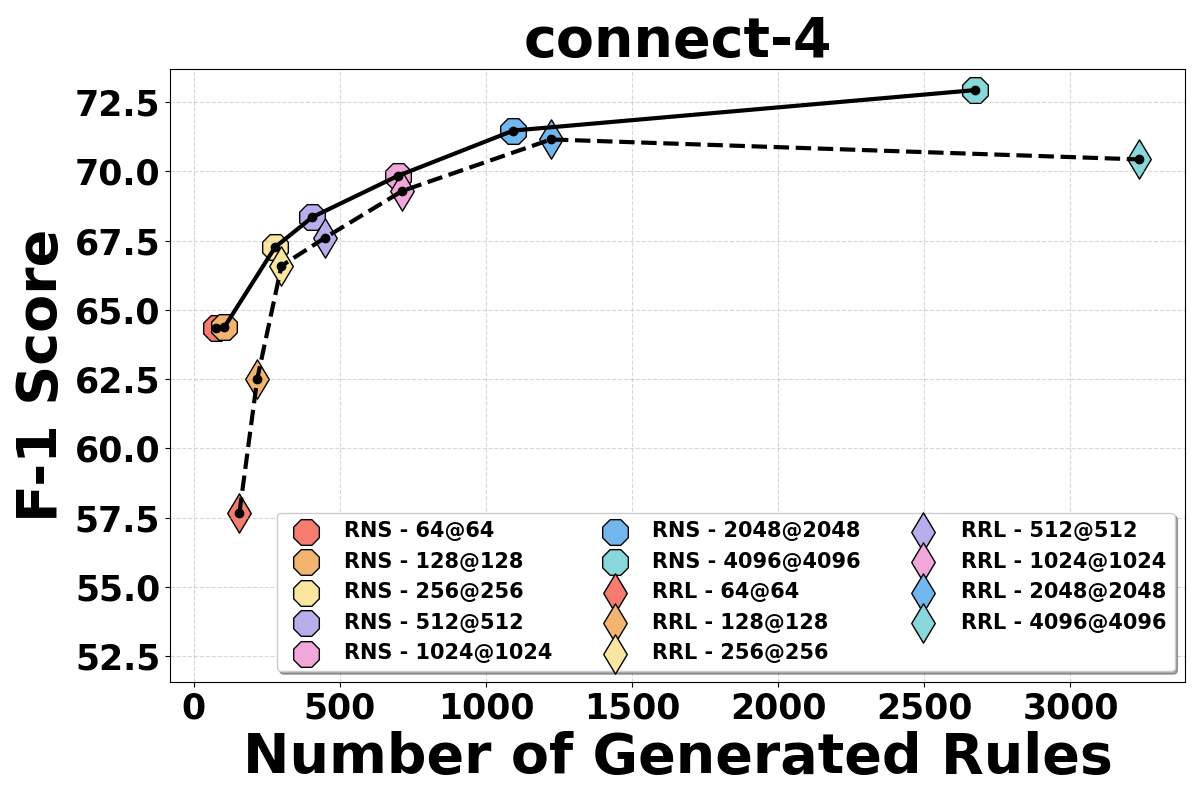}
    \label{fig:connect-4_number}
  \end{minipage}

  \begin{minipage}[t]{0.20\linewidth}
    \includegraphics[width=\linewidth]{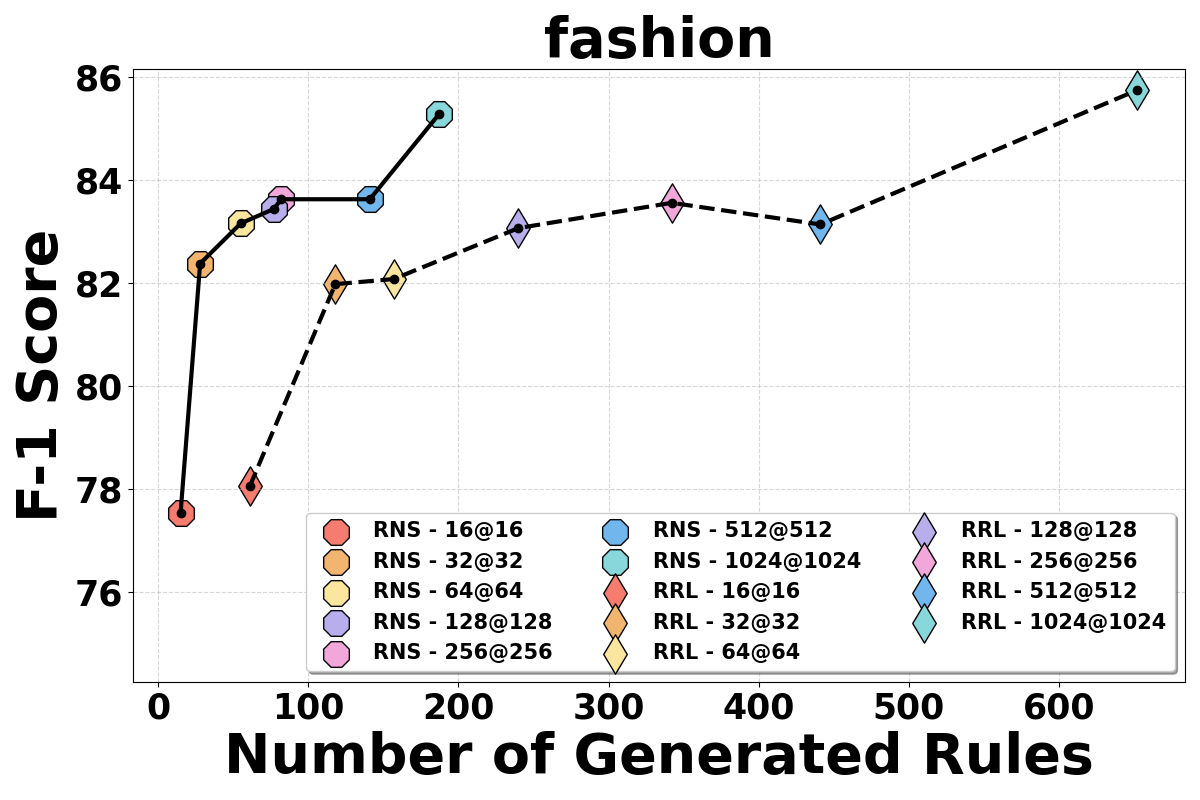}
    \label{fig:fashion_number}
  \end{minipage}%
  \hfill%
  \begin{minipage}[t]{0.20\linewidth}
    \includegraphics[width=\linewidth]{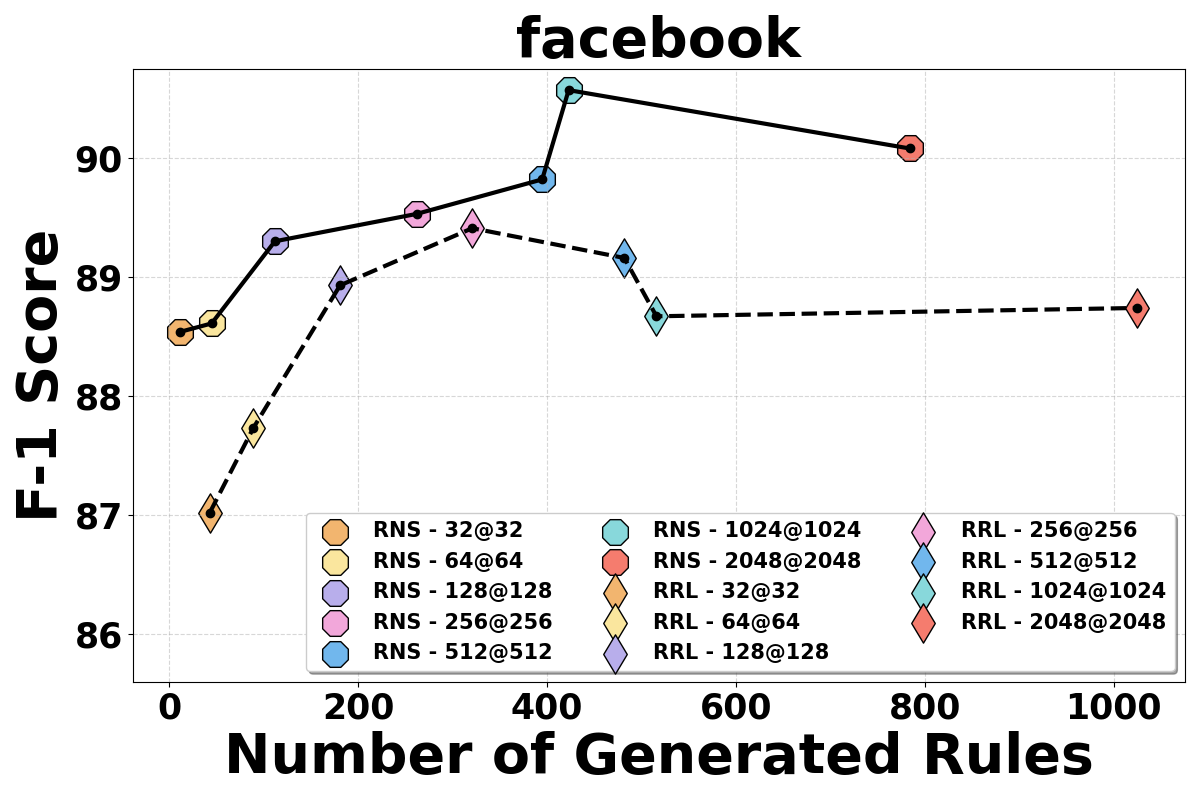}
    \label{fig:facebook_number}
  \end{minipage}%
  \hfill%
  \begin{minipage}[t]{0.20\linewidth}
    \includegraphics[width=\linewidth]{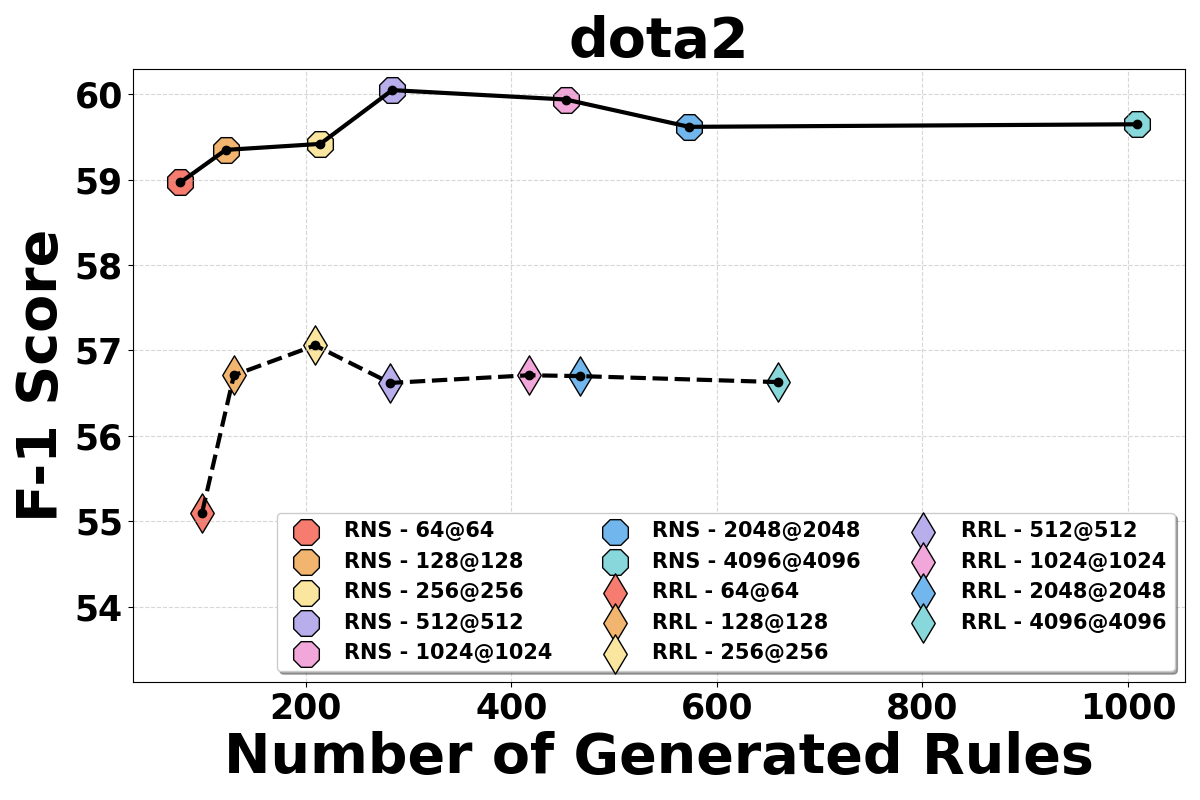}
    \label{fig:dota2_number}
  \end{minipage}%
  \hfill%
  \begin{minipage}[t]{0.20\linewidth}
    \includegraphics[width=\linewidth]{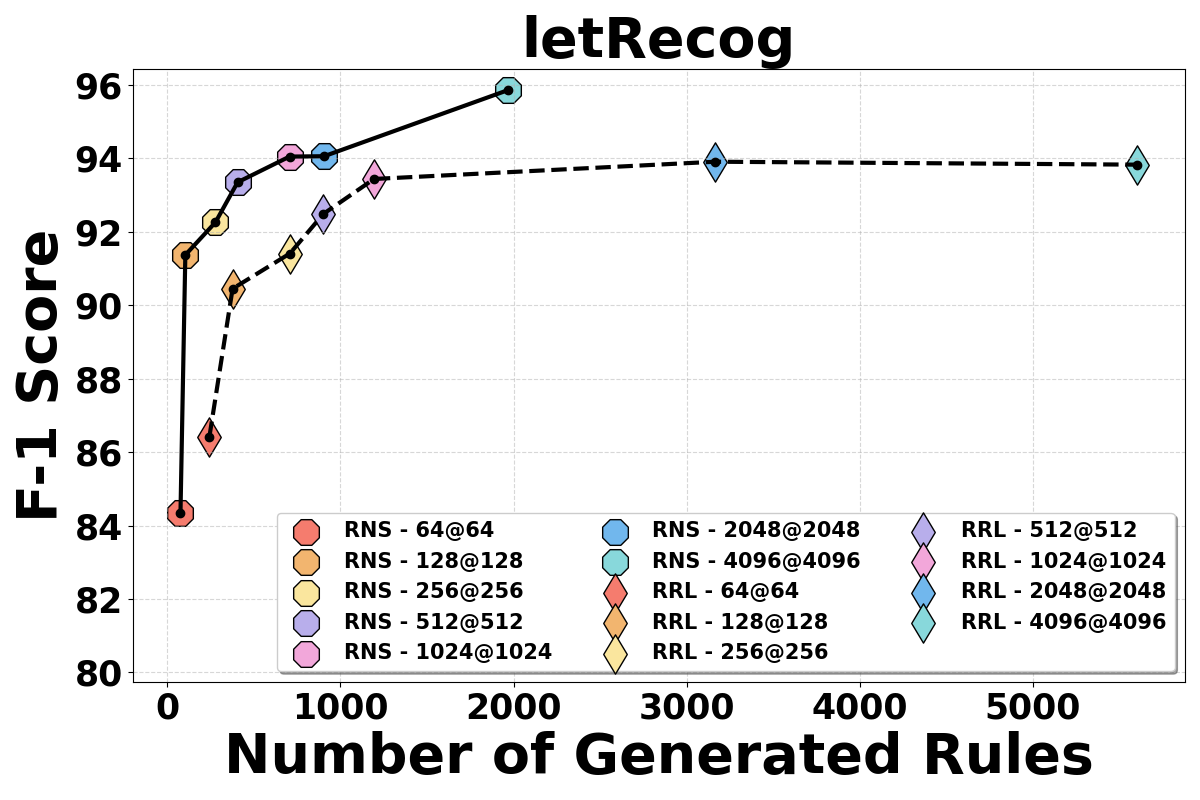}
    \label{fig:letRecog_number}
  \end{minipage}%
  \hfill%
  \begin{minipage}[t]{0.20\linewidth}
    \includegraphics[width=\linewidth]{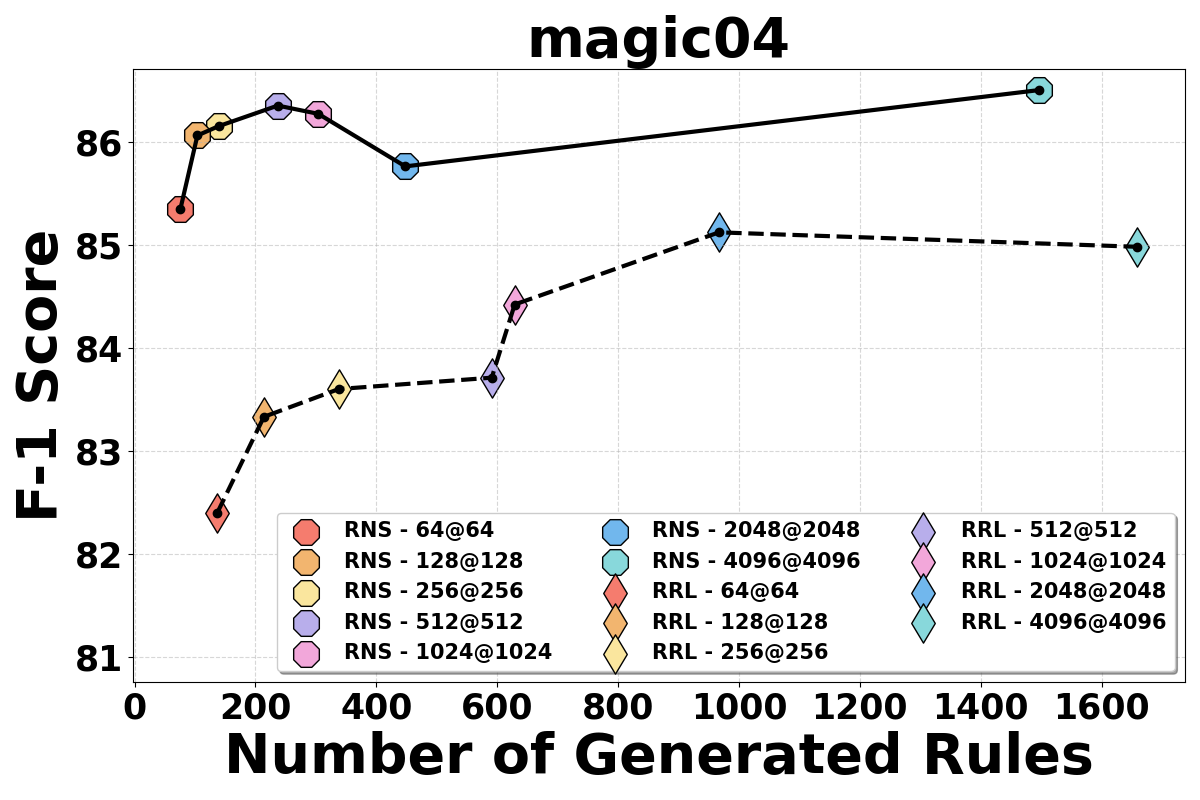}
    \label{fig:magic04_number}
  \end{minipage}

  \caption{Rule Number comparison across different architectures.}
  \label{fig:rule_number_all}
\end{figure*}

\begin{figure*}[ht]
  \centering
  \begin{minipage}[t]{0.20\linewidth}
    \includegraphics[width=\linewidth]{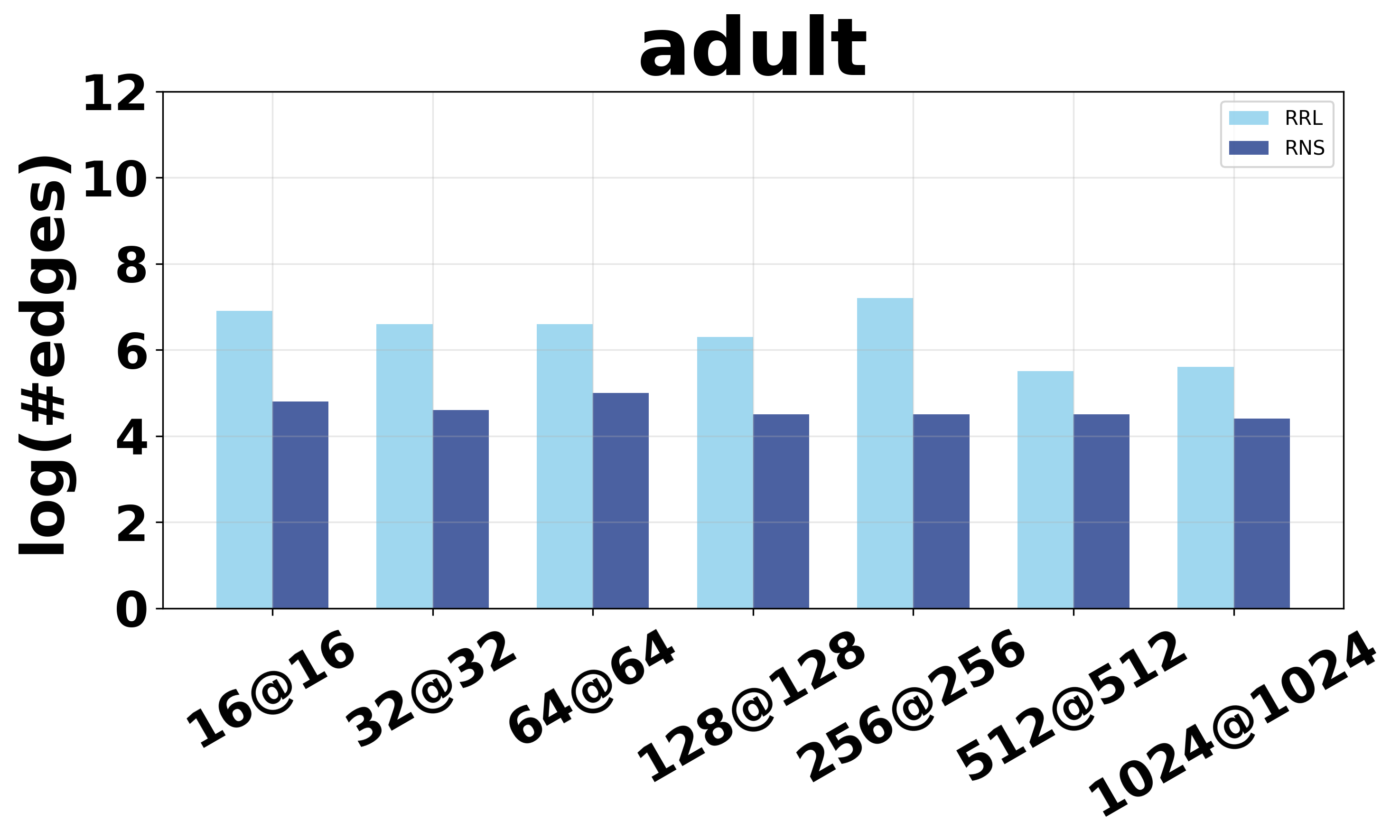}
    \label{fig:adult_length}
  \end{minipage}%
  \hfill%
  \begin{minipage}[t]{0.20\linewidth}
    \includegraphics[width=\linewidth]{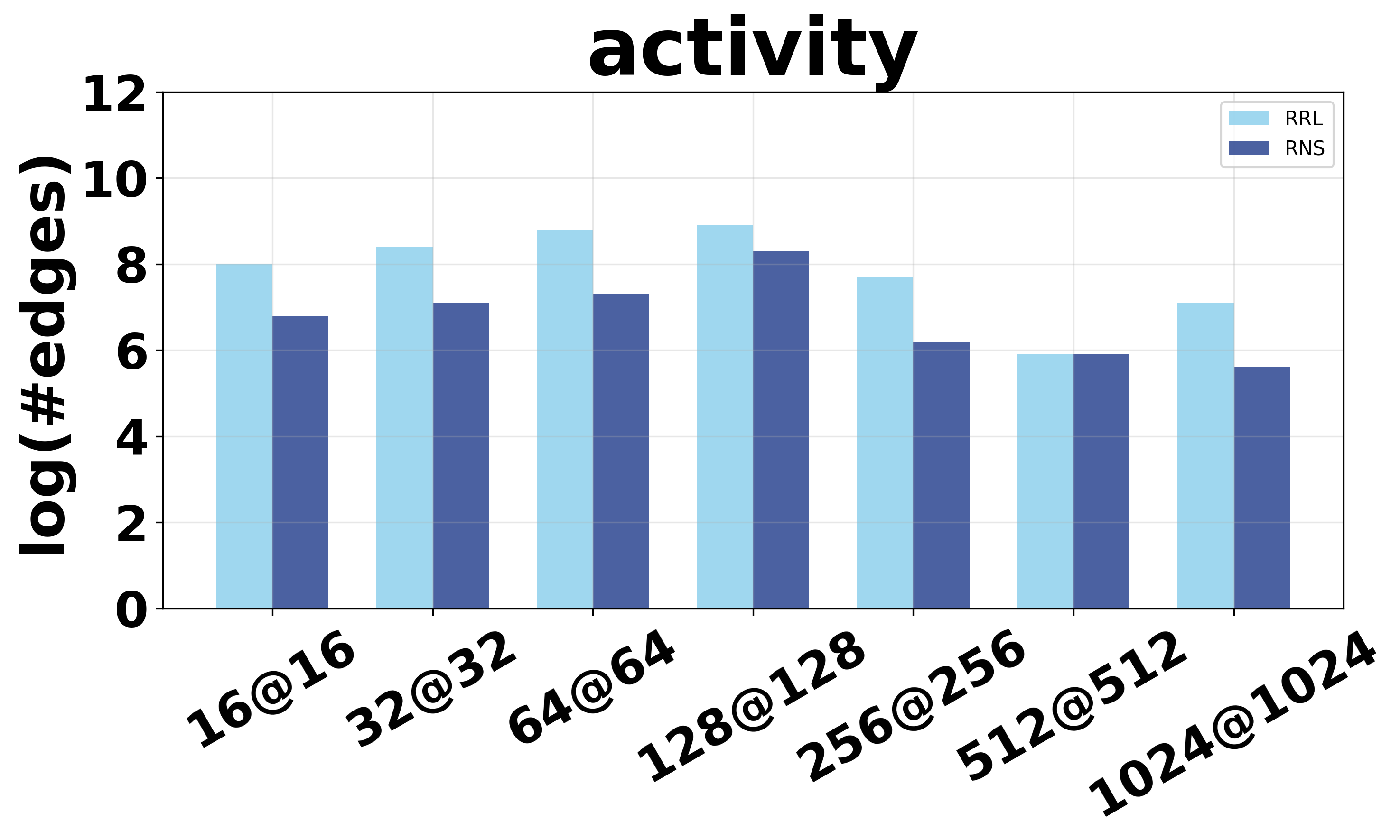}
    \label{fig:activity_length}
  \end{minipage}%
  \hfill%
  \begin{minipage}[t]{0.20\linewidth}
    \includegraphics[width=\linewidth]{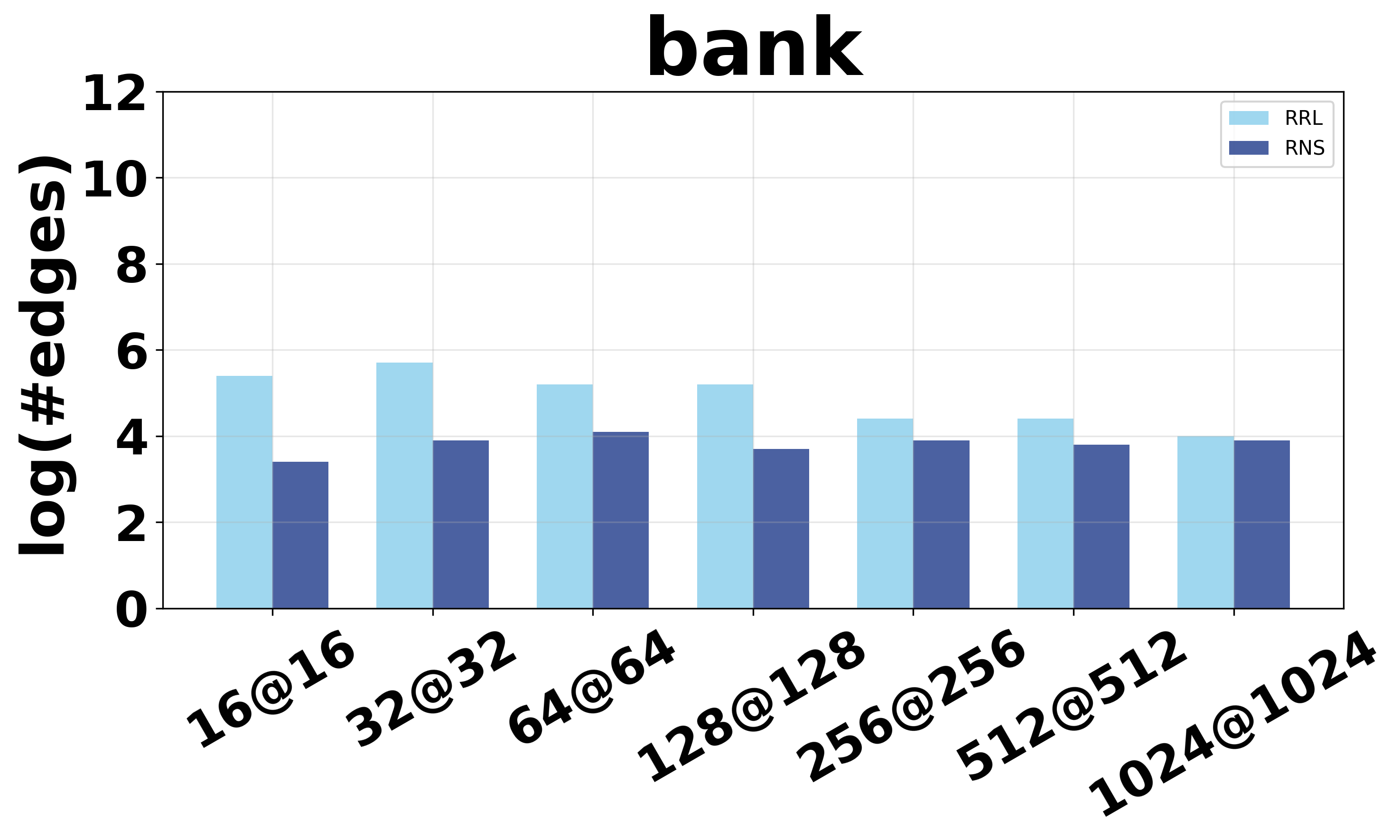}
    \label{fig:bank_length}
  \end{minipage}%
  \hfill%
  \begin{minipage}[t]{0.20\linewidth}
    \includegraphics[width=\linewidth]{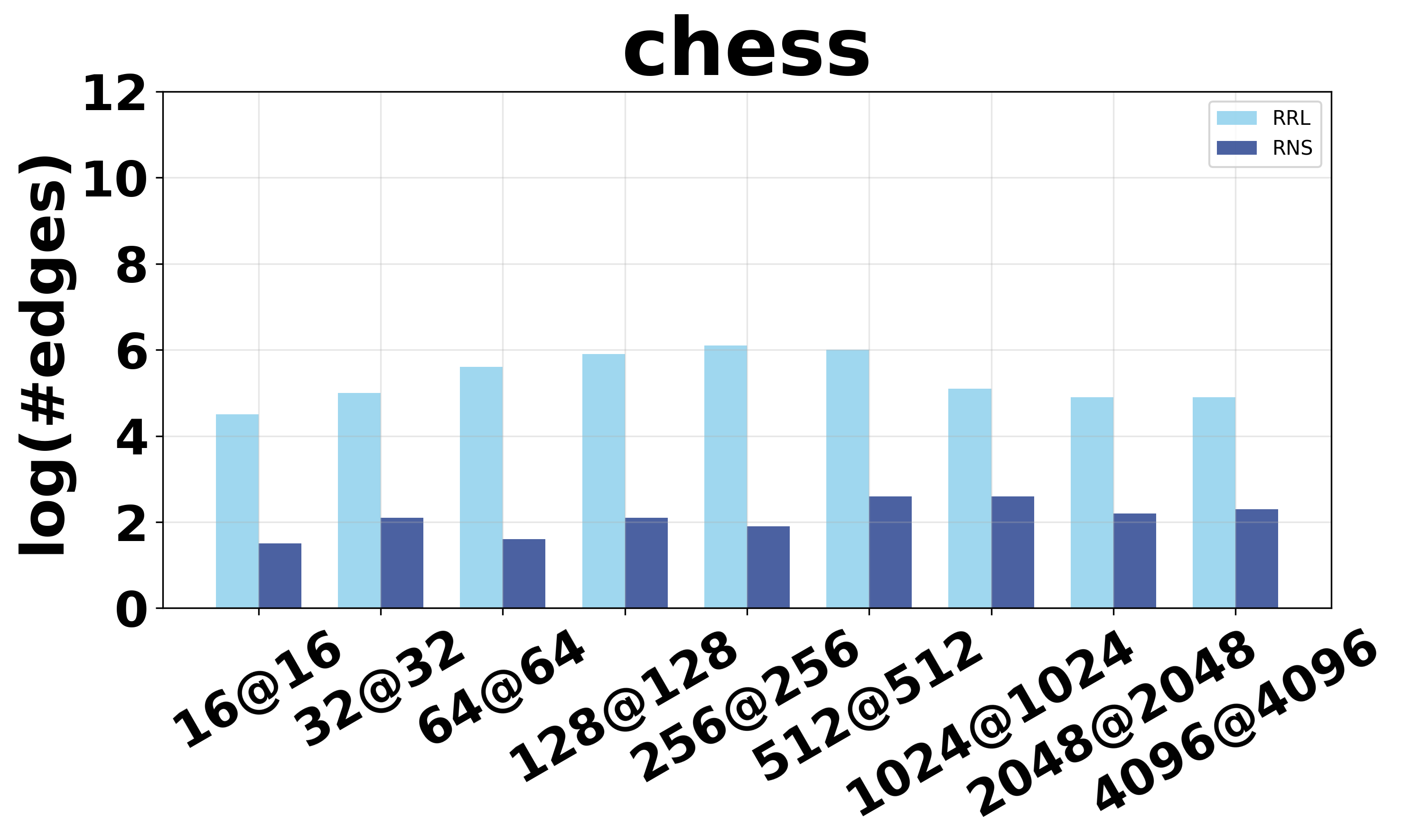}
    \label{fig:chess_length}
  \end{minipage}%
  \hfill%
  \begin{minipage}[t]{0.20\linewidth}
    \includegraphics[width=\linewidth]{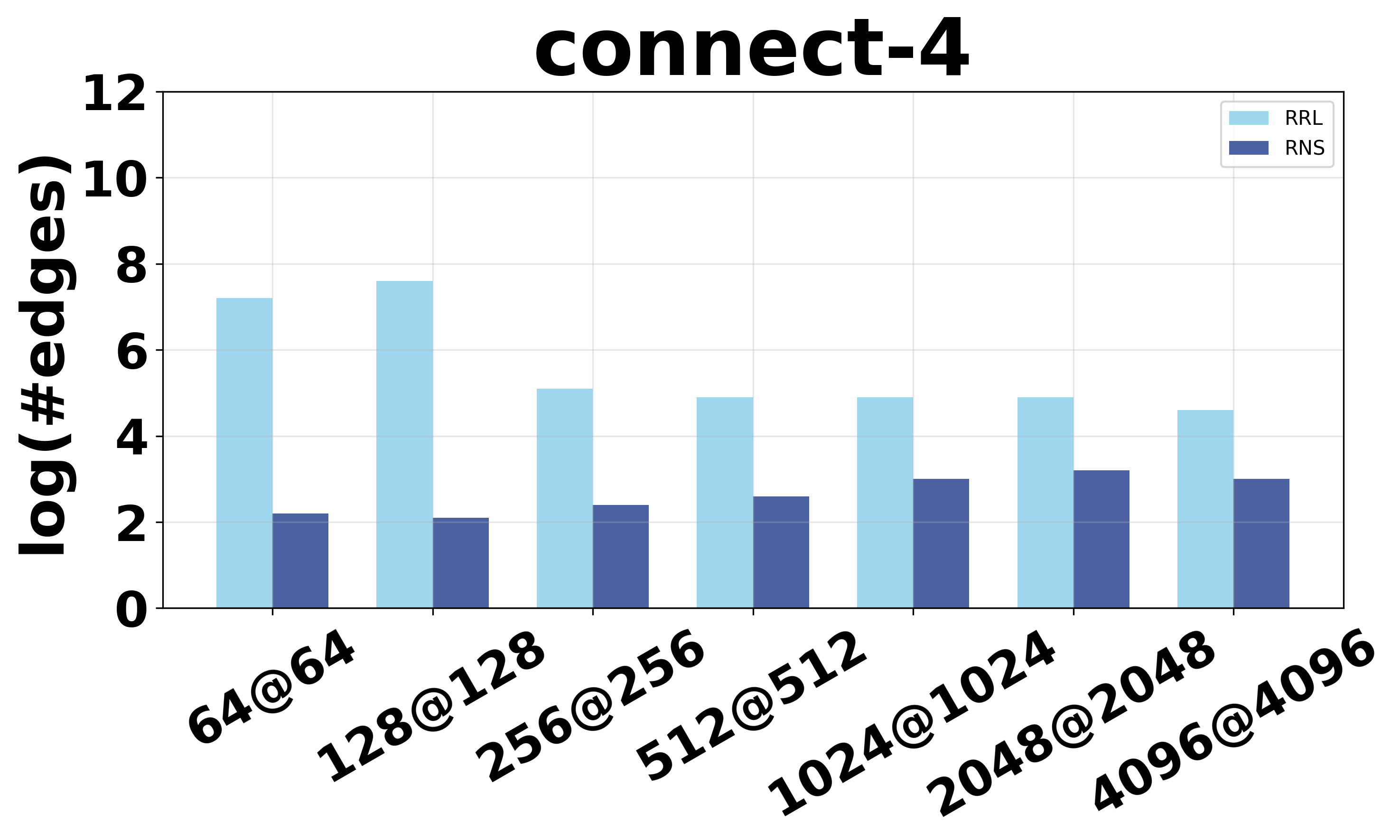}
    \label{fig:connect-4_length}
  \end{minipage}

  \begin{minipage}[t]{0.20\linewidth}
    \includegraphics[width=\linewidth]{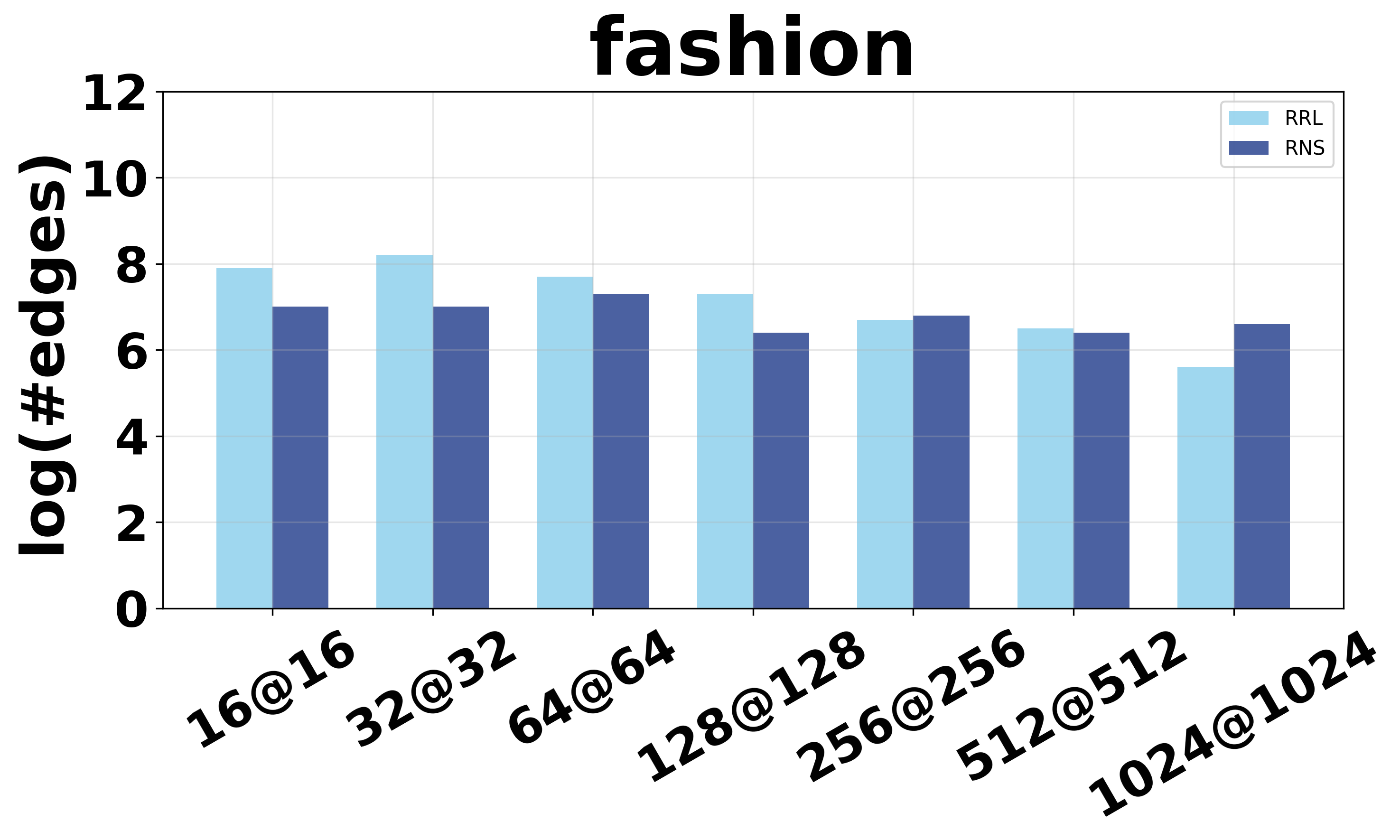}
    \label{fig:fashion_length}
  \end{minipage}%
  \hfill%
  \begin{minipage}[t]{0.20\linewidth}
    \includegraphics[width=\linewidth]{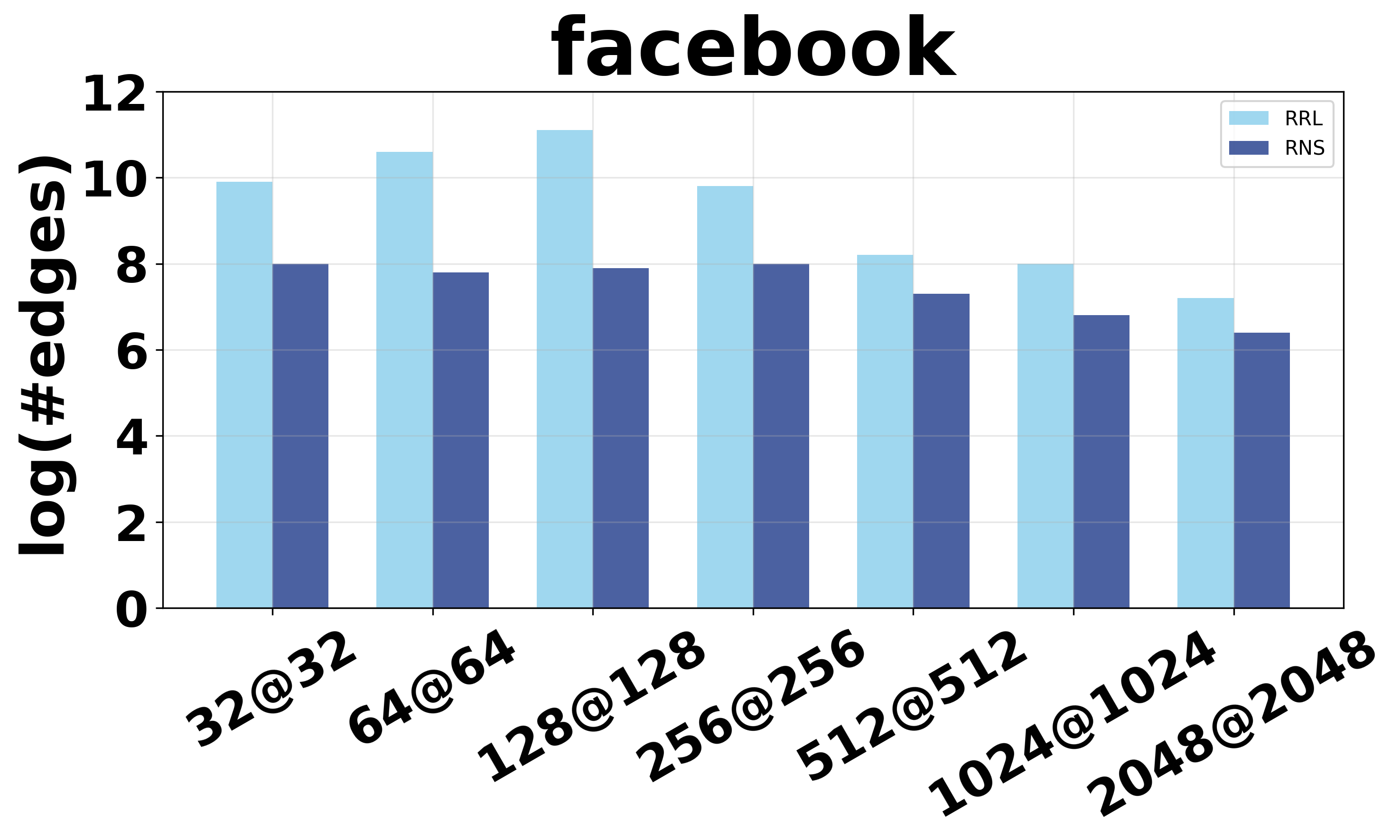}
    \label{fig:facebook_length}
  \end{minipage}%
  \hfill%
  \begin{minipage}[t]{0.20\linewidth}
    \includegraphics[width=\linewidth]{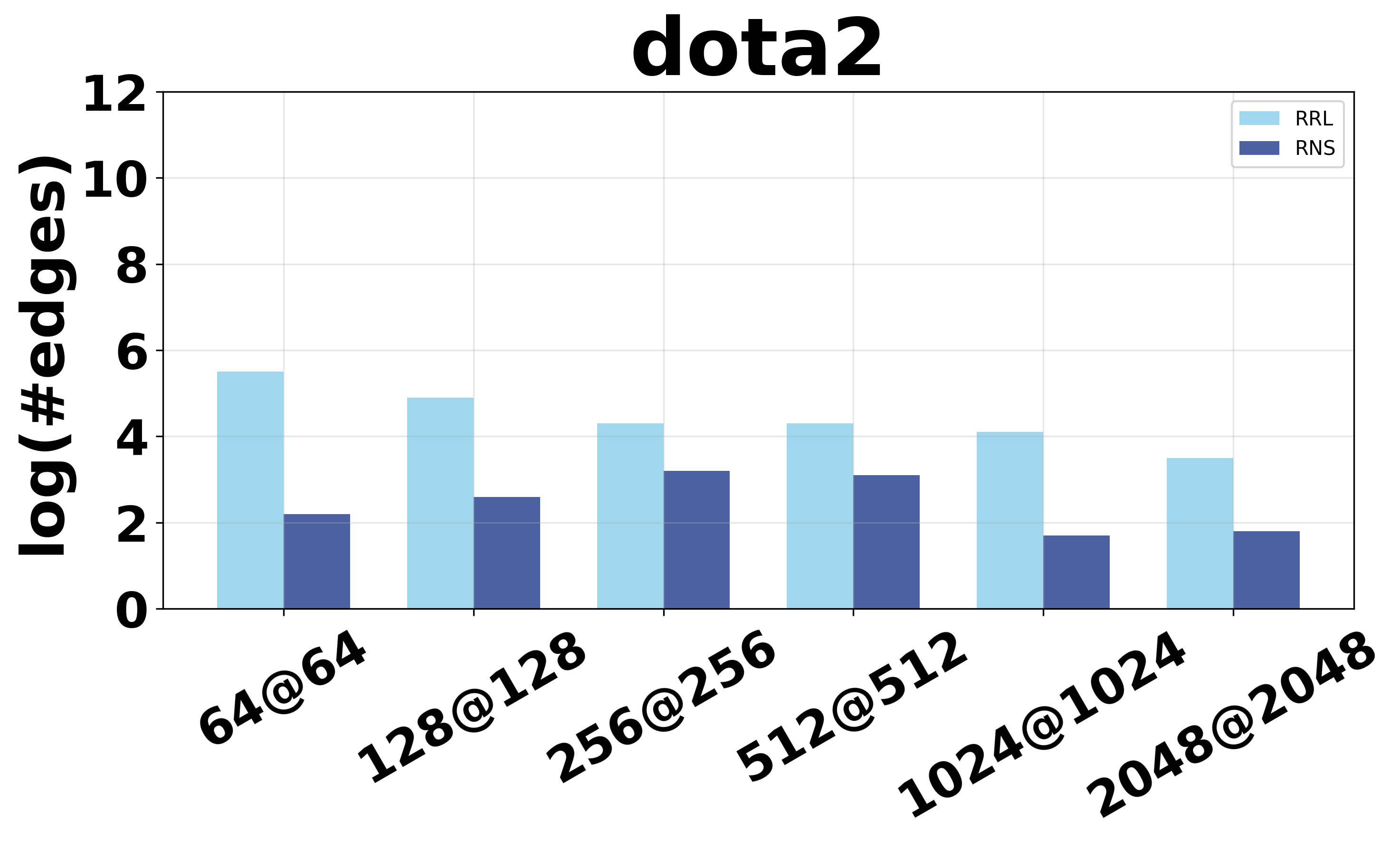}
    \label{fig:dota2_length}
  \end{minipage}%
  \hfill%
  \begin{minipage}[t]{0.20\linewidth}
    \includegraphics[width=\linewidth]{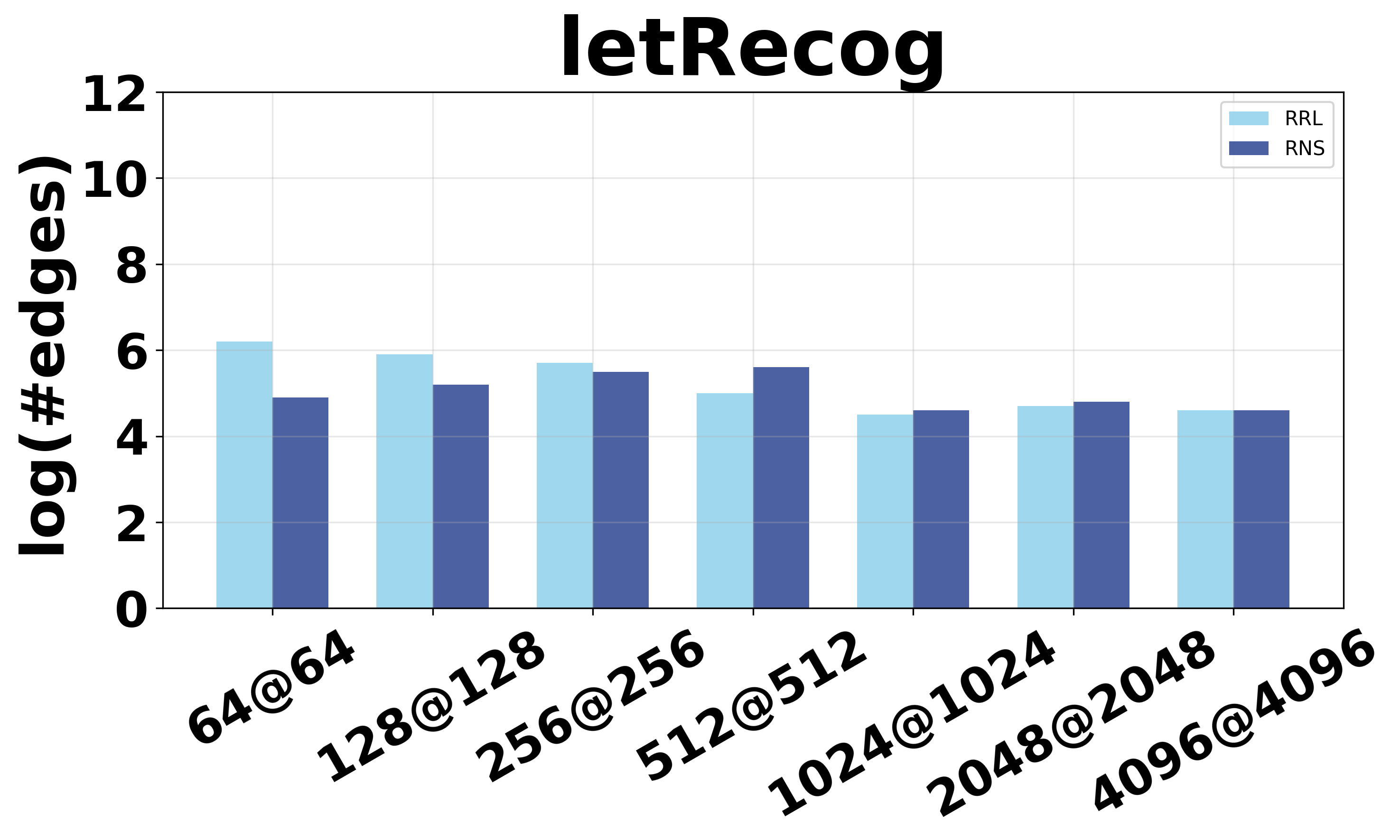}
    \label{fig:letRecog_length}
  \end{minipage}%
  \hfill%
  \begin{minipage}[t]{0.20\linewidth}
    \includegraphics[width=\linewidth]{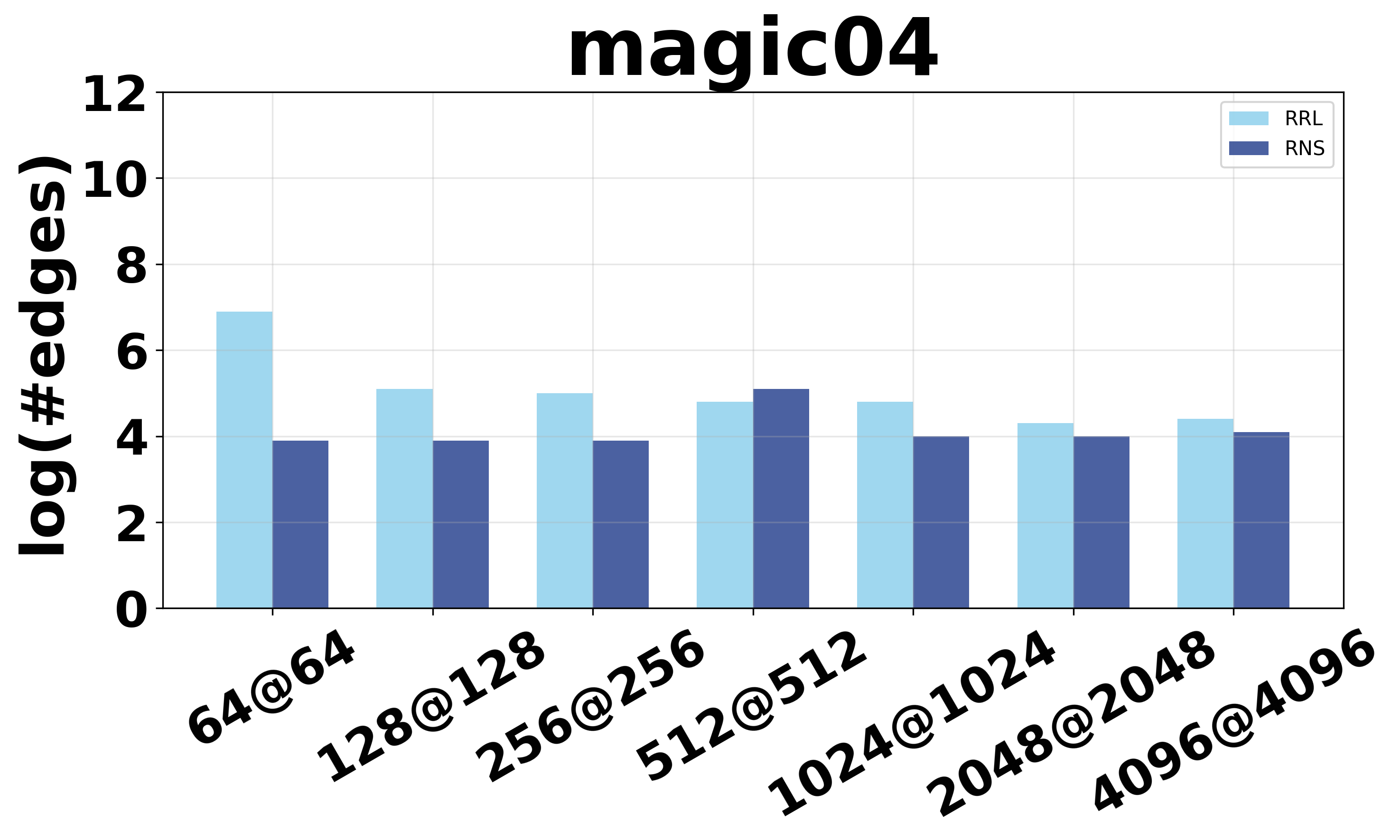}
    \label{fig:magic04_length}
  \end{minipage}

  \caption{Rule Length comparison across different architectures.}
  \label{fig:rule_length_all}
\end{figure*}

We evaluate learning efficiency by the number and length of learned rules across all datasets, as shown in Figure~\ref{fig:rule_number_all} and Figure~\ref{fig:rule_length_all}, respectively. Figure~\ref{fig:rule_number_all} presents scatter plots of F1 score against log(\#edges) across 10 datasets. The boundary connecting the results of different RNS architectures separates the upper left corner from the best baseline methods. This indicates that RNS consistently learns fewer rules while achieving high prediction accuracy across various scenarios.
The average length of rules in RNS trained on different datasets is shown in Figure~\ref{fig:rule_length_all}. The average rule length is less than 6 for all datasets except fashion and facebook, which are unstructured datasets and have more complex features. These results indicate that the rules learned by RNS are generally easy to understand across different scenarios.

\section{Rule Quality } \label{Quality all}

\textbf{Rule Quality Metrics.} Following prior work~\cite{10.1145/3583780.3614884,lakkaraju2016interpretable}, we evaluate rule quality using three key metrics: diversity, coverage, and single-rule accuracy. Let $I_r$ denote the set of data instances covered by rule $r$, and $D$ denote the complete dataset. The metrics are defined as:

\textbf{Single-rule Accuracy} measures the prediction accuracy of a single rule for the instances it covers:
\begin{equation}
\text{Single-rule Accuracy} = \frac{|\{i \in I_r : \text{rule prediction matches } y_i\}|}{|I_r|}
\end{equation}
This metric evaluates how accurately a single rule classifies the data instances it covers when used independently for prediction. It measures the proportion of correctly classified instances among all instances that satisfy the rule's conditions. Higher single-rule accuracy indicates that the rule is more reliable for making predictions on its own, without relying on other rules in the rule set.
\textbf{Coverage} quantifies the proportion of data instances covered by a rule:
\begin{equation}
\text{Coverage} = \frac{|I_r|}{|D|}
\end{equation}
This metric measures the scope or generality of a rule. Lower coverage indicates that rules are more specific and easier for human experts to understand, as they apply to a smaller, more focused subset of the data. Very high coverage may suggest overly general rules that lack specificity.

\textbf{Diversity} measures the overlap ratio between pairs of rules, with higher diversity reflecting that rules capture distinct, non-redundant logic:
\begin{equation}
\text{Diversity} = 1 - \frac{|I_i \cap I_j|}{|I_i \cup I_j|}
\end{equation}
This metric quantifies how different rules are from each other by measuring their overlap. Higher diversity values indicate that rules capture different patterns in the data with minimal redundancy, leading to a more comprehensive and non-redundant rule set. Low diversity suggests that multiple rules are covering similar data instances, which reduces the interpretability and efficiency of the rule set.

\begin{figure*}[ht]
  \centering
  \begin{minipage}[t]{0.20\linewidth}
    \includegraphics[width=\linewidth]{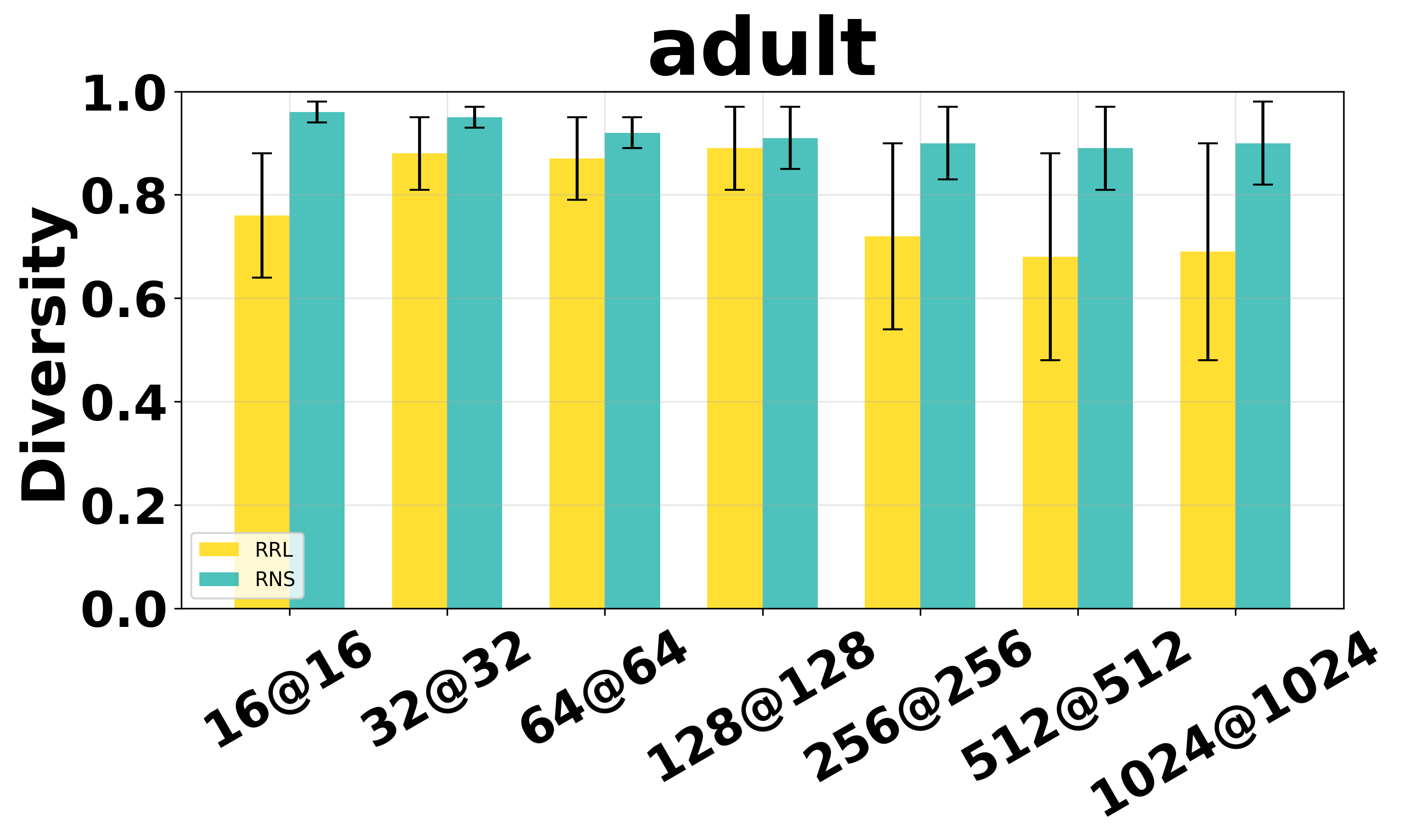}

  \end{minipage}%
  \hfill%
  \begin{minipage}[t]{0.20\linewidth}
    \includegraphics[width=\linewidth]{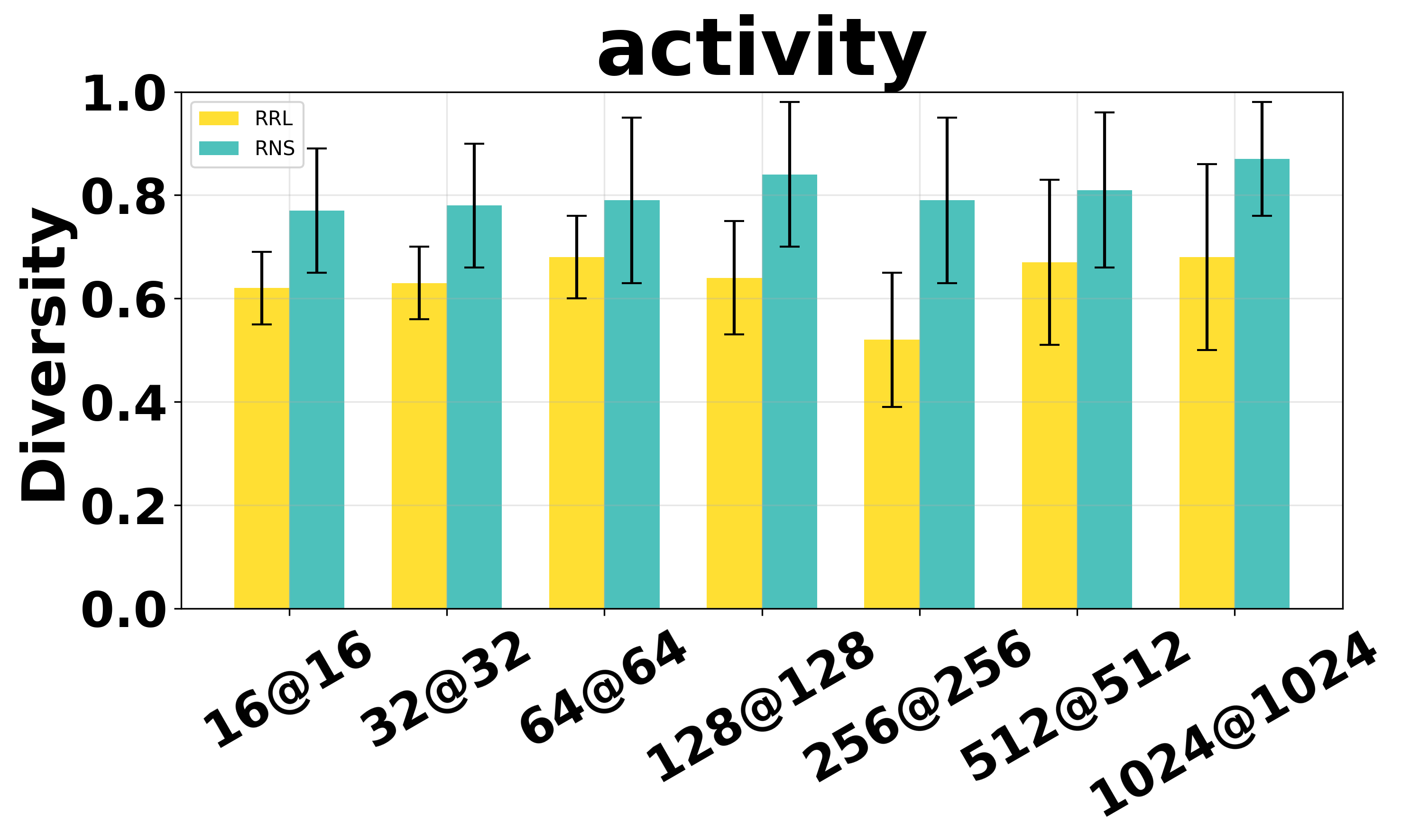}

  \end{minipage}%
  \hfill%
  \begin{minipage}[t]{0.20\linewidth}
    \includegraphics[width=\linewidth]{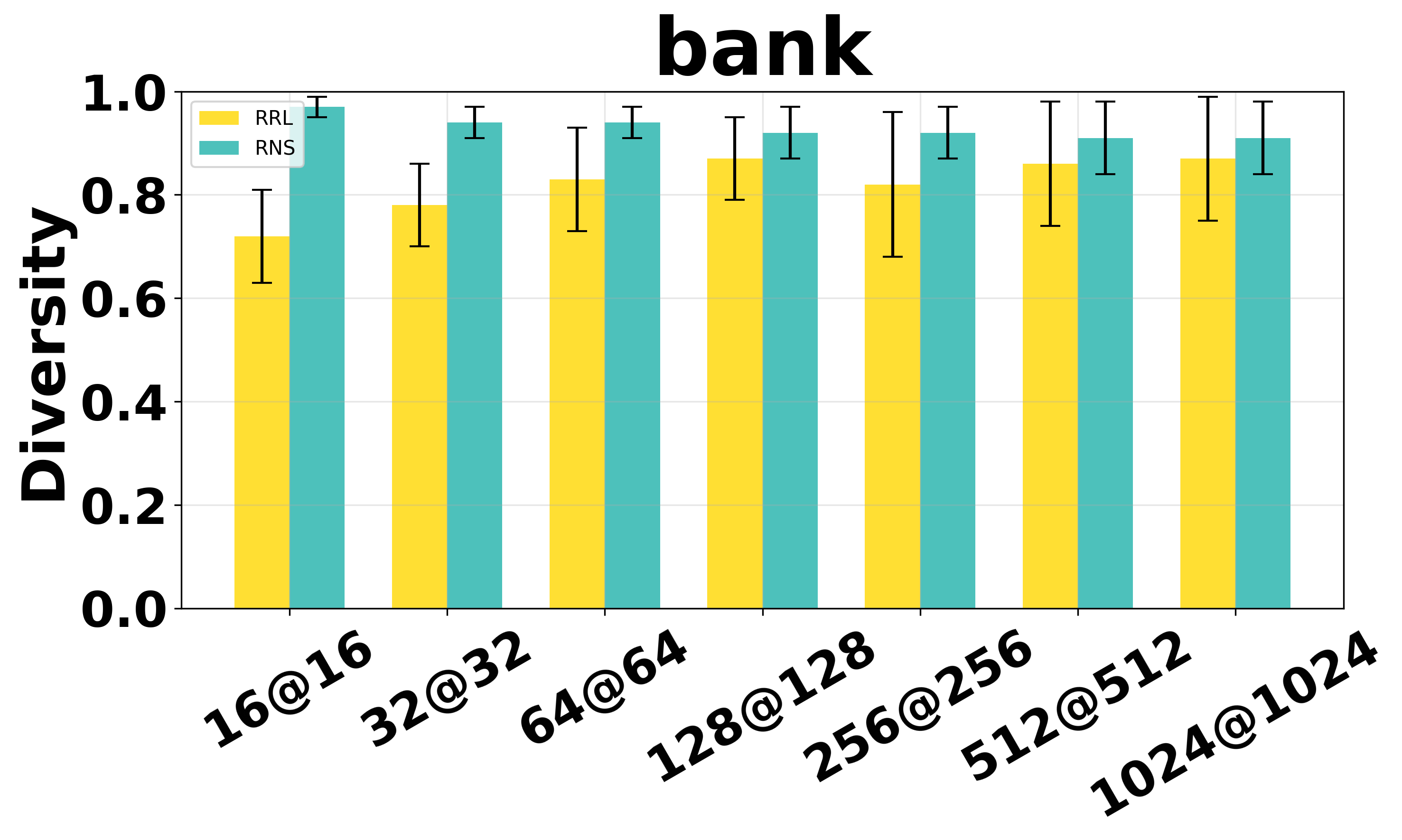}

  \end{minipage}%
  \hfill%
  \begin{minipage}[t]{0.20\linewidth}
    \includegraphics[width=\linewidth]{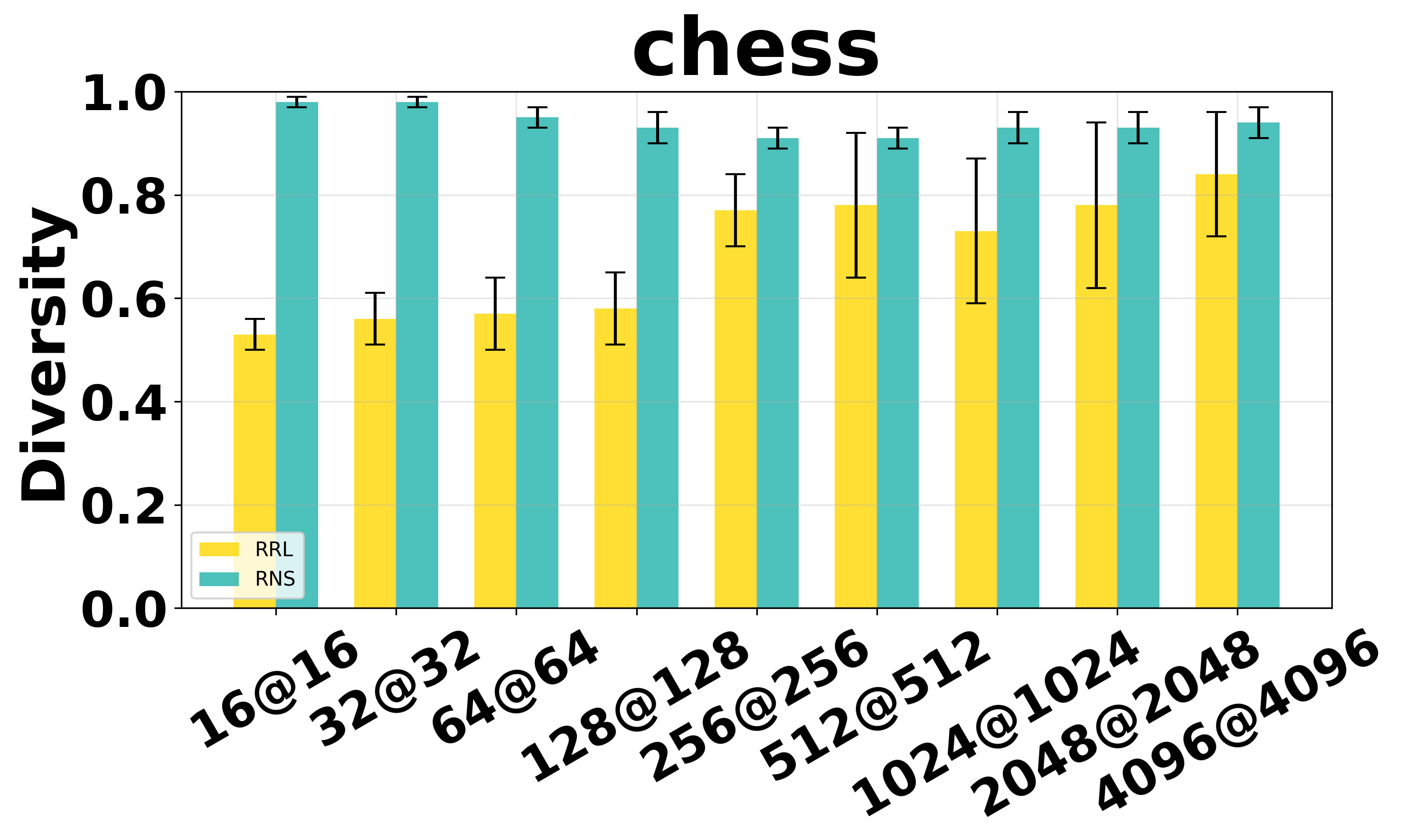}

  \end{minipage}%
  \hfill%
  \begin{minipage}[t]{0.20\linewidth}
    \includegraphics[width=\linewidth]{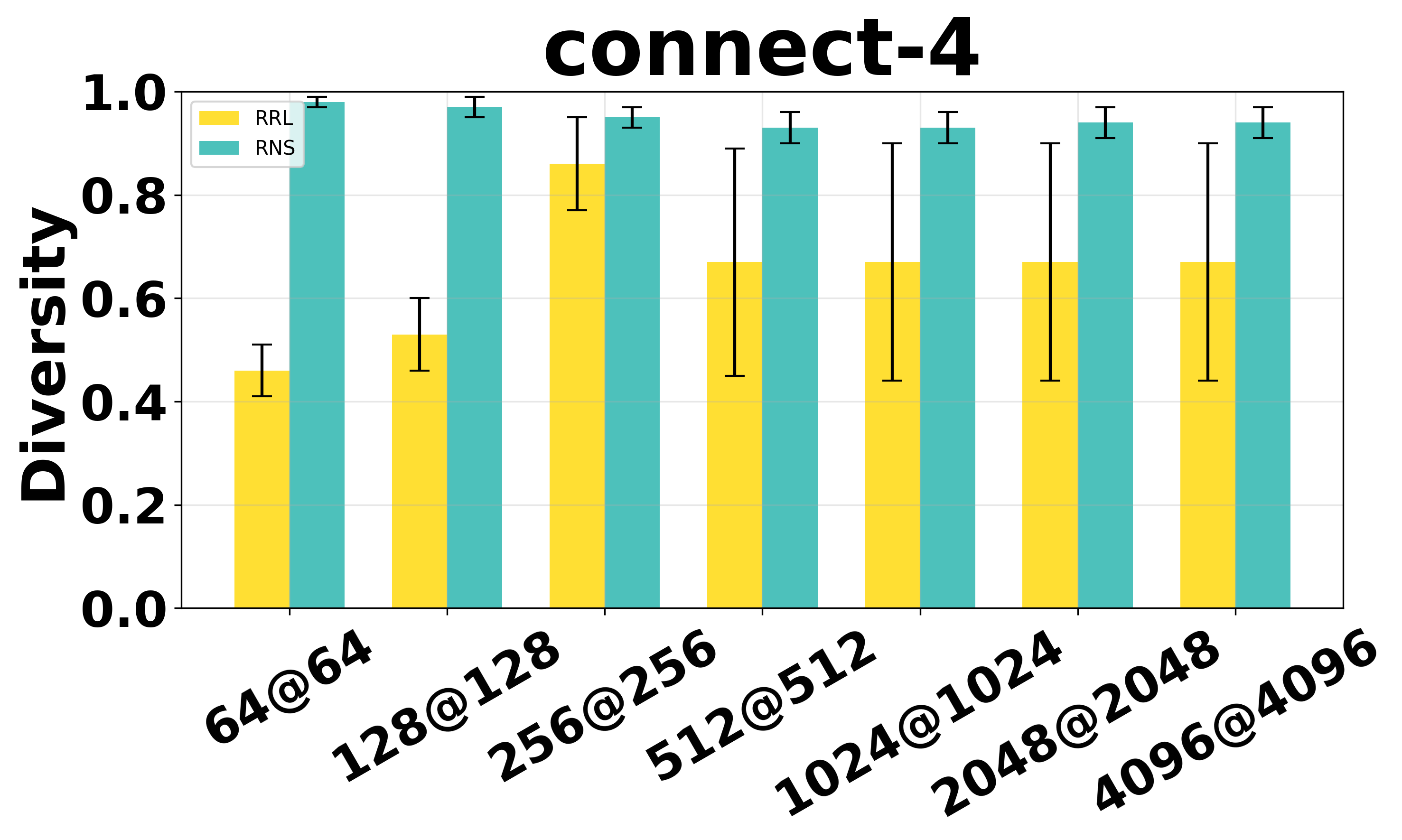}

  \end{minipage}

  \begin{minipage}[t]{0.20\linewidth}
    \includegraphics[width=\linewidth]{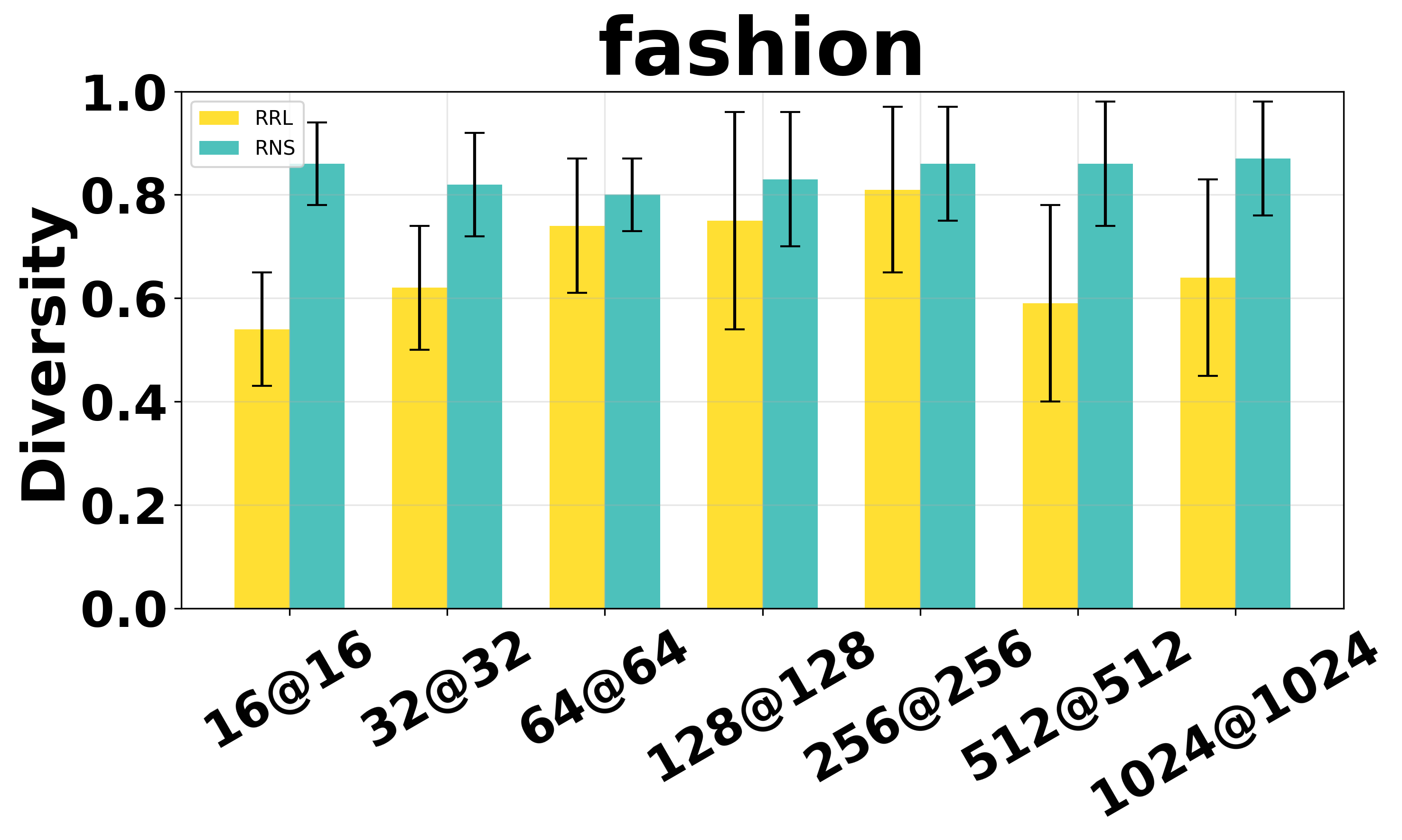}

  \end{minipage}%
  \hfill%
  \begin{minipage}[t]{0.20\linewidth}
    \includegraphics[width=\linewidth]{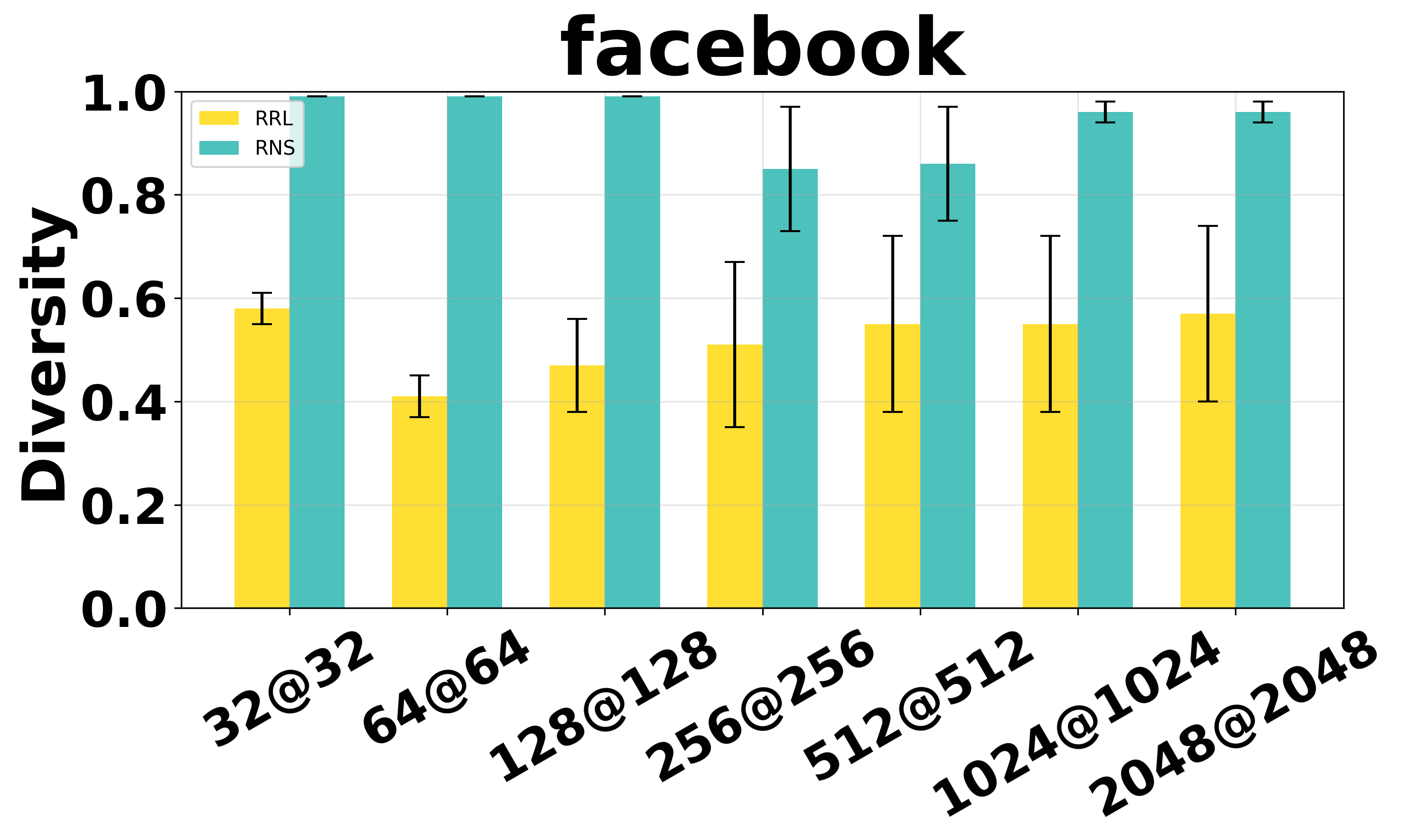}

  \end{minipage}%
  \hfill%
  \begin{minipage}[t]{0.20\linewidth}
    \includegraphics[width=\linewidth]{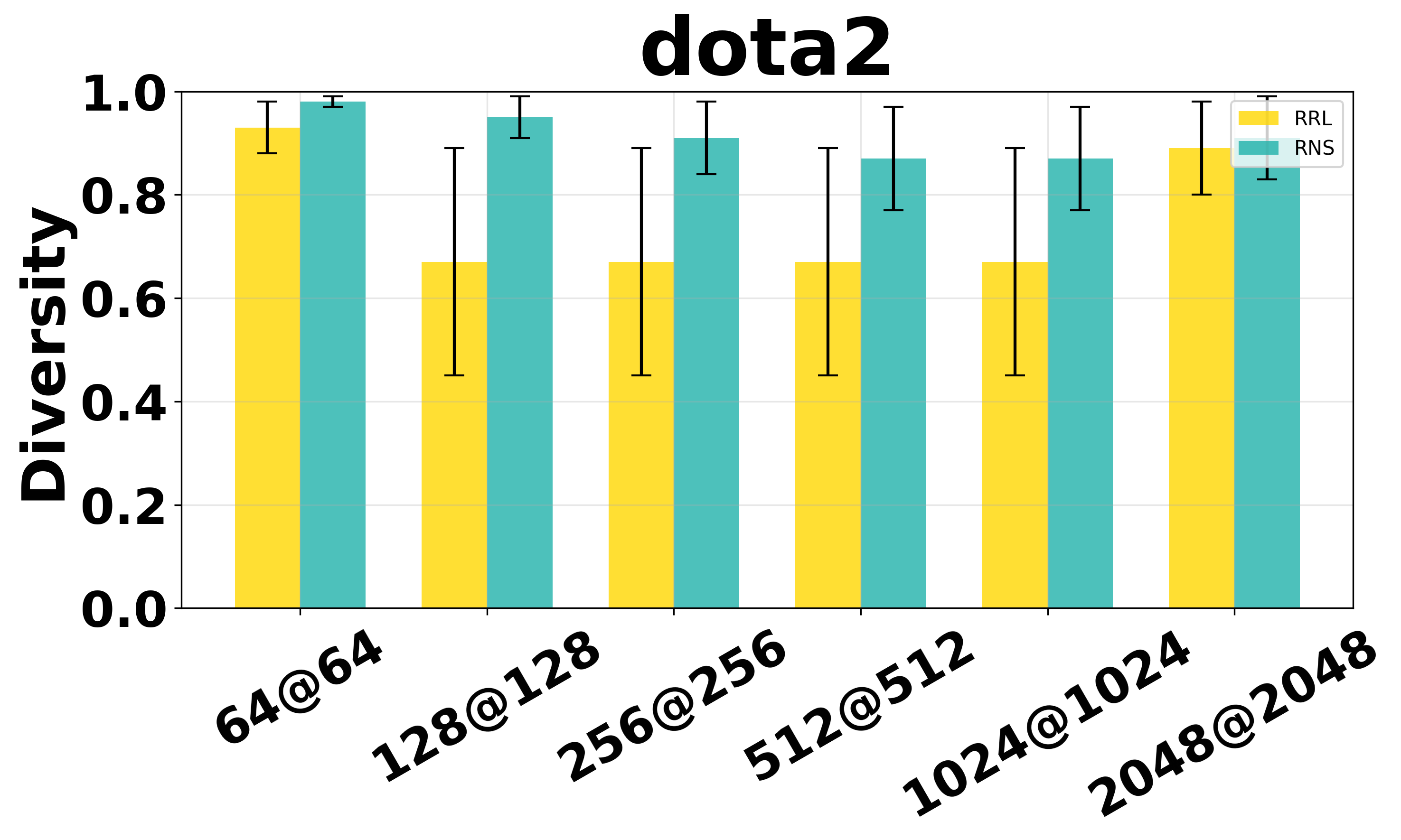}

  \end{minipage}%
  \hfill%
  \begin{minipage}[t]{0.20\linewidth}
    \includegraphics[width=\linewidth]{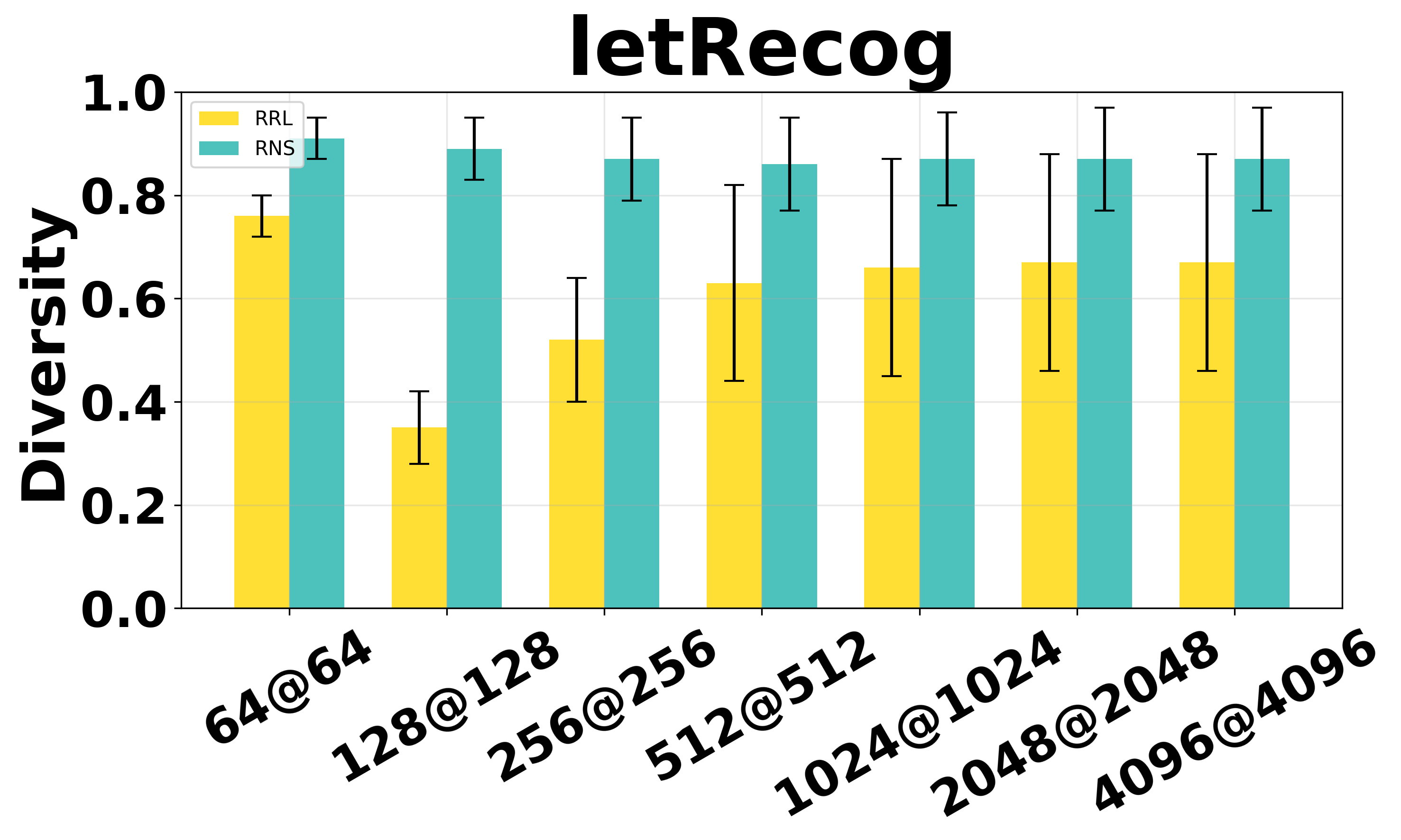}

  \end{minipage}%
  \hfill%
  \begin{minipage}[t]{0.20\linewidth}
    \includegraphics[width=\linewidth]{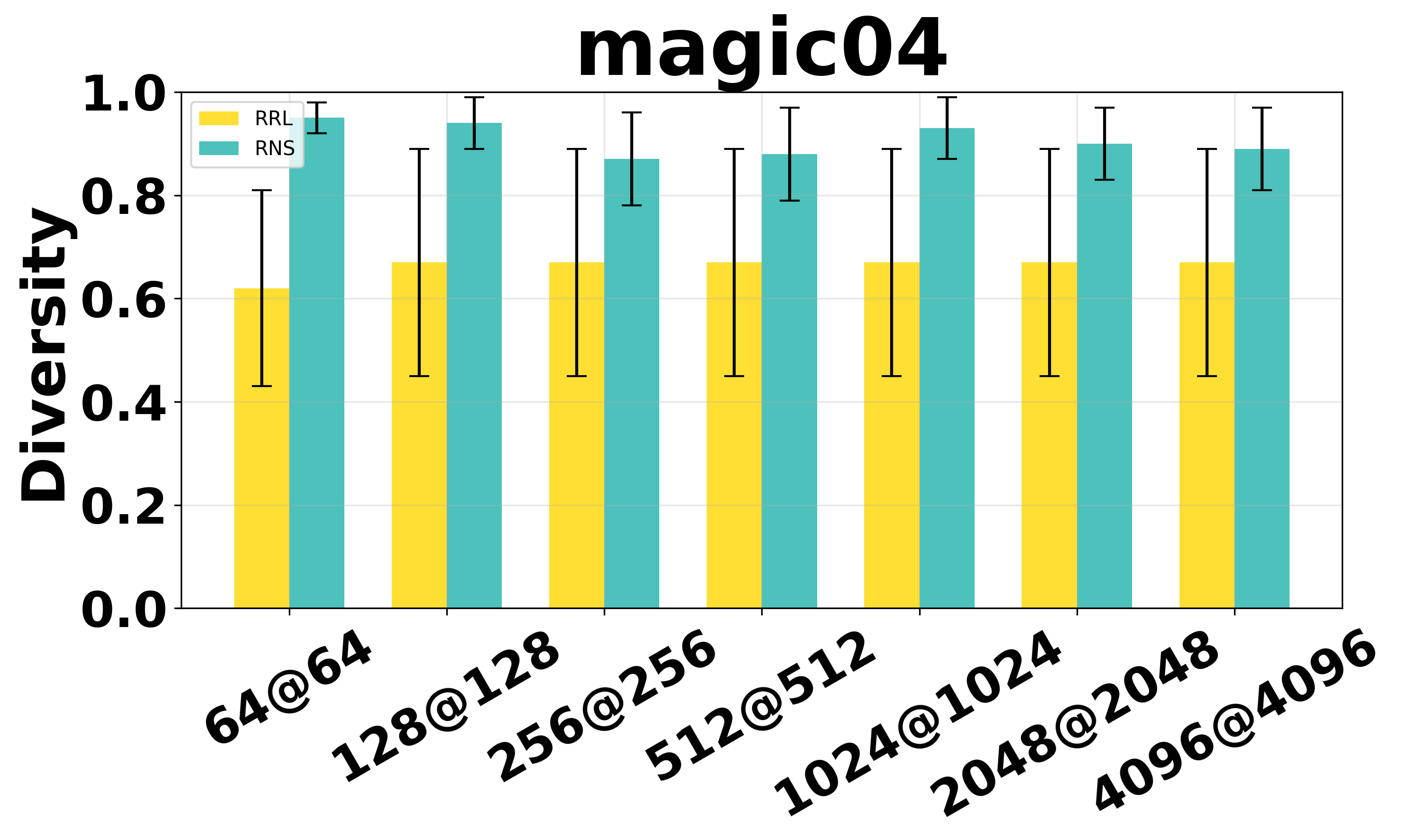}

  \end{minipage}

  \caption{Rule Diversity comparison across different architectures.}
  \label{fig:diversity_all}
\end{figure*}
\begin{figure*}[ht]
  \centering
  \begin{minipage}[t]{0.20\linewidth}
    \includegraphics[width=\linewidth]{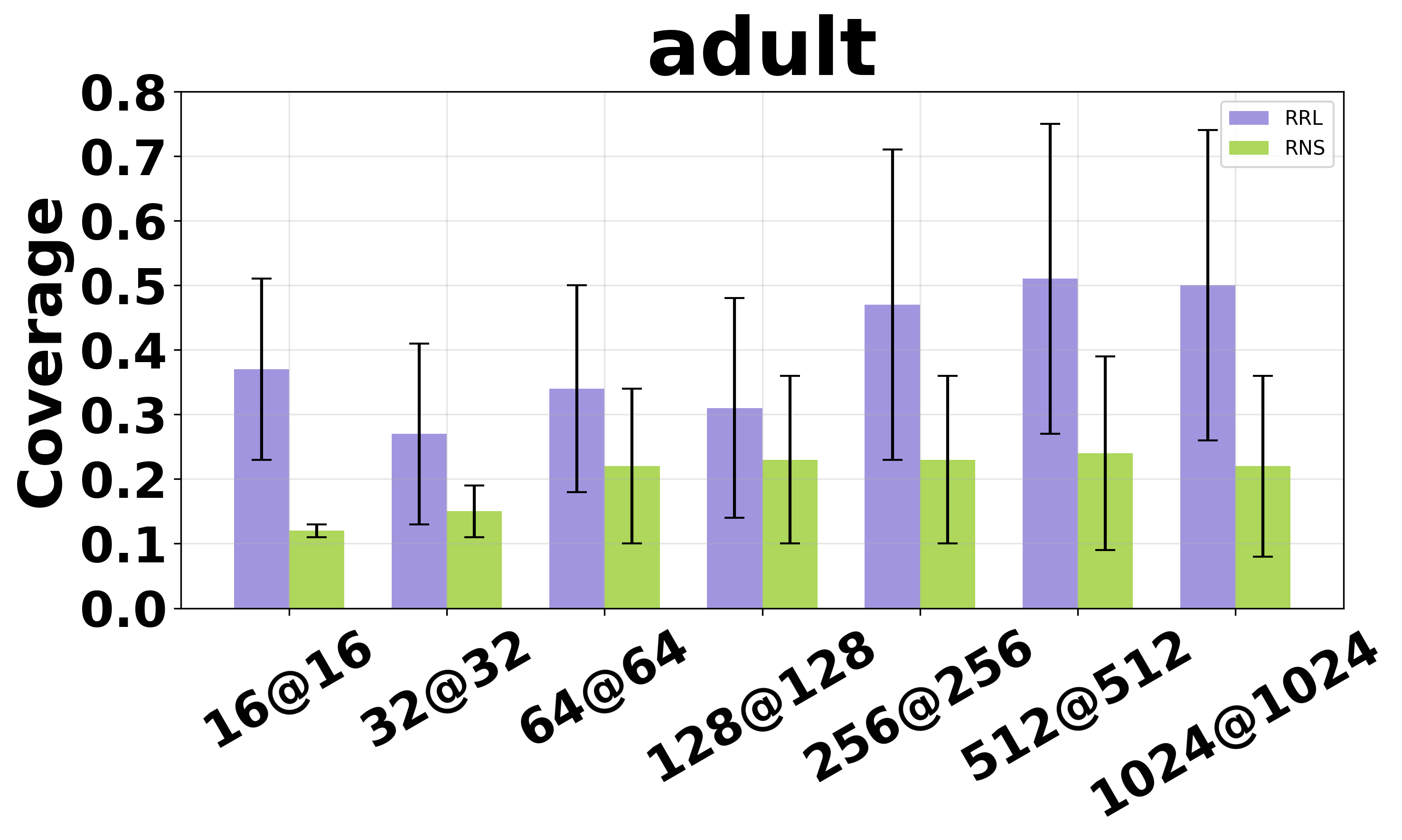}

  \end{minipage}%
  \hfill%
  \begin{minipage}[t]{0.20\linewidth}
    \includegraphics[width=\linewidth]{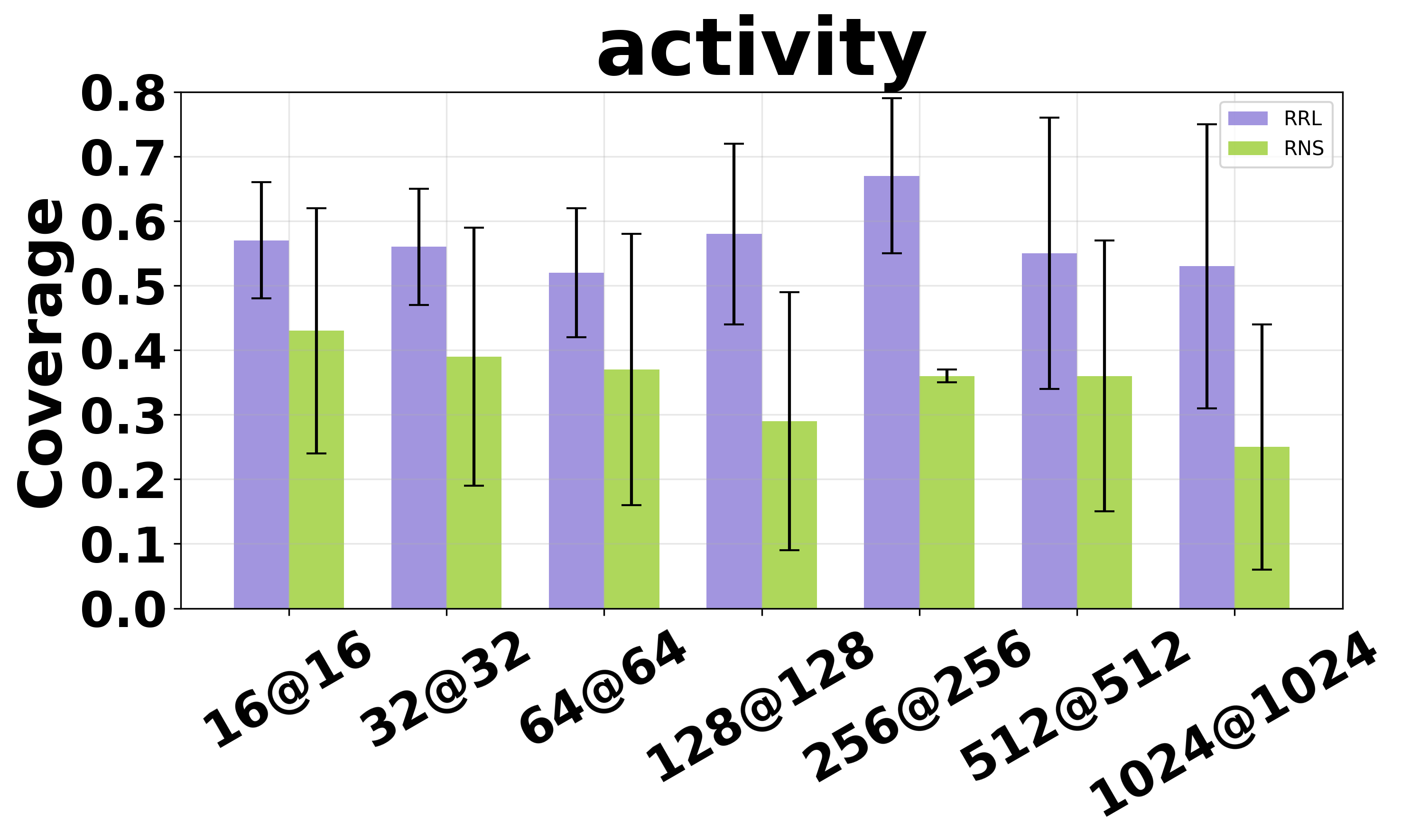}

  \end{minipage}%
  \hfill%
  \begin{minipage}[t]{0.20\linewidth}
    \includegraphics[width=\linewidth]{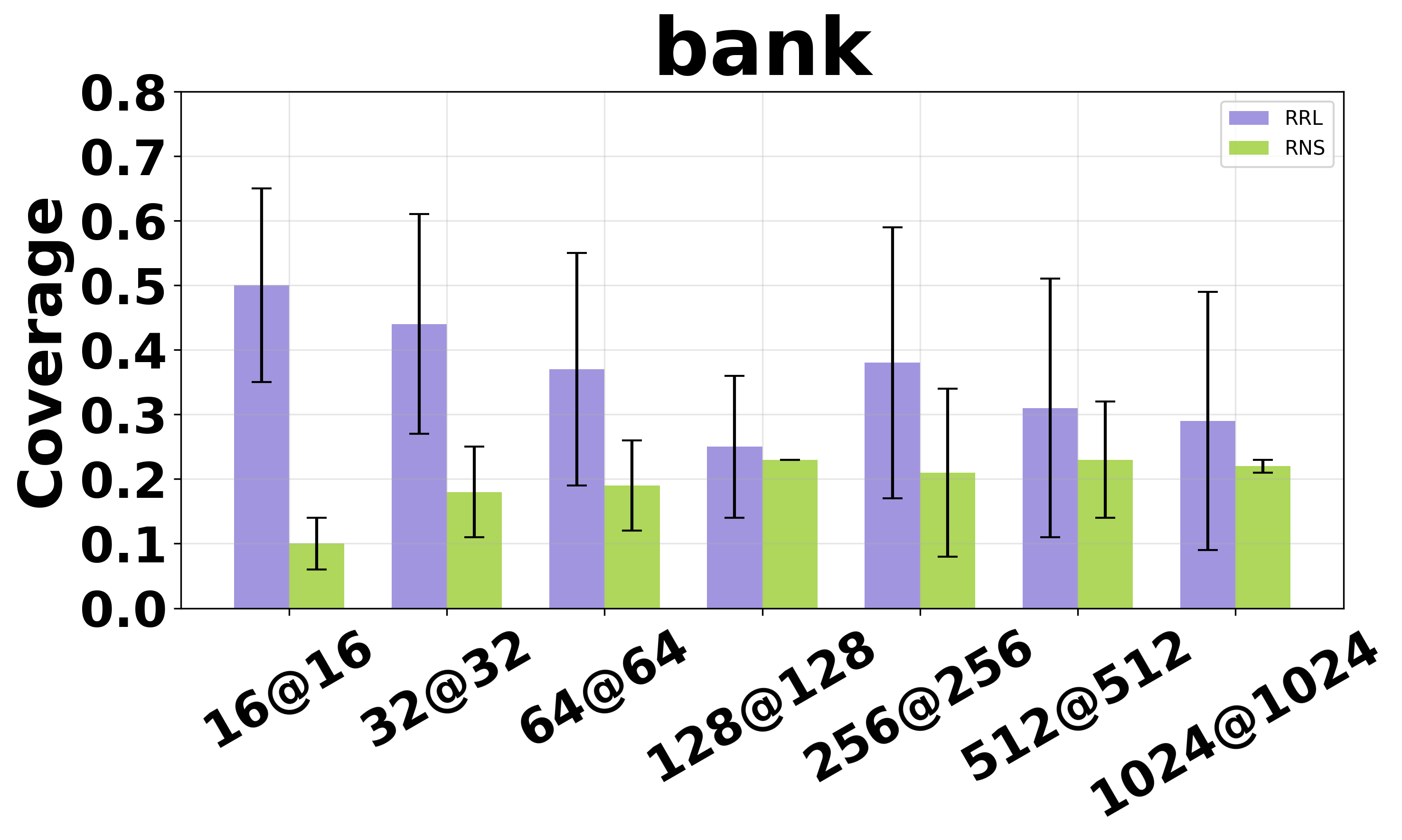}

  \end{minipage}%
  \hfill%
  \begin{minipage}[t]{0.20\linewidth}
    \includegraphics[width=\linewidth]{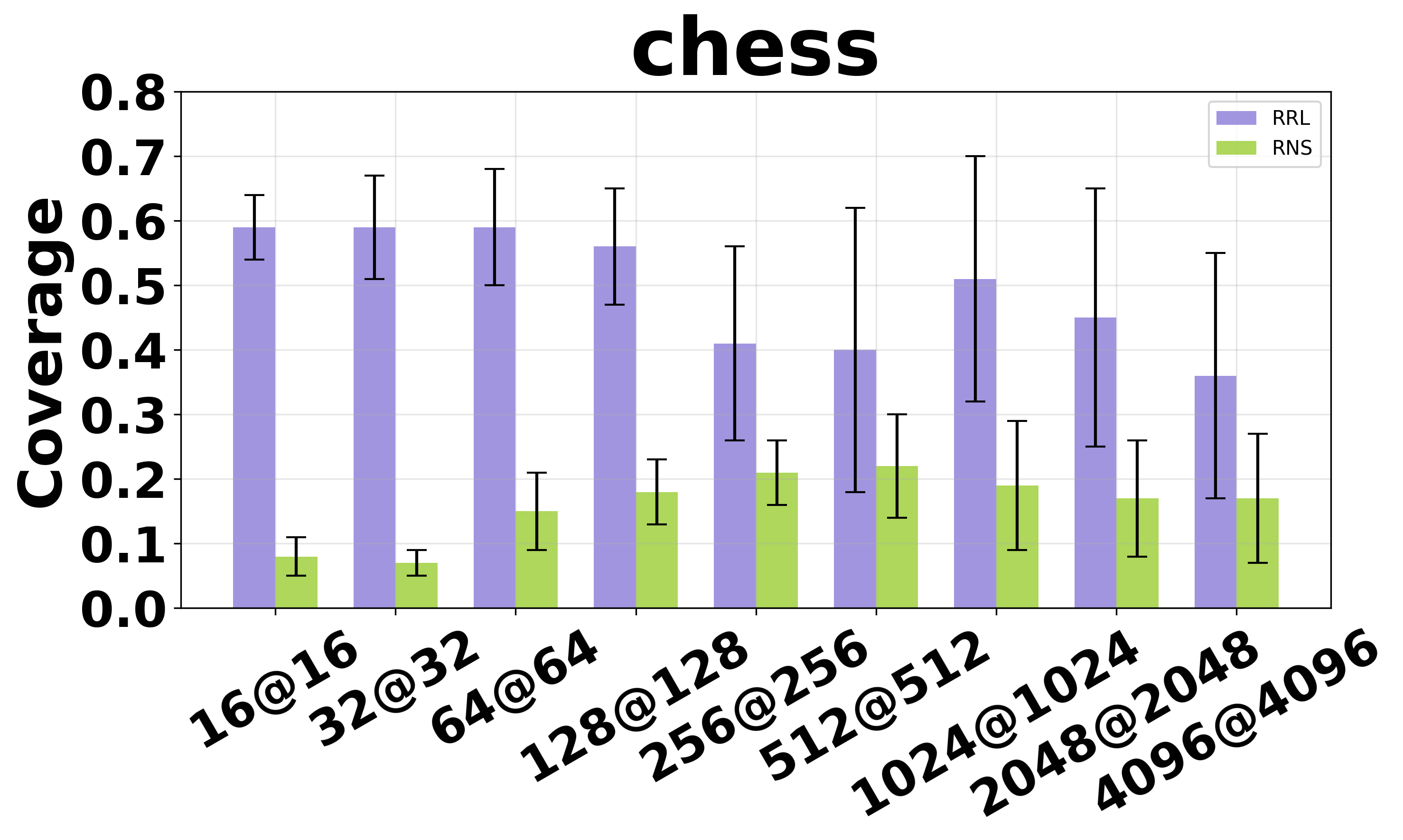}

  \end{minipage}%
  \hfill%
  \begin{minipage}[t]{0.20\linewidth}
    \includegraphics[width=\linewidth]{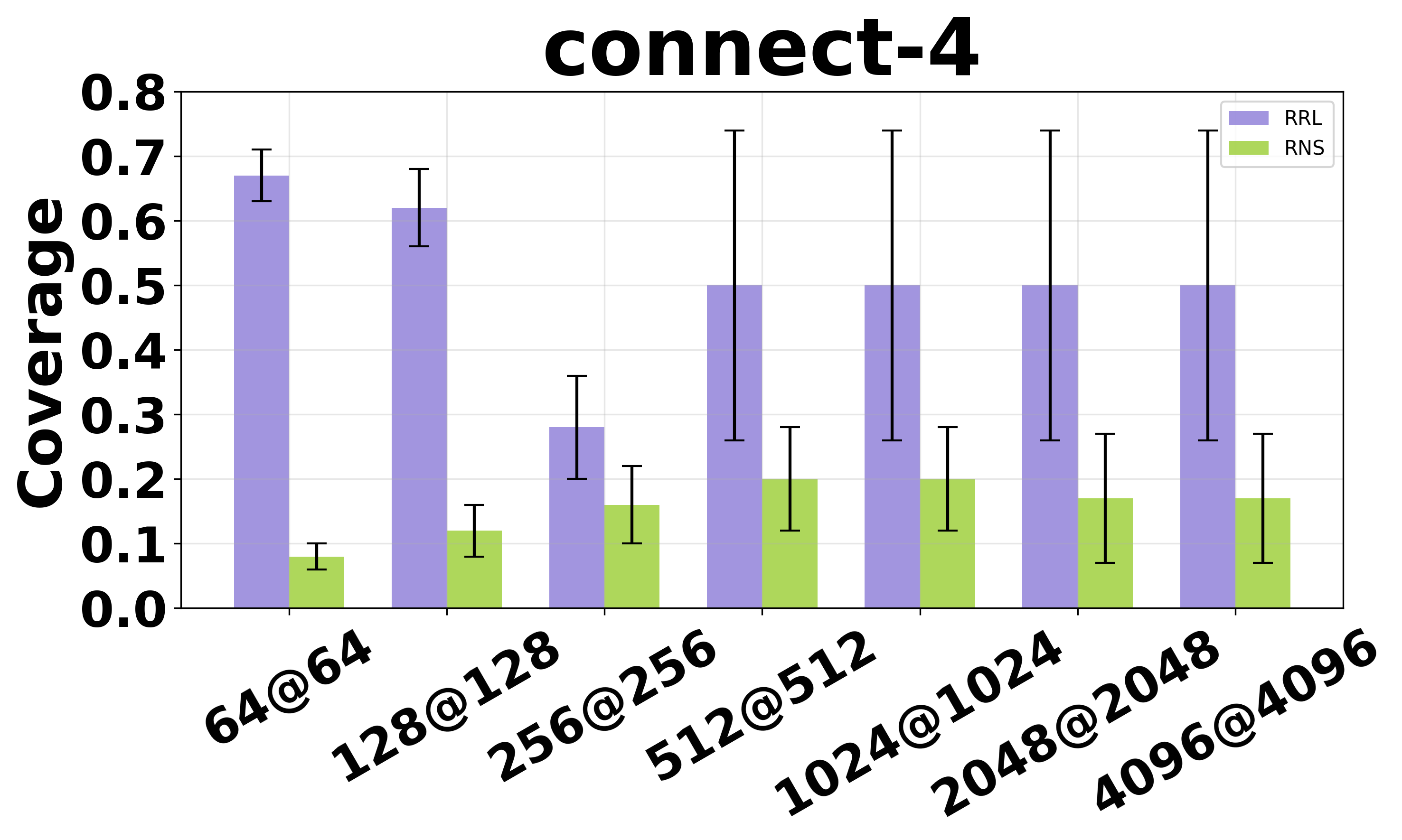}

  \end{minipage}

  \begin{minipage}[t]{0.20\linewidth}
    \includegraphics[width=\linewidth]{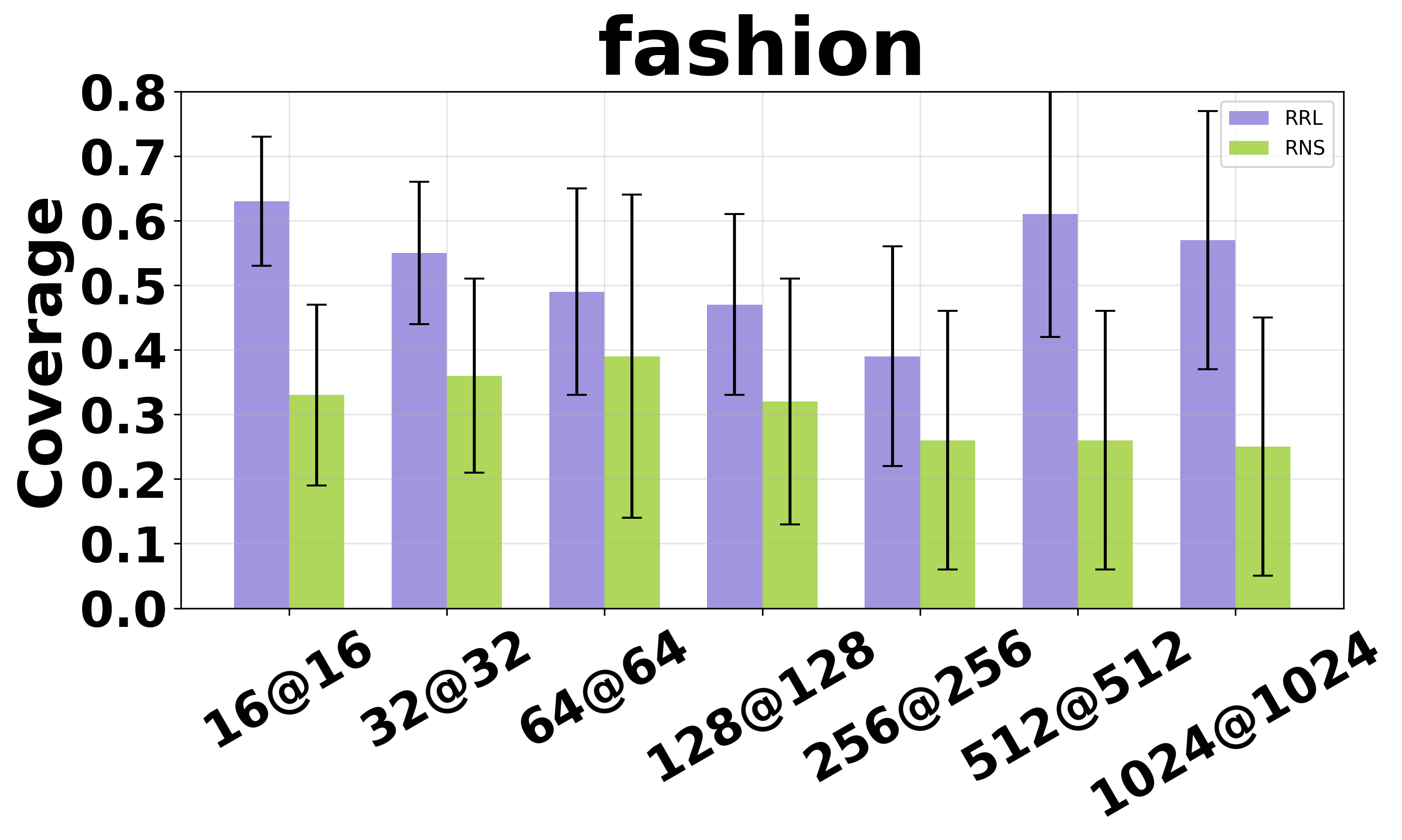}

  \end{minipage}%
  \hfill%
  \begin{minipage}[t]{0.20\linewidth}
    \includegraphics[width=\linewidth]{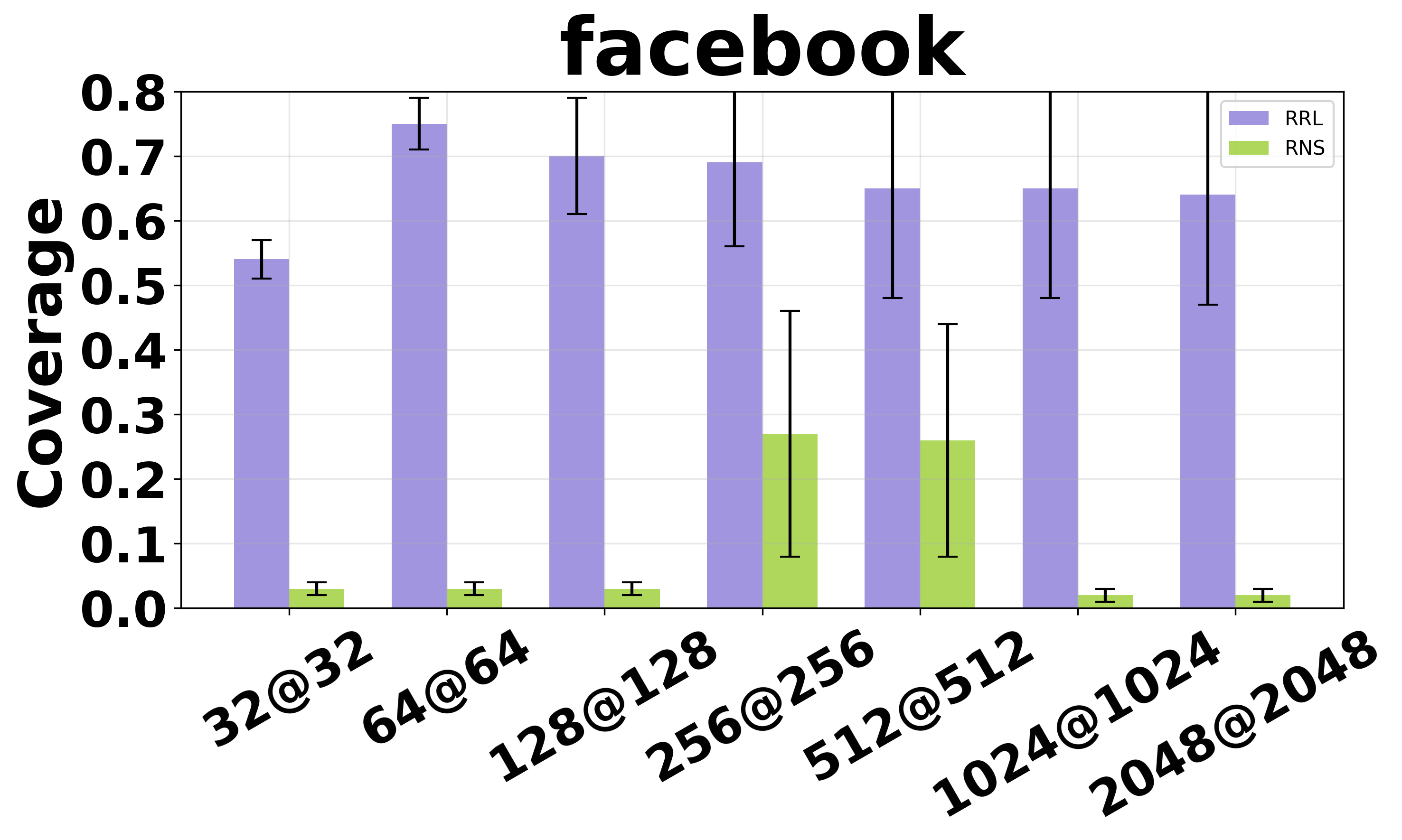}

  \end{minipage}%
  \hfill%
  \begin{minipage}[t]{0.20\linewidth}
    \includegraphics[width=\linewidth]{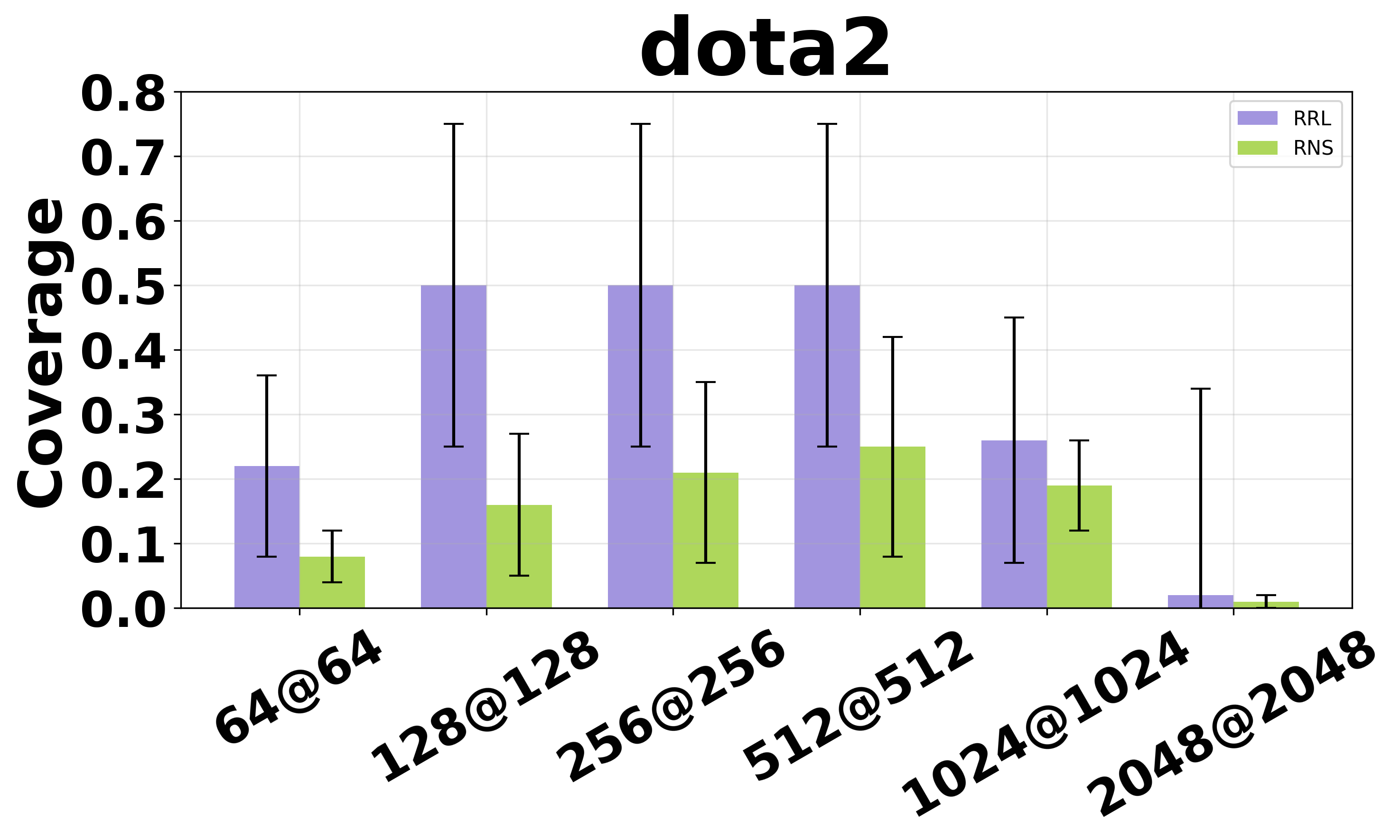}

  \end{minipage}%
  \hfill%
  \begin{minipage}[t]{0.20\linewidth}
    \includegraphics[width=\linewidth]{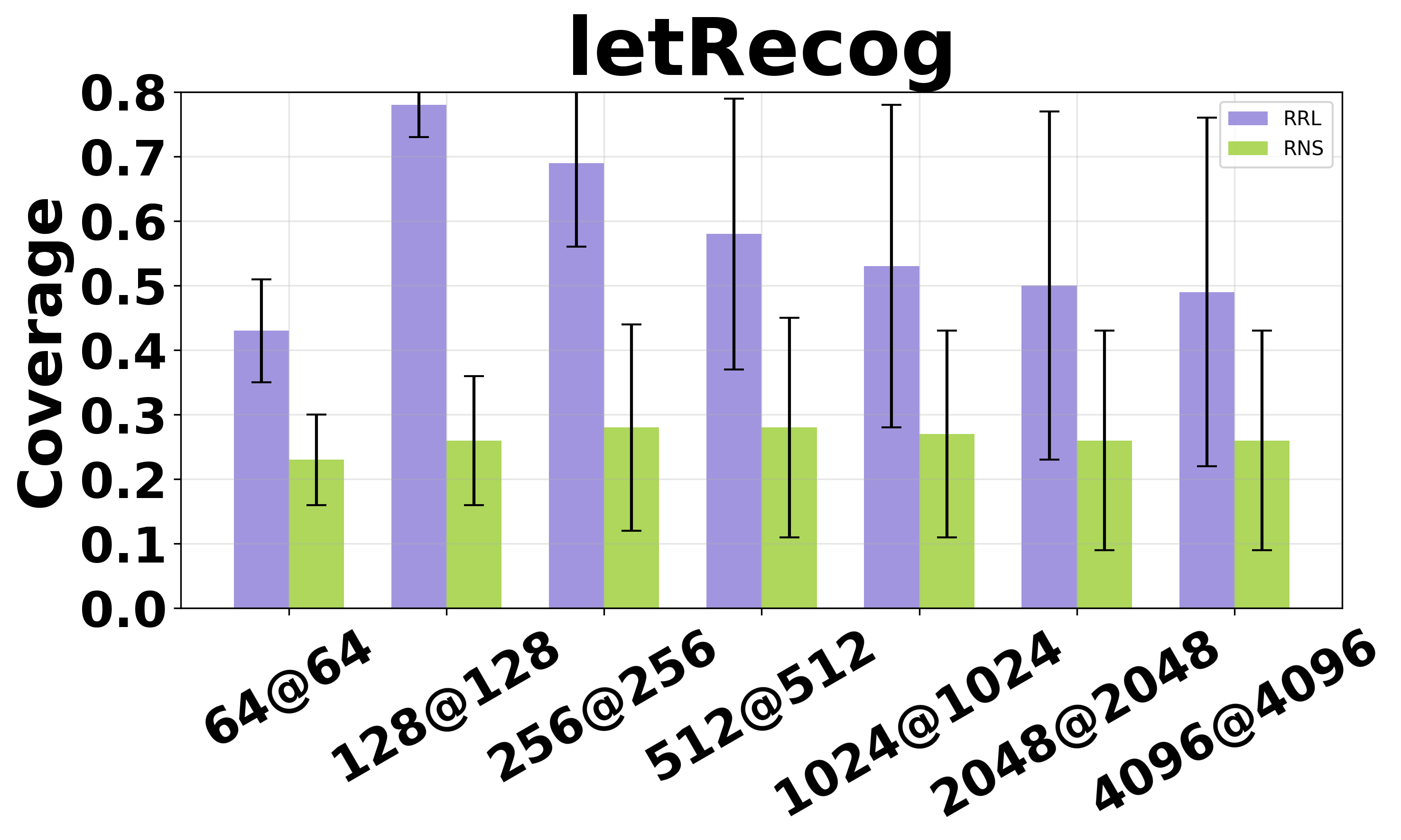}
    
  \end{minipage}%
  \hfill%
  \begin{minipage}[t]{0.20\linewidth}
    \includegraphics[width=\linewidth]{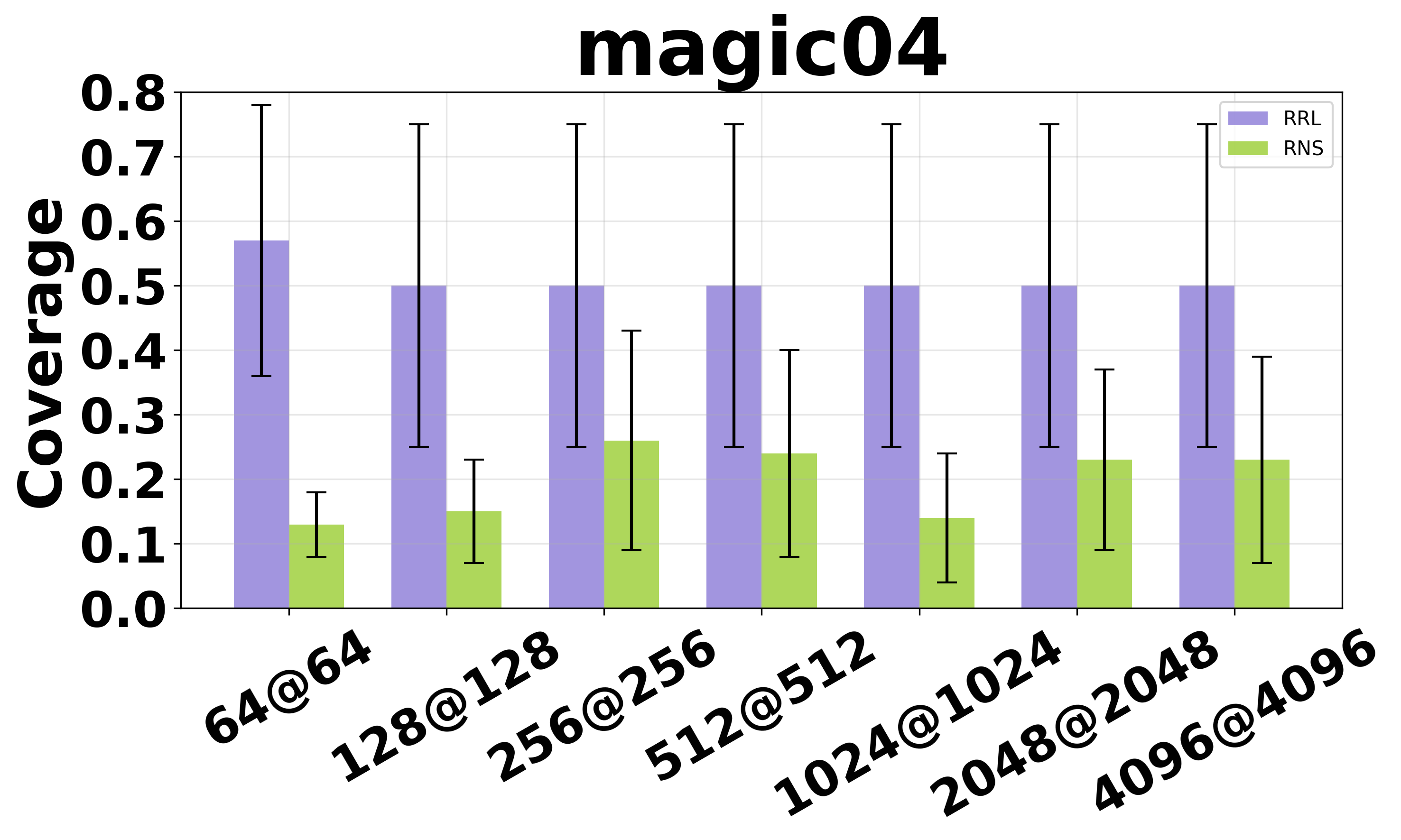}
    
  \end{minipage}

  \caption{Rule Coverage comparison across different architectures.}
  \label{fig:coverage_all}
\end{figure*}
\begin{figure*}[ht]
  \centering
  \begin{minipage}[t]{0.20\linewidth}
    \includegraphics[width=\linewidth]{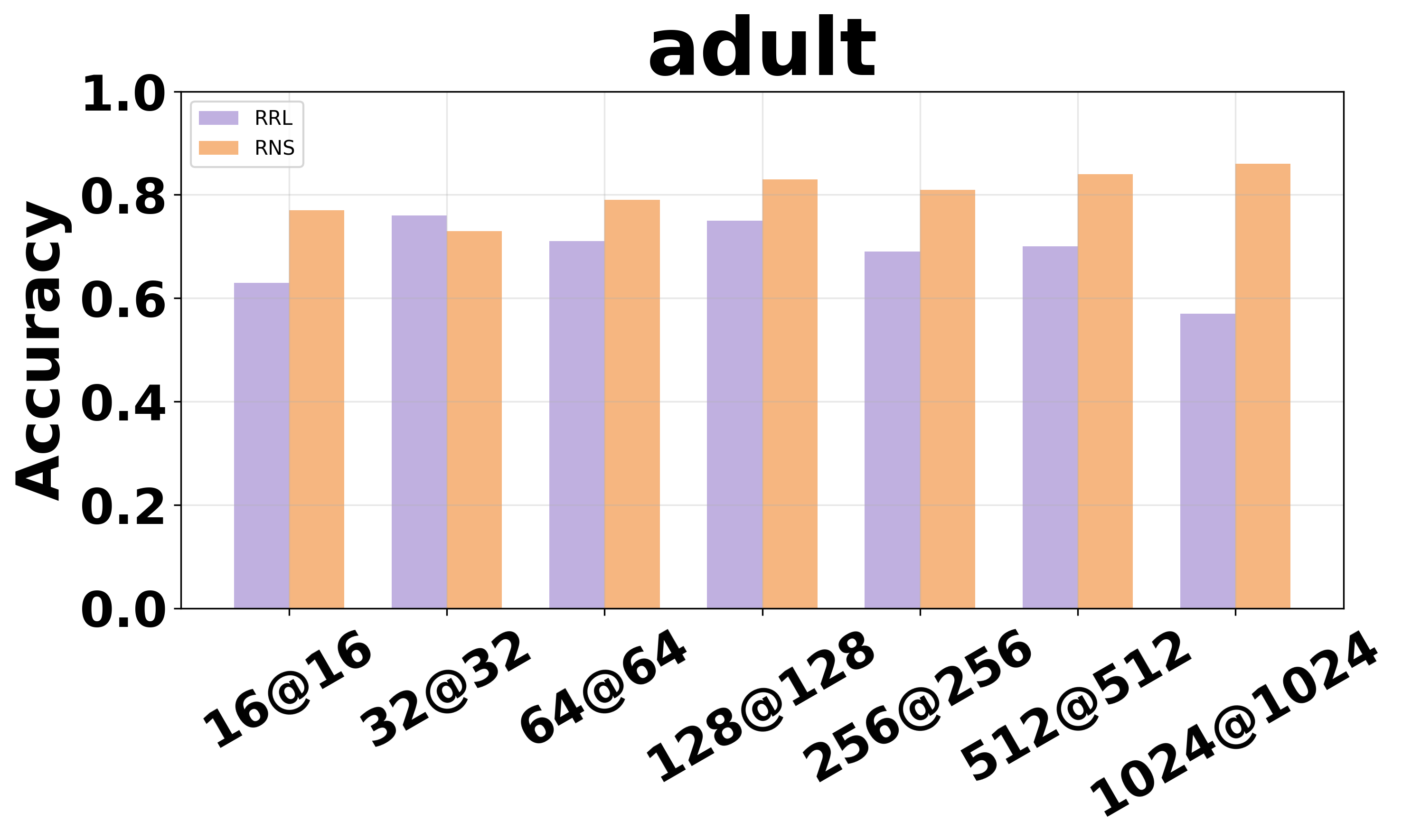}

  \end{minipage}%
  \hfill%
  \begin{minipage}[t]{0.20\linewidth}
    \includegraphics[width=\linewidth]{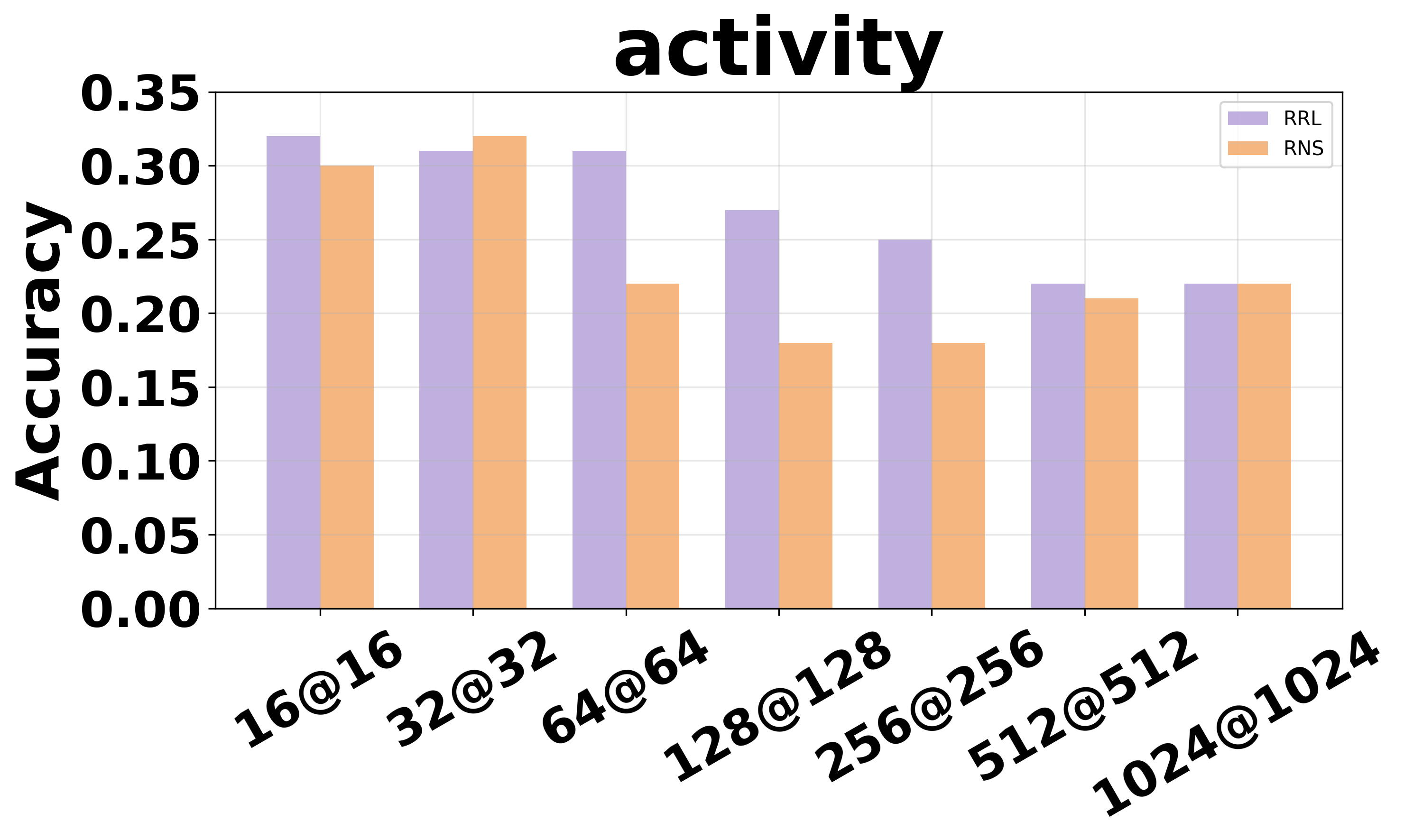}

  \end{minipage}%
  \hfill%
  \begin{minipage}[t]{0.20\linewidth}
    \includegraphics[width=\linewidth]{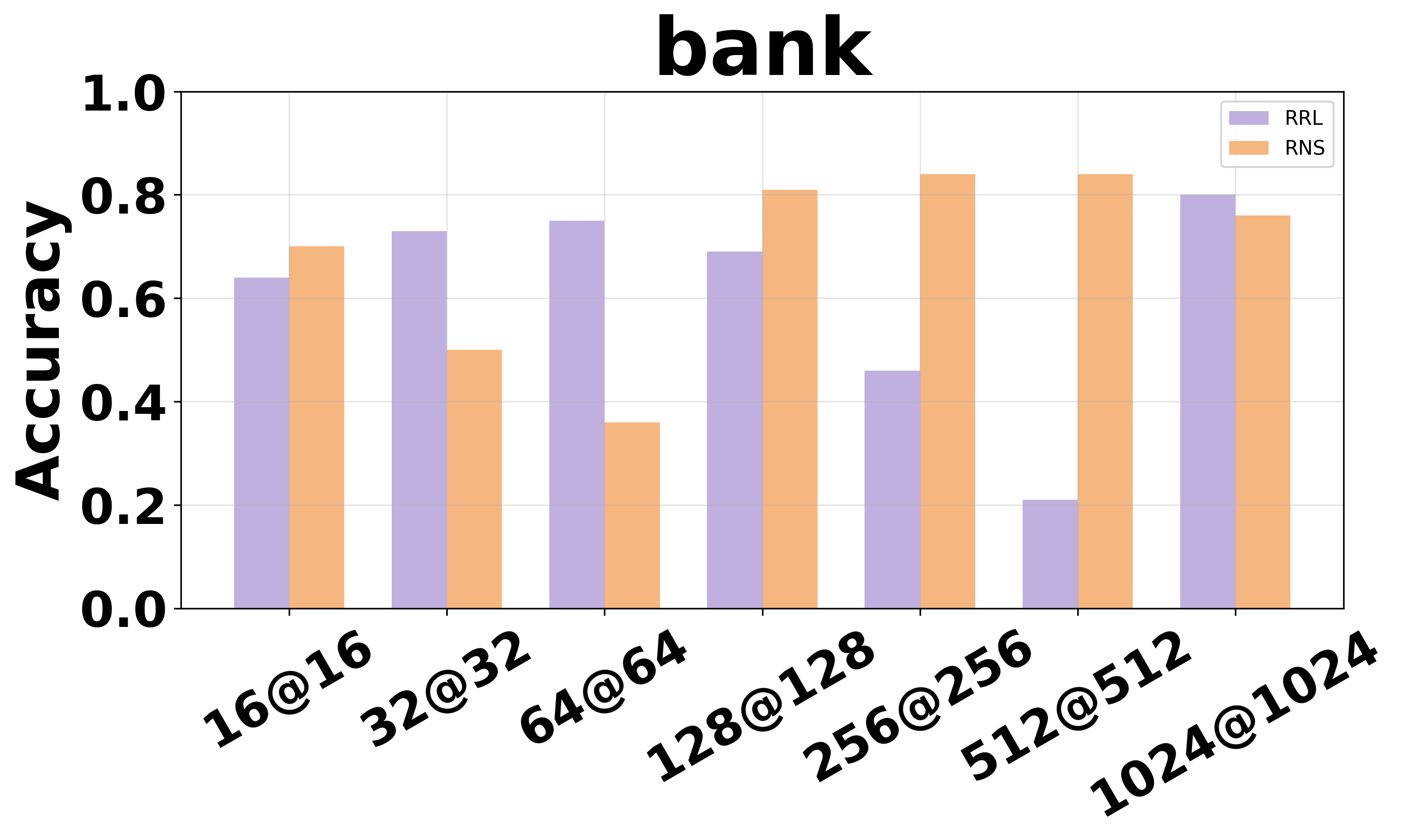}

  \end{minipage}%
  \hfill%
  \begin{minipage}[t]{0.20\linewidth}
    \includegraphics[width=\linewidth]{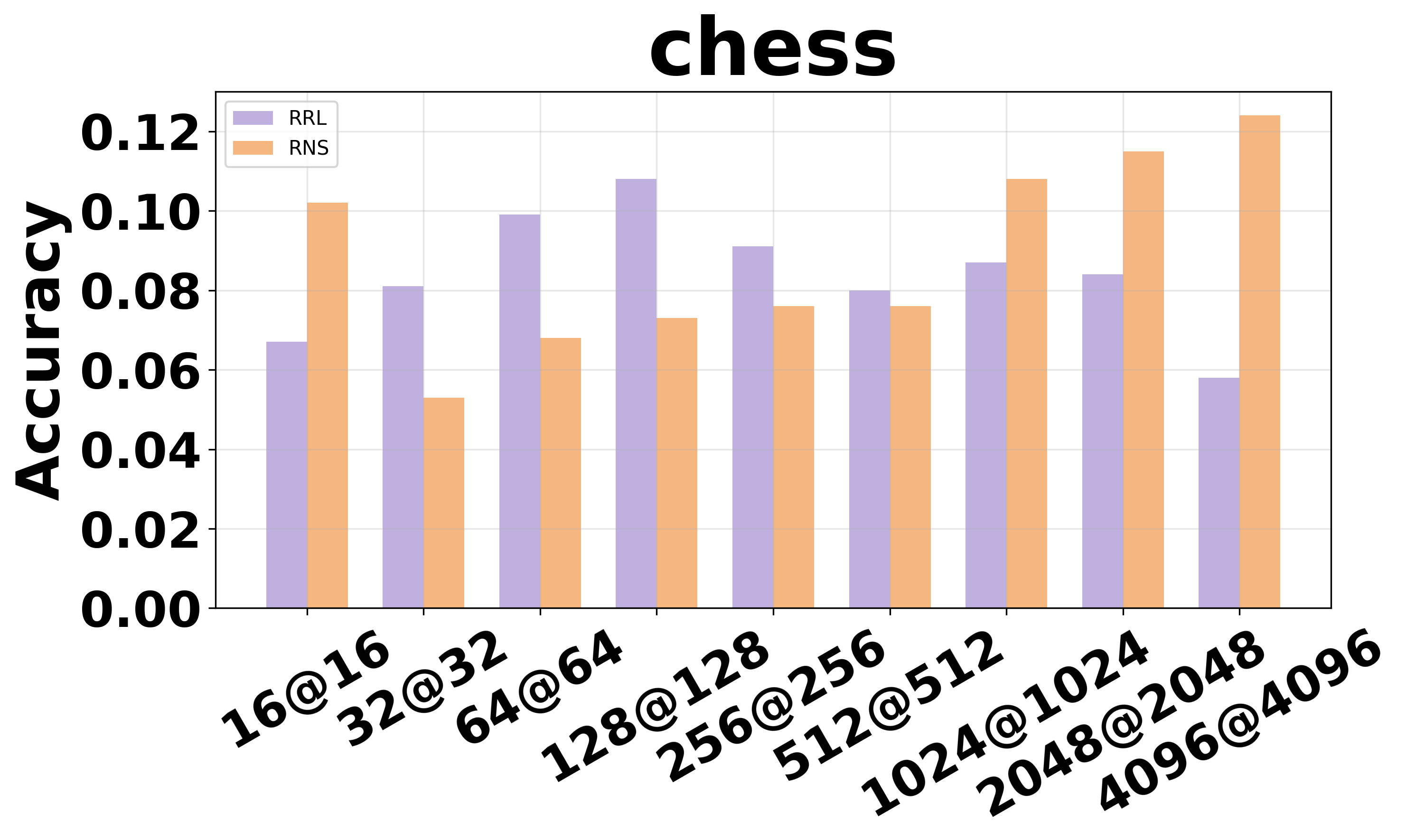}

  \end{minipage}%
  \hfill%
  \begin{minipage}[t]{0.20\linewidth}
    \includegraphics[width=\linewidth]{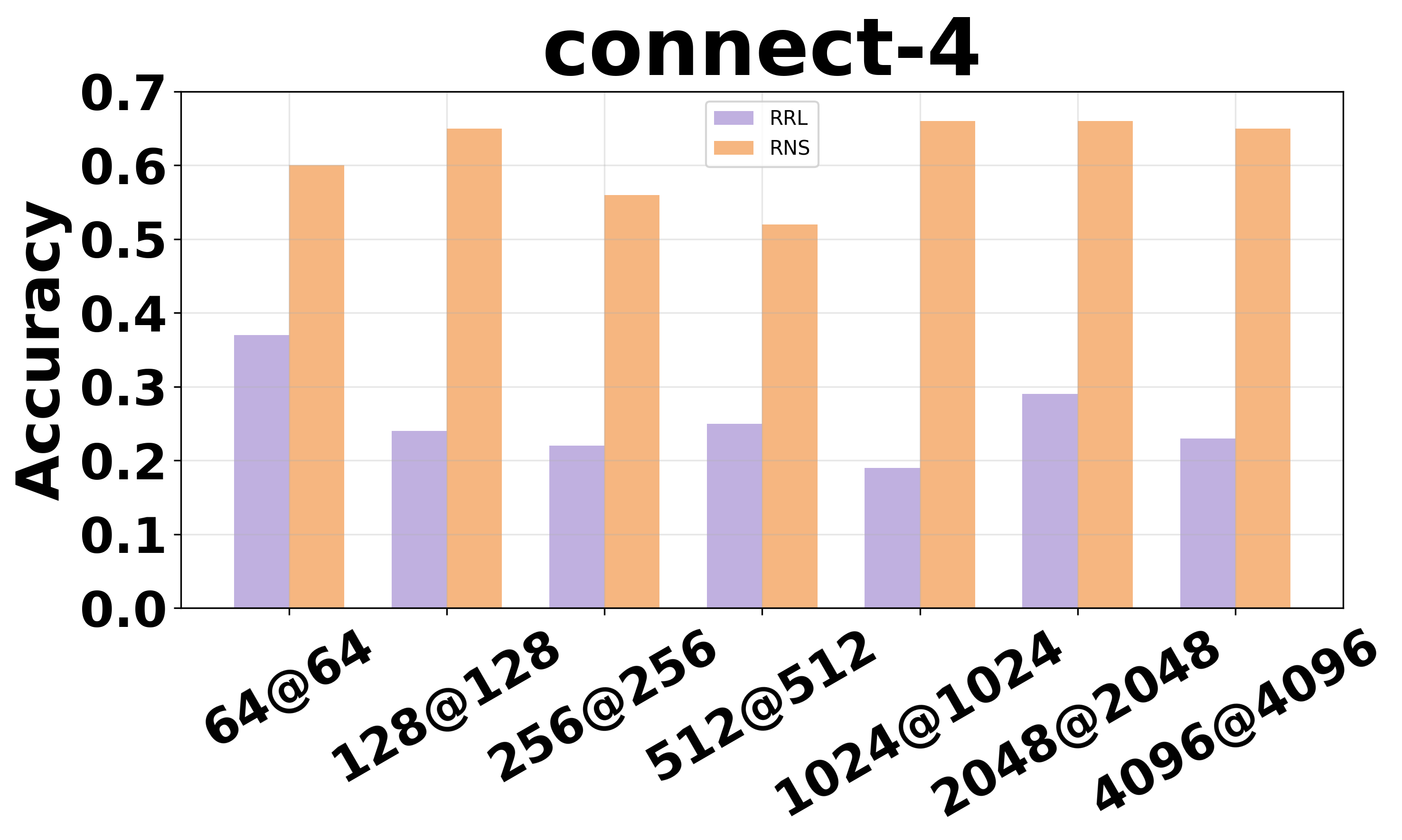}

  \end{minipage}

  \begin{minipage}[t]{0.20\linewidth}
    \includegraphics[width=\linewidth]{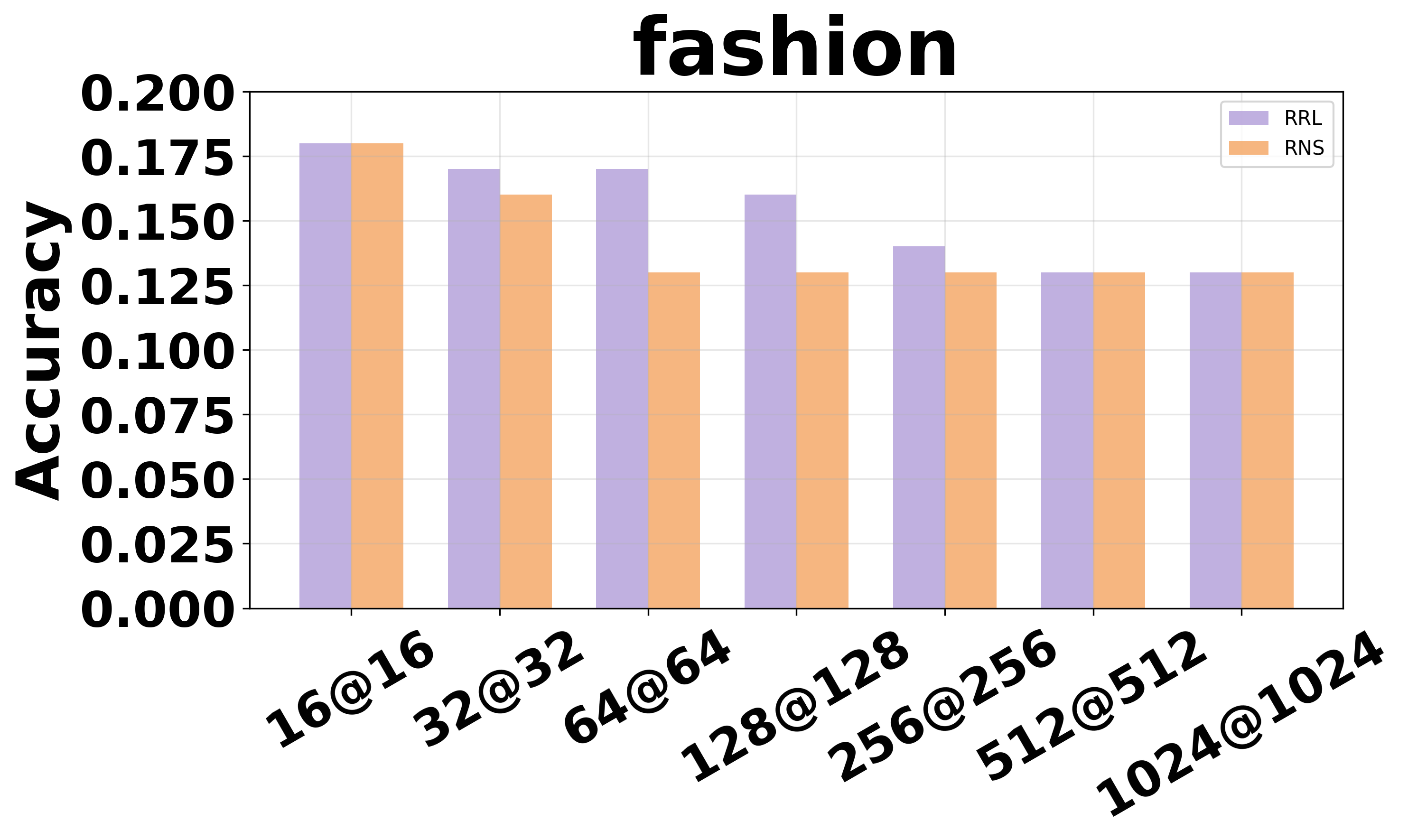}

  \end{minipage}%
  \hfill%
  \begin{minipage}[t]{0.20\linewidth}
    \includegraphics[width=\linewidth]{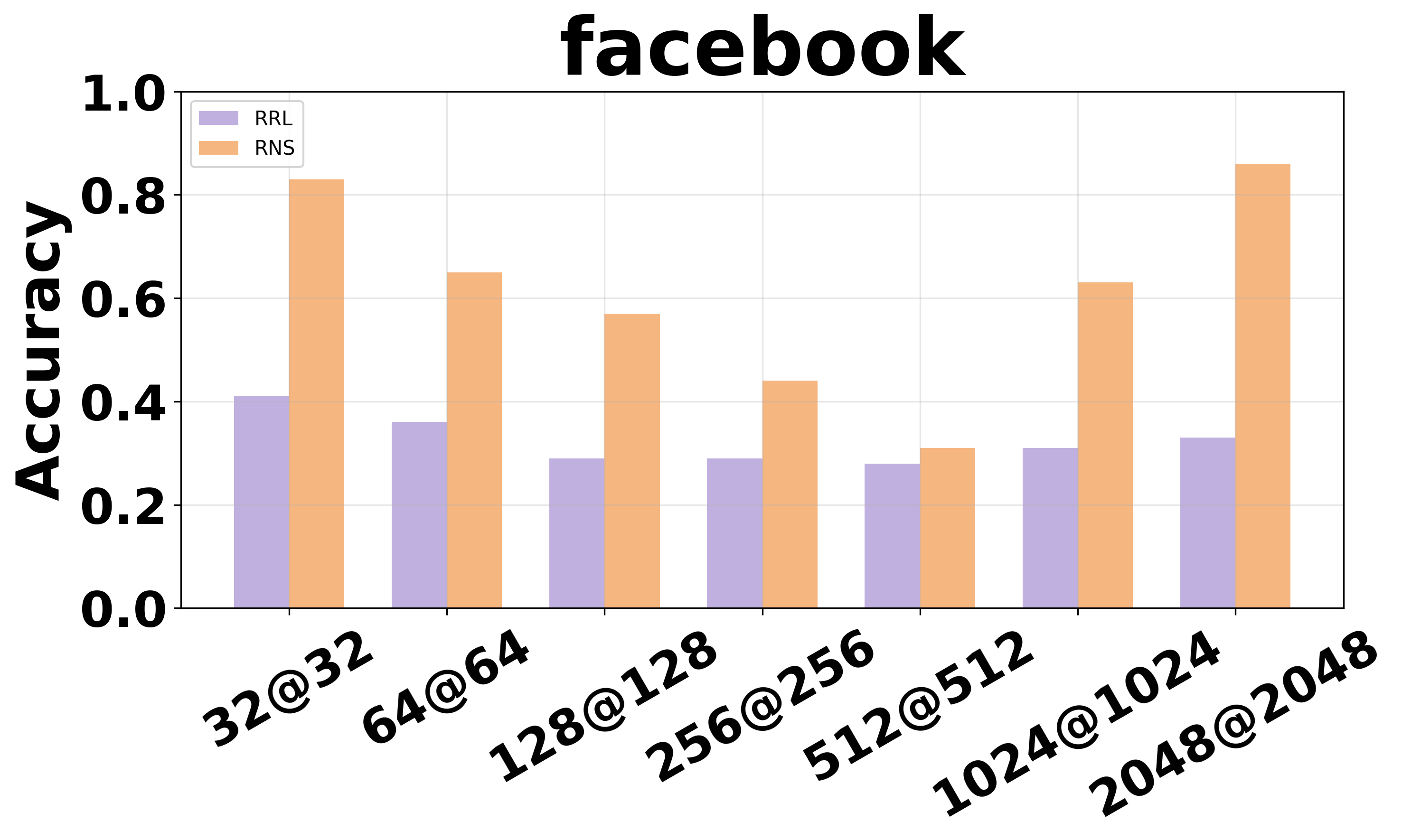}

  \end{minipage}%
  \hfill%
  \begin{minipage}[t]{0.20\linewidth}
    \includegraphics[width=\linewidth]{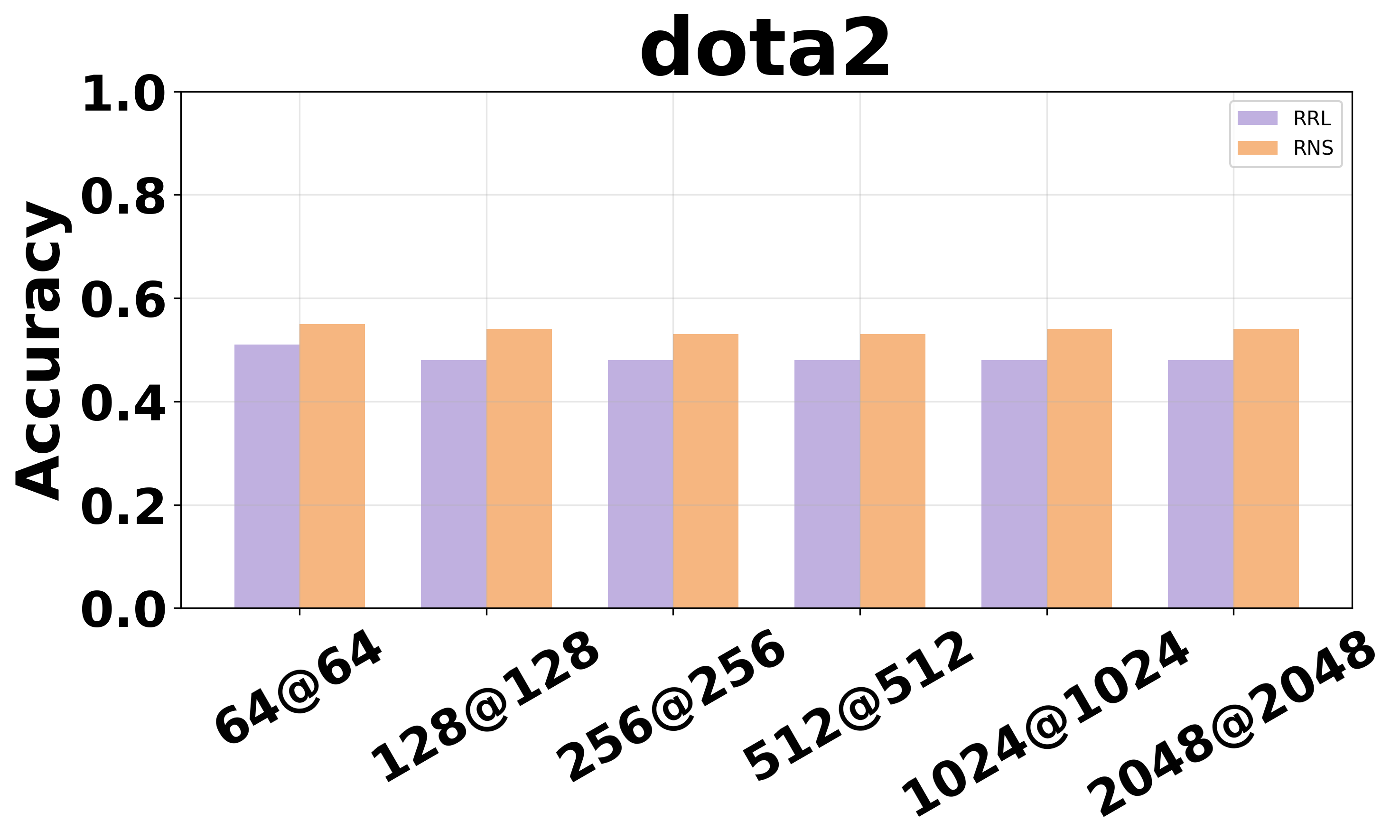}

  \end{minipage}%
  \hfill%
  \begin{minipage}[t]{0.20\linewidth}
    \includegraphics[width=\linewidth]{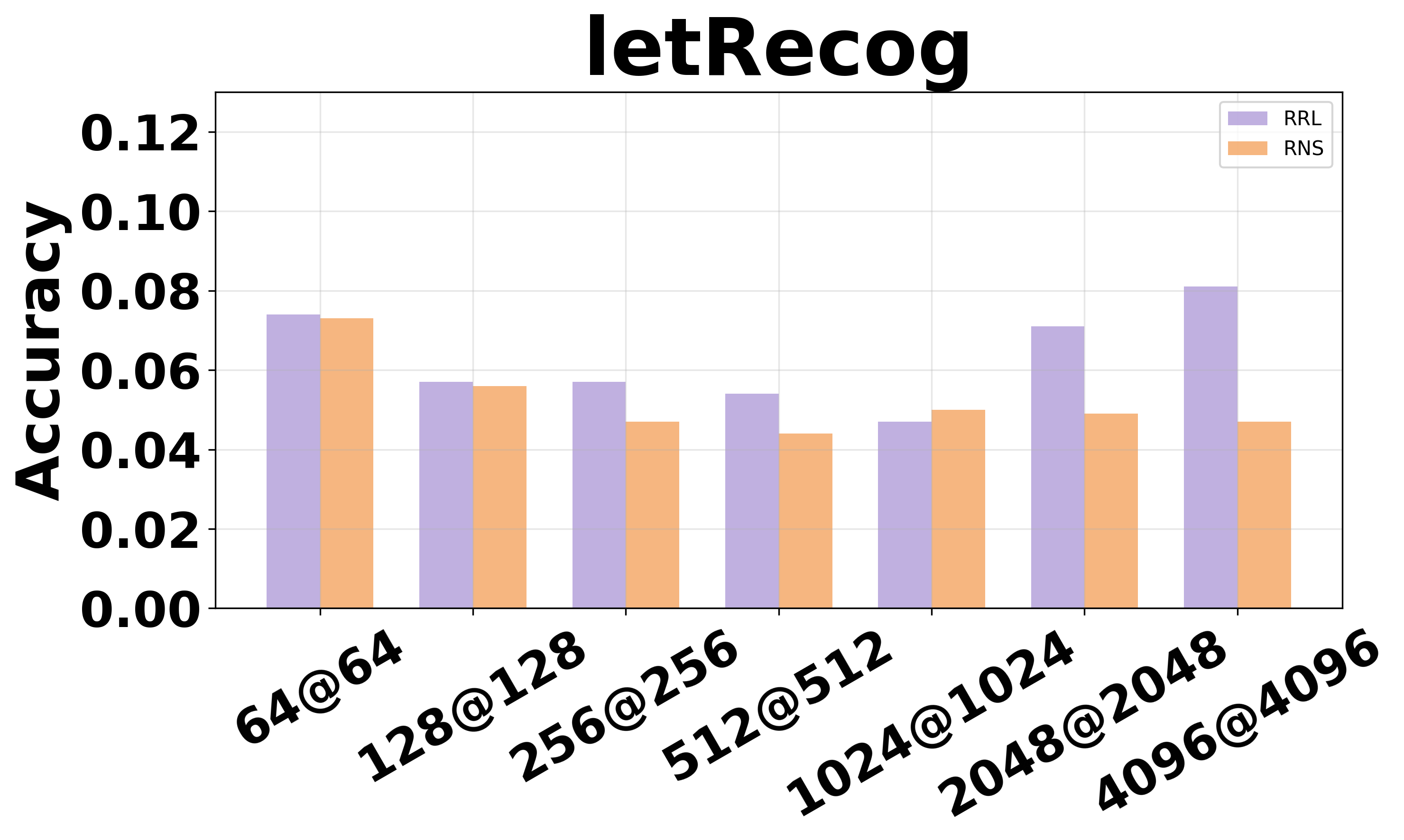}

  \end{minipage}%
  \hfill%
  \begin{minipage}[t]{0.20\linewidth}
    \includegraphics[width=\linewidth]{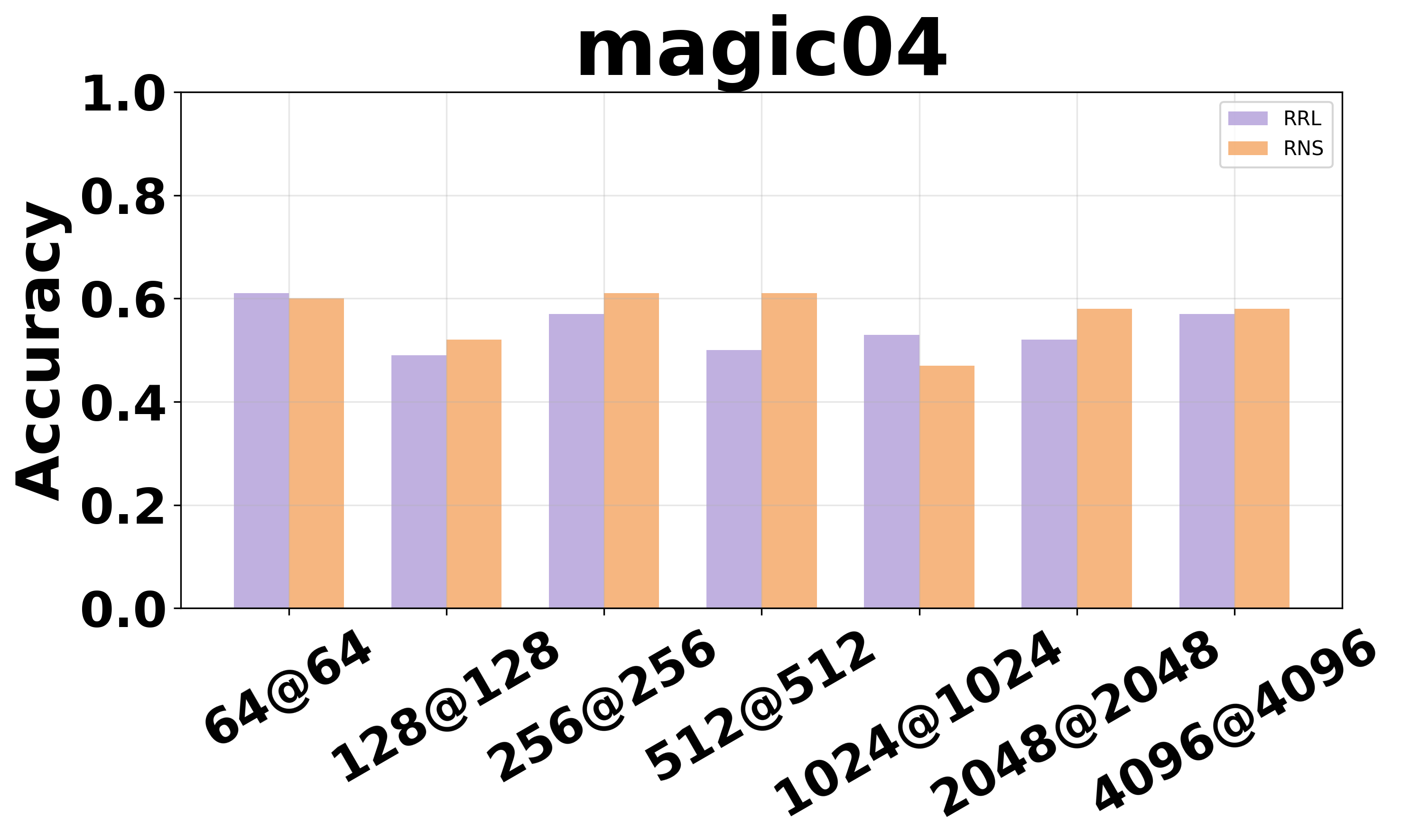}

  \end{minipage}

  \caption{Rule Accuracy comparison across different architectures.}
  \label{fig:accuracy_all}
\end{figure*}

 \textbf{Results.} We conduct extensive experiments across 10 datasets, as shown in Figure \ref{fig:diversity_all}, Figure~\ref{fig:coverage_all}, and Figure~\ref{fig:accuracy_all}. RNS consistently produces rules with superior results: higher accuracy, greater diversity, and lower coverage deviation. This indicates that the rules learned by RNS are easier to distinguish and exhibit better prediction and generalization power.

\section{Ablation Study (RQ4)} 
\label{Ablation_HCC}
\noindent\textbf{Negation Layer and HCC.} To evaluate the effectiveness of the Negation Layer and Heterogeneous Connection Constraint (HCC), we compare RNS trained with and without these components. The Negation Layer enables functional completeness by supporting negation operations in learned rules, while HCC narrows the search space for learning connections between logical layers, improving performance. Table~\ref{tab:ablation_study} shows F1 scores for RNS and its variants on the bank (small) and activity (large) datasets. The F1 score decreases without these components, highlighting their importance. HCC also improves learning efficiency. With all other factors constant, training time decreases by 11.22\% and 25\% for the activity and bank datasets, respectively, as shown in Table~\ref{tab:training_efficiency}, emphasizing its utility.

\noindent\textbf{Binning Function.} We evaluate RNS's performance with different binning functions introduced in Section~\ref{sec:binarization}. F1 scores for RNS with these methods are shown in Table~\ref{tab:binning_methods}. RNS exhibits strong flexibility, with RanInt performing the best due to its stochastic diversity, which helps prevent overfitting. Additionally, its computational efficiency and simplicity are advantageous. KInt, in contrast, clusters based solely on feature values, neglecting target variables, which can reduce its effectiveness. EntInt incorporates label information to minimize entropy within bins, potentially improving accuracy. AutoInt, while adaptable, incurs significant computational overhead due to its parameter optimization, posing challenges in practice.

\begin{table}[t!]
\centering
\setlength{\arrayrulewidth}{0.8pt}
\begin{tabular}{l|cc} 
 \toprule
 \textbf{Model Variant} & \textbf{Activity} & \textbf{Bank} \\ 
 \midrule
 RNS w/o Negation Layer & 97.68 & 76.92 \\ 
 RNS w/o HCC & 97.81 & 76.36 \\
 \midrule
 \textbf{RNS (Full)} & \textbf{98.37} & \textbf{77.18} \\
 \bottomrule
\end{tabular}
\caption{Ablation study: F1 Score (\%) of RNS and its variants on two datasets.}
\label{tab:ablation_study}
\end{table}

\begin{table}[t!]
\centering
\setlength{\arrayrulewidth}{0.8pt}
\begin{tabular}{l|cc} 
 \toprule
 \textbf{Configuration} & \textbf{Activity} & \textbf{Bank} \\ 
 \midrule
 RNS w/o HCC & 1h 38m 9s & 1h 24m 50s \\ 
 \textbf{RNS (with HCC)} & \textbf{1h 27m 3s} & \textbf{1h 3m 56s} \\
 \bottomrule
\end{tabular}
\caption{Training efficiency: Impact of Heterogeneous Connection Constraint (HCC) on training time.}
\label{tab:training_efficiency}
\end{table}

\begin{table}[t!]
\centering
\setlength{\arrayrulewidth}{0.8pt}
\begin{tabular}{l|cc} 
 \toprule
 \textbf{Binning Method} & \textbf{Activity} & \textbf{Bank} \\ 
 \midrule
 KInt (K-means) & 92.71 & 74.37 \\ 
 EntInt (Entropy) & 97.09 & 77.02 \\ 
 AutoInt (Auto-interval) & 95.62 & 75.59 \\
 \midrule
 \textbf{RanInt (Random)} & \textbf{98.80} & \textbf{77.18} \\ 
 \bottomrule
\end{tabular}
\caption{Feature preprocessing: F1 Score (\%) comparison of different binning methods.}
\label{tab:binning_methods}
\end{table}

\section{Hyperparameter Study (RQ5)}
\label{appendix:hyper_parameter}
\begin{figure}[t]
\centering
\begin{minipage}{0.5\textwidth}
    \includegraphics[width=\linewidth]{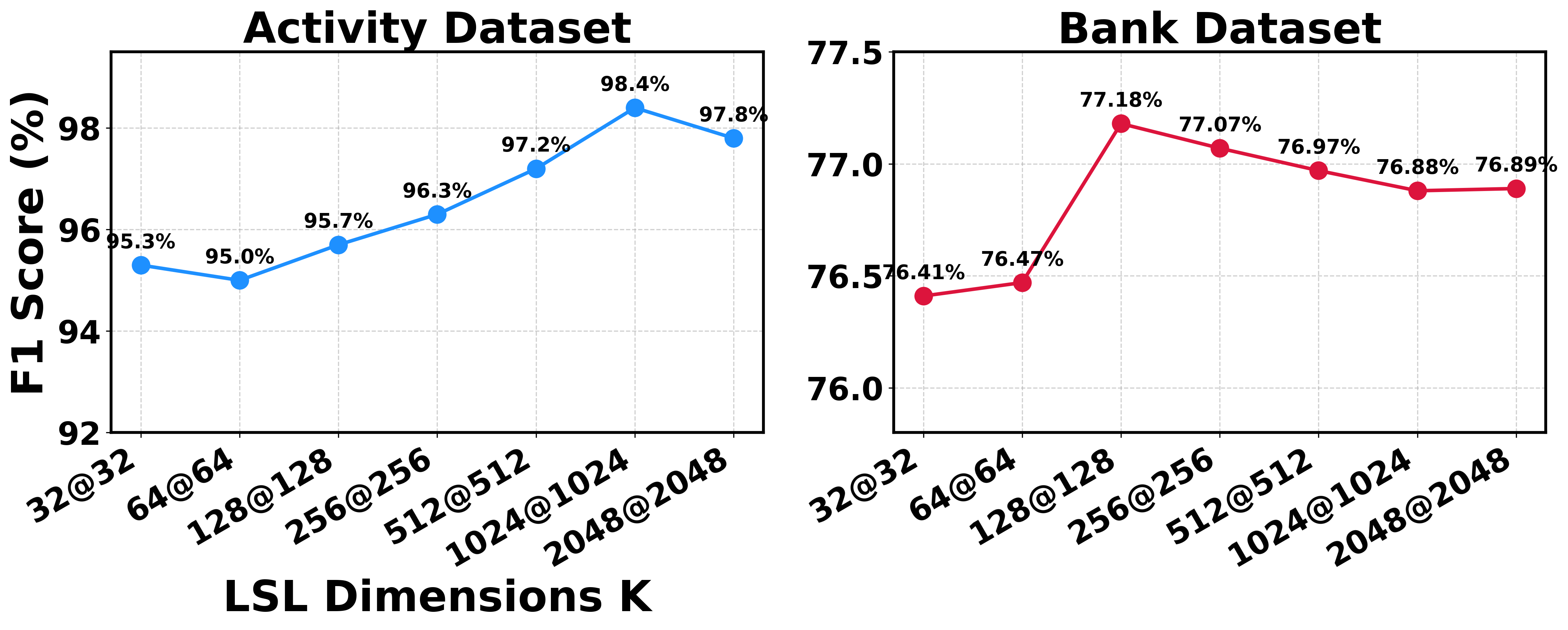}
\end{minipage}\hfill
\begin{minipage}{0.5\textwidth}
    \includegraphics[width=\linewidth]{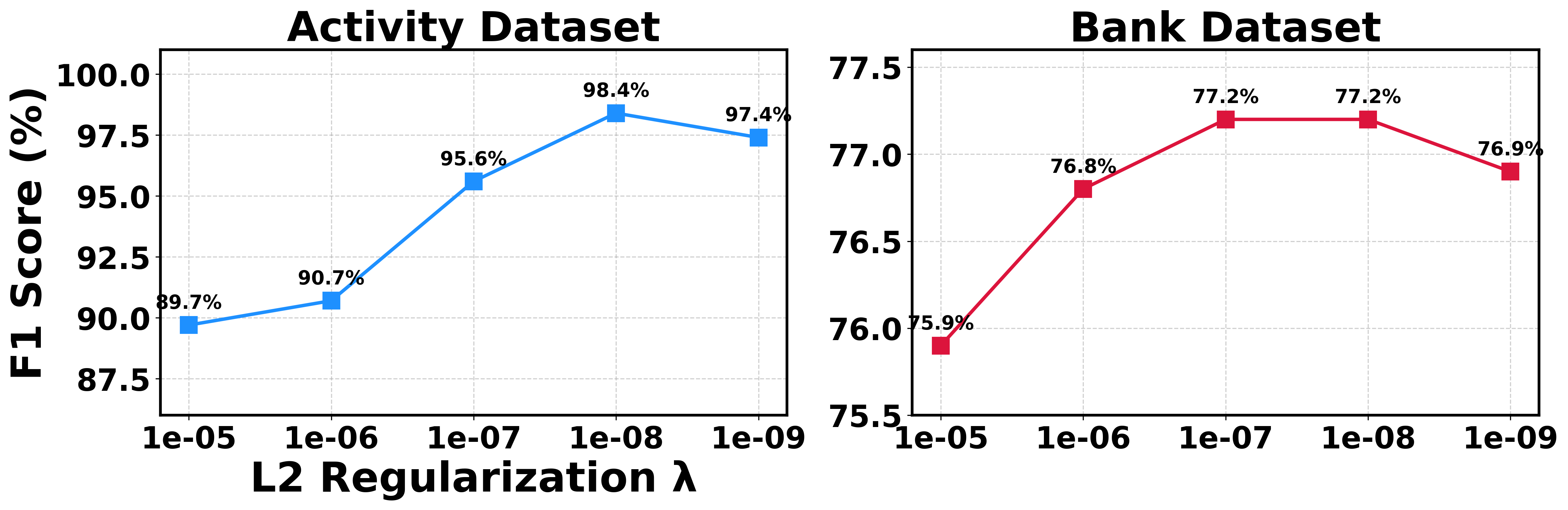}

\end{minipage}\hfill

\caption{Impact of LSL dimension K and L2 regularization for both \textit{activity} and \textit{bank}.}
\label{appendix:fig:K_l2}
\end{figure}

We evaluate LSL dimension $K$ and L2 regularization in Figure~\ref{appendix:fig:K_l2}. 
For the \textit{activity} dataset, F1 improves steadily with larger $K$, peaking at $1024@1024$, 
while the \textit{bank} dataset achieves its best performance earlier at $128@128$. 
Similarly, moderate L2 values (around $10^{-7}$--$10^{-8}$) yield the highest scores, 
with \textit{activity} reaching $98.4\%$ and \textit{bank} stabilizing near $77.2\%$. 
These results suggest that \textit{activity} dataset benefits from larger architectures and balanced regularization, 
whereas \textit{bank} saturates with smaller models and is less sensitive to $\lambda$.

\section{Gradient vanishing}
\label{gradient_vanishing}

Despite the high interpretability of discrete logical layers, training them is challenging due to their discrete parameters and non-differentiable structures. This challenge is addressed by drawing inspiration from the training methods used in binary neural networks, which involve searching for discrete solutions within a continuous space. ~\cite{wang2021scalable} leverages the logical activation functions proposed by~\cite{payani2019learning} in RRL:
\begin{equation}
    AND(h, W_{conn}^i) = \prod_{j=1}^n F_c(h_j, W_{i,j})
\end{equation}
\begin{equation}
    OR(h, W_{conn}^i) = 1 - \prod_{j=1}^n(1-F_d(h_j, W_{i,j}))
\end{equation}
where $F_c(h,w) = 1-w(1-h)$ and $F_d(h,w) = h\cdot w$. 

If $\mathbf{h}$ and $W_i$ are both binary vectors, then $\text{Conj}(\mathbf{h}, W_i) = \land_{W_{i,j}=1} h_j$ and $\text{Disj}(\mathbf{h}, W_i) = \lor_{W_{i,j}=1} h_j$. $F_c(\mathbf{h}, W)$ and $F_d(\mathbf{h}, W)$ decide how much $h_j$ would affect the operation according to $W_{i,j}$. After using continuous weights and logical activation functions, the AND and OR operators, denoted by $R$ and $S$, are defined as follows:

\begin{equation}
    r_i^{(l)} = AND(u^{(l-1)}, W_i^{(l,0)})
\end{equation}
\begin{equation}
    s_i^{(l)} = OR(u^{(l-1)}, W_i^{(l,1)})
\end{equation}
Although the whole logical layer becomes differentiable by applying this continuous relaxation, the above functions are not compatible with RNS. In this setting, the output of the node $h \in [0, 1]$ conflicts with our binarized setting, where all logical parameters are either $-1$ or $+1$. This not only breaks the inherently discrete nature of RNS but also suffers from the serious vanishing gradient problem. The reason can be found by analyzing the partial derivative of each node with respect to its directly connected weights and with respect to its directly connected nodes as follows:

\begin{equation}
    \frac{\partial r}{\partial W_{i,j}} = (u_j^{(l-1)}-1) \cdot \prod_{k\neq j}F_c(u_k^{(l-1)}, W_{i,k}^{(l,0)})
\end{equation}

\begin{equation}
    \frac{\partial r}{\partial u_{j}^{(l-1)}} = W_{i,j}^{(l,0)} \cdot \prod_{k\neq j}F_c(u_k^{(l-1)}, W_{i,k}^{(l,0)})
\end{equation}

Since $u_{j}^{(l-1)}$ and $W_{i,k}^{(l,0)}$ fall within the range of 0 to 1, the values of $F_c(\cdot)$ also lie within this range. If the number of inputs is large and most $F_c(\cdot)$ are not 1, the gradient tends toward 0 due to multiplications. Additionally, in the discrete setting, only when $u_{j}^{(l-1)}$ and $1 - F_d(\cdot)$ are all 1, can the gradient be non-zero, which is quite rare in practice.

\subsection{Detailed Analysis} \label{sec:vanish-and-solutions}

Training discrete logical layers in neural networks is challenging due to their binary parameters and non-differentiable operations. A common approach to make such models trainable is to relax the logical functions to a continuous, differentiable form~\cite{Wang_2024}. These relaxations enable gradient-based training, but they often suffer from severe \textbf{vanishing gradient} problems due to the multiplicative structure of the logical functions. Moreover, the RRL activations output values in $[0,1]$, which conflicts with a fully binarized $\pm1$ logic setting (such as our RNS approach) and breaks the discrete semantics of the network. We detail below why vanishing gradients occur in such formulations and describe the solutions proposed in RRL as well as the different strategy taken by RNS to overcome these issues.

\subsection{Vanishing Gradients in Multiplicative Logical Activations (RRL)}

In a conventional formulation of a logical AND gate~\cite{payani2019learning}, the output is the product of its binary inputs. For binary $x_i \in \{0,1\}$, one can write
\begin{equation}
\text{AND}(x_1,\ldots, x_n) = \prod_{i=1}^{n} x_i,
\end{equation}
and similarly, a logical OR can be written (using De Morgan's law) as
\begin{equation}
\text{OR}(x_1,\ldots, x_n) = 1 - \prod_{i=1}^{n}(1 - x_i).
\end{equation}

While these expressions are correct for binary values, their direct use in a neural network leads to zero gradients almost everywhere -- a phenomenon known as \emph{gradient vanishing}. The gradient of the AND with respect to one input is
\begin{equation}
\frac{\partial \text{AND}(x)}{\partial x_j} = \prod_{i \neq j} x_i,
\end{equation}
which is zero whenever any other input $x_i$ is zero. More generally, if $x_i$ are in $[0,1]$, this derivative equals the product of the other $(n-1)$ inputs. As $n$ grows or if the inputs are often fractional values $<1$, this product becomes extremely small -- decaying exponentially with $n$ -- thus severely diminishing the gradient signal. In the case of OR (product of $(1-x_i)$ terms), the same issue arises by symmetry (the gradient will contain a product of many factors in $[0,1]$). The consequence is that standard product-based logic units have large regions of the input space where the gradient is essentially zero, hampering learning.

In a fully discrete setting, if a conjunction has more than one false literal, changing any single input from 0 to 1 does not alter the output (since at least one false remains to make the AND false). Thus, the gradient at that point is exactly zero in all those input directions -- creating broad ``dead zones'' where the network cannot learn.

To enable gradient-based training, RRL replaces the hard logical operations with differentiable approximations that produce continuous outputs. Let $h=(h_1,\dots,h_n)$ be the input vector to a logical neuron (with $h_j \in [0,1]$ as relaxations of Boolean values) and let $W_i=(W_{i,1},\dots, W_{i,n})$ be a set of weights indicating which inputs are involved in the $i$-th clause (for example, $W_{i,j}\in\{0,1\}$ or $[0,1]$, with 1 meaning the $j$th input is included in the clause). RRL defines smoothed conjunction and disjunction functions as:

\begin{align}
\text{Conj}(h, W_i) &= \prod_{j=1}^{n} F_c(h_j,\,W_{i,j}), & F_c(h,w) &= 1 - w(1-h), \label{eq:rrl-conj}\\[6pt]
\text{Disj}(h, W_i) &= 1 - \prod_{j=1}^{n}\Big(1 - F_d(h_j,\,W_{i,j})\Big), & F_d(h,w) &= h \cdot w, \label{eq:rrl-disj}
\end{align}

where $F_c(h,w)$ and $F_d(h,w)$ are specific smooth blending functions for inputs $h$ and weights $w$. If $h \in \{0,1\}^n$ and $W_i$ is binary, $\text{Conj}(h, W_i)$ reduces to the logical AND of the selected inputs (those with $W_{i,j}=1$), and $\text{Disj}(h, W_i)$ becomes the logical OR. For continuous $h$ and $W_i$, these give differentiable outputs in $[0,1]$.

However, the gradient remains problematic. Consider $r_i = \text{Conj}(h,W_i)$. Using the chain rule:
\begin{align}
\frac{\partial r_i}{\partial W_{i,j}} &=\;\Big(\prod_{k \neq j} F_c(h_k, W_{i,k})\Big)\;\frac{\partial F_c(h_j, W_{i,j})}{\partial W_{i,j}} =\;(h_j - 1)\;\prod_{k \neq j} F_c(h_k, W_{i,k})~, \label{eq:rrl-grad-w}\\[6pt]
\frac{\partial r_i}{\partial h_j} &=\;\Big(\prod_{k \neq j} F_c(h_k, W_{i,k})\Big)\;\frac{\partial F_c(h_j, W_{i,j})}{\partial h_j} =\;W_{i,j}\;\prod_{k \neq j} F_c(h_k, W_{i,k})~. \label{eq:rrl-grad-h}
\end{align}

Each partial derivative is proportional to a product of $(n-1)$ terms $F_c(h_k,W_{i,k})$, each of which lies in $[0,1]$. Unless all those terms are very close to 1, the product will be small; if any term is 0, the product (and thus the gradient) is zero. In other words, the magnitude of the gradient \emph{shrinks multiplicatively} with every additional input in the clause.

A similar calculation for $s_i = \text{Disj}(h,W_i)$ yields, by symmetry (with $G_k := 1 - F_d(h_k,W_{i,k})$):
\begin{align}
\frac{\partial s_i}{\partial W_{i,j}} \;=\; h_j \prod_{k \neq j} G_k~, & \qquad \frac{\partial s_i}{\partial h_j} \;=\; W_{i,j} \prod_{k \neq j} G_k~, \label{eq:rrl-grad-disj}
\end{align}

which again contains a product of $(n-1)$ factors $G_k \in [0,1]$. Thus, for large $n$ or for typical fractional values of $h_j$ and $W_{i,j}$, these gradients \textbf{vanish} to near-zero. In practice, when many inputs of an AND are not extremely close to 1, the gradient through that AND node becomes negligibly small. And in the discrete limit ($h, W \in \{0,1\}$), $\frac{\partial r_i}{\partial W_{i,j}}$ and $\frac{\partial r_i}{\partial h_j}$ are zero in almost all cases: only if \emph{all} other inputs of the AND are 1 (so that the output is sensitive to the remaining input/literal) will a change in that input make a difference. Such a situation---e.g., a clause where all but one literal is true---is rare, meaning the network spends most of its time in regimes where the loss gradient is zero with respect to each individual parameter. This underscores the fundamental incompatibility of naive multiplicative logical units with gradient-based learning: the more literals in a rule, the more pronounced the vanishing gradient problem becomes.

\paragraph{Incompatibility with $\pm1$ Binarization (RNS).}
An additional issue is that the RRL activations $\text{Conj}(h, W)$ and $\text{Disj}(h, W)$ produce outputs in the range $[0,1]$ (since they are essentially probabilities or fractional truth values). This is incompatible with the design of RNS, where all internal logical signals are meant to be binary $\{\!-\!1,+1\}$ values at runtime. In RNS, we require a hard True or False, encoded as $+1$ or $-1$. Using RRL's continuous outputs inside an RNS network would break the discrete semantics and require additional mechanisms to re-binarize the outputs at each layer. Moreover, as discussed above, those continuous outputs suffer from vanishing gradients when used in deep clauses. Thus, while RRL's relaxation makes a logical layer differentiable, it does so at the cost of deviating from binary $\pm1$ representation and encountering extremely small gradients for large rules. These drawbacks motivate alternative approaches that maintain binary representations and avoid multiplicative shrinkage.

\subsection{RRL's Mitigations: Log-Domain Smoothing and Gradient Grafting}

RRL and related frameworks have proposed a couple of techniques to alleviate the vanishing gradient and training difficulties while still using product-based logic. We briefly summarize two such strategies: a log-domain scaled activation function, and a hybrid training method known as gradient grafting.

\paragraph{Log-space smoothing of logical activations.}
One idea introduced with RRL is to modify the product formulation by amplifying small values in the product through a log transformation. Specifically, define a projection function $P(v)$ that boosts a small product $v$:
\begin{equation}
P(v) \;=\; \frac{1}{\,1 - \log v\,}\,, \qquad \frac{dP}{dv} \;=\; \frac{P(v)^2}{\,v\,}~. \label{eq:P-def}
\end{equation}

Here $v>0$ is a value (in practice, $v$ will be a small positive number representing the product of several terms). The function $P(v)$ is chosen such that when $v$ is small, $-\log v$ is large, so $P(v)$ will significantly exceed $v$ (for example, if $v=10^{-3}$, then $P(v) \approx 1/(1-(-6.9)) \approx 1/7.9 \approx 0.127$, whereas $v$ itself is $0.001$). In this way, $P(v)$ stretches the lower end of the range upward.

RRL incorporates this into the logical units by first computing the product of inputs in log-space and then projecting back. Using a small $\varepsilon>0$ to avoid $\log 0$, the \emph{improved} conjunction and disjunction are defined as:
\begin{align}
\text{Conj}^+(h,W_i) \;&=\; P\!\Bigg(\;\prod_{j=1}^{n}\big(F_c(h_j, W_{i,j}) + \varepsilon\big)\Bigg)\,, \label{eq:conj-plus}\\[6pt]
\text{Disj}^+(h,W_i) \;&=\; 1 \;-\; P\!\Bigg(\;\prod_{j=1}^{n}\big(1 - F_d(h_j, W_{i,j}) + \varepsilon\big)\Bigg)\,. \label{eq:disj-plus}
\end{align}

When $\varepsilon \to 0$ and $h, W$ are binary, $\text{Conj}^+$ and $\text{Disj}^+$ recover the exact AND/OR as before. However, for intermediate values, these use $P(\cdot)$ to prevent the product from becoming too small. If we let 
\[v \;=\; \prod_{j=1}^{n}\big(F_c(h_j, W_{i,j}) + \varepsilon\big)\]
be the raw product inside $P$, then by the chain rule the gradient of $\text{Conj}^+$ with respect to a parameter becomes:
\begin{align}
\frac{\partial\,\text{Conj}^+}{\partial W_{i,j}} &=\; \frac{P(v)^2}{\,v\,}\;\Bigg(\prod_{k \neq j}\big(F_c(h_k, W_{i,k})+\varepsilon\big)\Bigg)\;\frac{\partial F_c(h_j, W_{i,j})}{\partial W_{i,j}}\,, \label{eq:conj-plus-grad-w}\\[6pt]
\frac{\partial\,\text{Conj}^+}{\partial h_j} &=\; \frac{P(v)^2}{\,v\,}\;\Bigg(\prod_{k \neq j}\big(F_c(h_k, W_{i,k})+\varepsilon\big)\Bigg)\;\frac{\partial F_c(h_j, W_{i,j})}{\partial h_j}\,. \label{eq:conj-plus-grad-h}
\end{align}

Comparing these to the original gradients \eqref{eq:rrl-grad-w}--\eqref{eq:rrl-grad-h}, we see that the pure product term $\prod_{k \neq j} F_c(h_k,W_{i,k})$ is now multiplied by $\frac{P(v)^2}{v}$. For moderately small $v$, this factor can be significantly larger than 1, thereby attenuating the vanishing effect. Intuitively, instead of the gradient shrinking proportional to $v$ (the product of many small terms), it shrinks proportional to $P(v)^2$, and since $P(v) > v$ when $v$ is small, the decay is slower. This log-space trick can appreciably increase gradient magnitudes when the clause length $n$ is not too large or the inputs are not too extreme.

Comparing these to the original gradients \eqref{eq:rrl-grad-w}--\eqref{eq:rrl-grad-h}, we see that the pure product term $\prod_{k \neq j} F_c(h_k,W_{i,k})$ is now multiplied by $\frac{P(v)^2}{v}$. For moderately small $v$, this factor can be significantly larger than 1, thereby attenuating the vanishing effect. 

To understand how the gradient shrinks, let's analyze this mathematically. Consider the product term:
\begin{equation}
v = \prod_{j=1}^{n} F_c(h_j, W_{i,j}) = \prod_{j=1}^{n} (1 - W_{i,j}(1-h_j))
\end{equation}

For typical intermediate values where $h_j \approx 0.5$ and $W_{i,j} \approx 0.5$, we have:
\begin{equation}
F_c(h_j, W_{i,j}) \approx 1 - 0.5(1-0.5) = 0.75
\end{equation}

Therefore, the product becomes:
\begin{equation}
v \approx (0.75)^n
\end{equation}

As $n$ increases, this decays exponentially:
\begin{align}
n = 10: \quad &v \approx 0.75^{10} \approx 0.056 \\
n = 20: \quad &v \approx 0.75^{20} \approx 0.003 \\
n = 50: \quad &v \approx 0.75^{50} \approx 5.7 \times 10^{-7} \\
n = 100: \quad &v \approx 0.75^{100} \approx 3.2 \times 10^{-13}
\end{align}

The gradient magnitude without log-smoothing is proportional to $v$:
\begin{equation}
\left|\frac{\partial r_i}{\partial W_{i,j}}\right| \propto v \approx \alpha^n \quad \text{where } \alpha < 1
\end{equation}

With the log-space projection $P(v)$, the gradient becomes:
\begin{equation}
\left|\frac{\partial\,\text{Conj}^+}{\partial W_{i,j}}\right| \propto \frac{P(v)^2}{v} = \frac{1}{v(1-\log v)^2}
\end{equation}

For small $v$, we have $P(v) \approx \frac{1}{-\log v}$, so:
\begin{equation}
\frac{P(v)^2}{v} \approx \frac{1}{v(\log v)^2}
\end{equation}

Let's evaluate this for different clause sizes:
\begin{align}
n = 10: \quad v \approx 0.056, \quad &\frac{P(v)^2}{v} \approx \frac{1}{0.056 \times (-2.88)^2} \approx 2.15 \\
n = 20: \quad v \approx 0.003, \quad &\frac{P(v)^2}{v} \approx \frac{1}{0.003 \times (-5.81)^2} \approx 9.87 \\
n = 50: \quad v \approx 5.7 \times 10^{-7}, \quad &\frac{P(v)^2}{v} \approx \frac{1}{5.7 \times 10^{-7} \times (-14.4)^2} \approx 8.5 \times 10^{3}
\end{align}

While $P(v)^2/v$ grows as $v$ shrinks, for very small $v$ (large $n$), the growth is only polynomial in $\log(1/v) \approx n\log(1/\alpha)$, not enough to fully compensate for the exponential decay. Eventually, for very large $n$:
\begin{equation}
\lim_{n \to \infty} \frac{P(v)^2}{v} \approx \frac{1}{\alpha^n \cdot n^2 \cdot (\log \alpha)^2} \to 0
\end{equation}

While $P(v)^2/v$ grows as $v$ shrinks, for very small $v$ (large $n$), the growth is only polynomial in $n$, not enough to fully compensate for the exponential decay. 

To see this clearly, recall that $v \approx \alpha^n$ where $\alpha < 1$ (e.g., $\alpha = 0.75$). Then:
\begin{equation}
P(v) = \frac{1}{1 - \log v} = \frac{1}{1 - \log(\alpha^n)} = \frac{1}{1 - n\log\alpha}
\end{equation}

Since $\alpha < 1$, we have $\log\alpha < 0$, so $-\log\alpha > 0$. For large $n$:
\begin{equation}
P(v) \approx \frac{1}{n|\log\alpha|}
\end{equation}

Therefore:
\begin{equation}
\frac{P(v)^2}{v} \approx \frac{1/n^2(\log\alpha)^2}{\alpha^n} = \frac{1}{\alpha^n \cdot n^2 \cdot (\log\alpha)^2}
\end{equation}

The key observation is:
\begin{itemize}
\item The numerator grows as $\mathcal{O}(1/n^2)$ (polynomial decay)
\item The denominator decays as $\mathcal{O}(\alpha^n)$ (exponential decay)
\end{itemize}

Since exponential decay dominates polynomial growth:
\begin{equation}
\lim_{n \to \infty} \frac{1}{\alpha^n \cdot n^2 \cdot (\log\alpha)^2} = \lim_{n \to \infty} \frac{1}{n^2 (\log\alpha)^2} \cdot \frac{1}{\alpha^n} \to 0
\end{equation}

because $\frac{1}{\alpha^n}$ approaches 0 much faster than $\frac{1}{n^2}$ approaches infinity.

Thus, while the log-space trick delays the vanishing, it cannot prevent it for large $n$. Intuitively, instead of the gradient shrinking proportional to $v$ (the product of many small terms), it shrinks proportional to $P(v)^2$, and since $P(v) > v$ when $v$ is small, the decay is slower. This log-space trick can appreciably increase gradient magnitudes when the clause length $n$ is not too large or the inputs are not too extreme.

However, this modification does \textbf{not fully eliminate} the vanishing gradient problem. When $n$ is very large or many inputs are significantly below 1, the initial product $v$ becomes extremely tiny (e.g. $10^{-10}$ or smaller), making $\log v$ a large negative number. In such extreme cases, $P(v)$ itself approaches 0 (since $1 - \log v$ is huge), and consequently $P(v)^2/v$ can also become very small. In the worst case, if any factor in the product is 0, $v=0$ and no finite smoothing can help ($P(0)$ is undefined without $\varepsilon$ and effectively $P(v)^2/v$ remains near 0 for very small $v$). Thus, for very complex clauses or highly non-saturated inputs, the gradients may still collapse to nearly zero. Moreover, once we ultimately project the network to discrete weights and inputs ($h, W \in \{0,1\}$), the gradient at those exact binary points is again zero in most directions (as discussed earlier). In summary, $P(v)$-based smoothing improves gradient flow for moderately small signals but does not fundamentally overcome vanishing gradients when training very large logical expressions. Additional strategies are required to train purely discrete models.

\paragraph{Gradient Grafting for discrete training.}
Another technique used in RRL to handle training with binary decisions is \textbf{Gradient Grafting}. The idea is to maintain two parallel models during training: a continuous one that is used for backpropagation, and a discrete one that defines the actual loss. Let $\theta$ denote the set of trainable continuous parameters (weights) and $q(\theta)$ be a binarization function that maps these to discrete values (for instance, thresholding each weight to 0 or 1). We denote by $\hat{Y} = F(\theta, X)$ the output of the continuous model on input $X$, and by $\bar{Y} = F(q(\theta), X)$ the output of the corresponding binarized model (i.e., the actual logical network with hard decisions). Training proceeds by computing the loss $L(\bar{Y})$ on the discrete model's output, but then updating $\theta$ using gradients from the continuous model. In formula:
\begin{equation}
\theta_{t+1} \;=\; \theta_t \;-\; \eta \;\bigg[\frac{\partial L(\bar{Y})}{\partial \bar{Y}}\bigg]\;\bigg[\frac{\partial \hat{Y}}{\partial \theta_t}\bigg]~, \label{eq:grad-grafting}
\end{equation}

where $\eta$ is the learning rate. In this update, $\frac{\partial L(\bar{Y})}{\partial \bar{Y}}$ is the gradient of the loss with respect to the discrete model's output (this measures how the final loss would change if the discrete output changed), and $\frac{\partial \hat{Y}}{\partial \theta_t}$ is the Jacobian of the continuous model's output with respect to its parameters (which is well-defined and non-zero because $\hat{Y}$ is produced by smooth activations). The chain of these two terms provides an effective surrogate gradient for $\theta$ that steers the discrete model's loss $L(\bar{Y})$ downward, even though $\bar{Y}$ itself has zero or undefined gradients w.r.t. $\theta$. In essence, the discrete network's error signal is ``grafted'' onto the continuous network's sensitivity.

This approach circumvents the vanishing gradient at the discrete points by always following the continuous proxy's gradients, and can successfully optimize a logical network where direct backpropagation would fail. Gradient grafting, however, comes at the cost of increased training complexity. The optimization is no longer a simple gradient descent on a single well-defined objective; instead, it couples two models (one binary, one continuous) and relies on their interplay. The update in Eq.~\eqref{eq:grad-grafting} is not the true gradient of any single loss function with respect to $\theta$, since $L(\bar{Y})$ is evaluated on a different forward path than $\hat{Y}$. Thus, careful tuning and heuristics may be needed to make this training scheme converge reliably. Nonetheless, this method has been shown to help train discrete logical networks that would otherwise be stuck due to vanishing gradients.

In summary, RRL's approach to training discrete logical rules involves smoothing the logical operations (to keep gradients alive) and using a hybrid training procedure to inject discrete loss information, partially mitigating the vanishing gradient issue.

\subsection{RNS: \texorpdfstring{$\boldsymbol{\pm1}$ Encoding and $\min$/$\max$ Logical Activations}{±1 Encoding and min/max Logical Activations}}

RNS takes a fundamentally different route to avoid vanishing gradients: it redesigns the logical neuron computations and encoding so that gradients do not collapse in the first place, even at discrete points. There are two key aspects to this strategy:

\textbf{(1) $\pm1$ Binary Encoding (No Zeros).} Instead of representing False as $0$ and True as $1$, RNS encodes Boolean values as $-1$ (false) and $+1$ (true). All intermediate logical signals in the network use this $\{\!-\!1, +1\}$ domain. The advantage of this encoding is that multiplying by $-1$ flips a signal's truth value while multiplying by $+1$ leaves it unchanged -- and importantly, \textbf{zero is never an output of a logical unit.} This means we never encounter the situation of a gradient being multiplied by 0 (which was a major issue in the $0/1$ encoding). By design, removing 0 from the state space ensures that no single input can annihilate the gradient by being zero.

\textbf{(2) Min/Max-Based Logical Operations (Non-multiplicative).} Instead of using products to represent AND/OR, RNS uses \emph{piecewise-linear extremum functions} that exactly mimic logic under the $\pm1$ encoding. Specifically, if a set of inputs $\{x_1, x_2, \dots, x_n\}$ are all either $-1$ or $+1$, we define:
\[
\text{AND}(x_1,\ldots,x_n) \;=\; \min\{x_1,\ldots,x_n\}, \qquad \text{OR}(x_1,\ldots,x_n) \;=\; \max\{x_1,\ldots,x_n\}.
\]

For example, if any $x_j = -1$ (False), the minimum will be $-1$, correctly giving AND = False; only if all $x_j = +1$ will the minimum be $+1$ (True). Similarly, the maximum returns $+1$ if any input is true. These operators perfectly realize the Boolean logic of AND/OR for $\pm1$ inputs, but unlike products, they are not multiplicative and do not cause exponential shrinkage of gradients. Instead, they behave as selectors: the AND output is whichever input is the ``most false'' (most negative), and the OR output is the ``most true'' (most positive).

Because $\min$ and $\max$ are piecewise linear functions, we can define well-behaved subgradients for them. Suppose $y = \max(x_1,\ldots,x_n)$. In the case of no ties, exactly one input attains this maximum value; say $x_k$ is the largest. A small change in $x_k$ will change $y$ equally (a 1-to-1 slope), whereas changes in any other $x_j$ (that are below the max) have no effect on $y$ (as long as they don't exceed $x_k$). One convenient choice of subgradient is to assign $\frac{\partial y}{\partial x_k} = 1$ for one of the maximal indices $k$, and $\frac{\partial y}{\partial x_j} = 0$ for all other $j\neq k$. More generally, when there are ties (multiple inputs share the max value), the gradient can be split among them. A common subgradient is:
\begin{equation}
\frac{\partial y}{\partial x_i} \;=\; \begin{cases} 
\frac{1}{|M|} &\text{if $i \in M$,}\\[6pt] 
0 &\text{if $i \notin M$,} 
\end{cases} \qquad\text{where $M = \{ j : x_j = \max_k x_k\}$}~. \label{eq:minmax-subgrad}
\end{equation}

An analogous definition applies for $y = \min(x_1,\ldots,x_n)$: the subgradient is $1/|m|$ for inputs $i$ that attain the minimum and 0 for others (where $m = \{j: x_j = \min_k x_k\}$). In words, the gradient of a $\max$ gate is distributed equally to the input(s) that are currently ``winning'' (i.e., the True inputs in an OR, or the least False inputs in an AND when all are true). Crucially, this gradient \emph{does not vanish with $n$}: at least one input of an AND/OR receives a substantial gradient (of order 1), indicating it is responsible for the output. Even in the worst case of a tie among all $n$ inputs (e.g., all inputs are $-1$ for an AND, or all $+1$ for an OR), each input would get a gradient of $1/n$ -- which decays only linearly with $n$, not exponentially. In most cases, only one or a few inputs determine the extremum, and they get a full magnitude gradient. There is no multiplicative chaining of many factors as in RRL's $F_c$ or $F_d$ products. Additionally, negation in RNS is handled simply by a sign-flip: each literal may have a trainable indicator that either uses the variable as $+1$ (positive literal) or negates it ($-1$). Negation is just multiplication by $-1$, which is trivially differentiable (its subgradient is $-1$ when active, or essentially treated as a constant factor).

\textbf{Training with Straight-Through Estimators (STE).} Both the $\pm1$ encoding and the use of $\min/\max$ activations are still inherently non-differentiable as functions (the $\max$ has a kink where two inputs are equal, and the sign function that produces $\pm1$ outputs is discontinuous). However, RNS is amenable to standard techniques for training binarized networks, particularly the \emph{straight-through estimator} (STE) for gradients. In backpropagation, whenever a gradient hits a non-differentiable threshold (such as the sign function that produces $\pm1$ outputs), RNS simply treats that operation as an identity mapping for the sake of gradient computation. This is a common approach in binary neural networks to approximate gradients through quantization. In the case of the $\min/\max$ gates, we use the subgradient defined in Eq.~\eqref{eq:minmax-subgrad} during backpropagation. The combination of STE and subgradient for $\min/\max$ ensures that at \emph{discrete} operating points, the network can still propagate meaningful gradients.

For example, consider an AND implemented as $y = \min(x_1,\ldots,x_n)$ with all inputs either $+1$ or $-1$. Suppose $y = -1$ (False) because at least one $x_j = -1$. In the forward pass, this yields $y=-1$. In the backward pass, the subgradient will assign a non-zero derivative to each input that was equal to the minimum (i.e., to each $x_j$ that is $-1$). This correctly identifies that increasing any of those $x_j$ from $-1$ to $+1$ (i.e., flipping a false literal to true) would change the AND output to a higher value (potentially making the whole conjunction true if that was the only false). Thus, even though the forward function is flat for changes in any single input (since you need all falses to flip for the AND to turn true), the chosen subgradient provides a direction for learning: it tells the network to try flipping those false literals. By contrast, in a multiplicative AND, if more than one input is false, the true gradient is strictly zero for a single-input change -- there is no signal at all indicating which inputs are candidates to flip. An STE cannot magically invent a meaningful gradient in that case without risking divergence from the actual function's behavior. In RNS, however, the STE is effectively aligning with the inherent combinatorial structure of the logic: it distributes blame to all currently-false conditions for an AND (or to all currently-true conditions for an OR that outputs true, in case of ties). This yields a robust training signal even in fully discrete regimes.

\subsection{Why RNS Avoids Vanishing Gradients}

In summary, the design choices in RNS eliminate the root causes of vanishing gradients that plague RRL's product-based logic:

\begin{itemize}
\item \textbf{Activation Form:} RRL uses multiplicative activations (product for AND, complement-product for OR), which cause gradient magnitudes to contract multiplicatively with clause size. Even the improved RRL with the $P(v)$ log-smoothing still relies on a product (modulated by a corrective factor) and can suffer when many terms are far from 1. In contrast, RNS uses $\min/\max$ activations, which are piecewise linear selectors. There is no long product over many inputs; the gradient comes from identifying the extremal input(s). This fundamental difference means RNS does not inherently squeeze the gradient as the number of inputs grows.

\item \textbf{Binary Encoding:} RRL's $0/1$ encoding introduces an actual zero output (false = 0) that can outright nullify gradients (any factor of 0 in a product zeroes out the whole gradient). RNS's $\pm1$ encoding avoids this so that a false value is $-1$ instead of $0$, so it never multiplicatively annihilates an entire gradient. Every input always has the potential to influence the output by changing sign, and the gradient methods used in RNS take advantage of that.

\item \textbf{Gradient Scaling with Clause Size:} In RRL, the magnitude of a gradient component is on the order of $\prod_{k\neq j}\alpha_k$ for some $\alpha_k \in [0,1]$ related to each of the other inputs (e.g. $F_c(h_k, W_{i,k})$). This leads to an \emph{exponential decay} in gradient as $n$ increases, unless all $\alpha_k$ are extremely close to 1. The log-domain trick rescales this by a factor of $P(v)^2/v$, which slows the decay but does not stop it for very large $n$. In RNS, by contrast, a gradient to a decisive input is $\mathcal{O}(1)$ -- it does not diminish with $n$ at all (one input typically gets full gradient 1 if it alone determines the output). In worst-case tie scenarios, the gradient might be split among $n$ inputs, giving each about $1/n$, i.e., decaying linearly with $n$, which is far milder than exponential decay. Thus, RNS scales to clauses with many literals without facing an overwhelming vanishing gradient issue.

\item \textbf{Discrete Training Behavior:} RRL must resort to special training techniques like gradient grafting to handle discrete parameters, because directly backpropagating through a binarized product-form network yields zero gradients in most places. RNS, on the other hand, can be trained with standard backpropagation augmented with STE, which is a simpler and more direct approach. The reason STE works well for RNS is that its surrogate gradient (identity for the sign function, plus the $\min/\max$ subgradient) aligns with actual changes that would affect the output. In RRL's case, an STE would have to assign gradients to inputs that in reality do not affect the output unless combined with others -- a fundamentally ambiguous credit assignment. Therefore, RNS avoids the need for a two-model training setup; one can optimize the $\pm1$ network in one go by using surrogate gradients, without the gradients vanishing.
\end{itemize}

Overall, by using $\pm1$ encoding and min/max logic, RNS preserves discrete interpretability while ensuring that gradient signals remain strong and informative. The network is able to learn large logical formulas (many-input clauses) because it never multiplies a long chain of fractional terms during backpropagation. Each logical neuron in RNS passes a gradient to the input(s) that currently determine its output, providing a clear learning direction. These properties enable RNS to train effectively, where a product-based logical network would struggle or stall due to vanishing gradients.

\end{document}